\documentclass[11pt]{article}

\usepackage{amsopn}

\usepackage{amsmath}
\usepackage{amsfonts}
\usepackage{mathrsfs}
\usepackage{dsfont}

\usepackage{multicol}
\usepackage{booktabs}
\usepackage{array}

\usepackage{amssymb}
\usepackage{enumerate}
\usepackage{graphicx}
\usepackage{mathrsfs}

\usepackage{a4}
\usepackage{url}
\usepackage{algorithm}
\usepackage{algorithmicx}
\usepackage{algpseudocode}
\usepackage{color}

\newcommand{\bea}{\begin{eqnarray*}}
\newcommand{\eea}{\end{eqnarray*}}
\newcommand{\be}{\begin{eqnarray}}
\newcommand{\ee}{\end{eqnarray}}

\def\GP{\mathsf{GP}}
\newcommand{\GPmodel}[1]{\GP(0,\ms_{#1}^2 K^{(#1)})}
\newcommand{\GPmodelg}{\GP(0,\ms^2 K)}
\def\dd{\mathrm{d}}

\def\Arg{\mathrm{Arg}}

\def\diag{\mathrm{diag}}

\def\Pr{\mathrm{Prob}}

\def\tr{\mathrm{trace}}

\def\mt{\theta}

\DeclareMathOperator{\PR}{\mathsf{PR}}

\def\IMSE{\mathsf{IMSE}}
\def\MSE{\mathsf{MSE}}
\def\Bias{\mathsf{Bias}}
\def\ISE{\mathsf{ISE}}

\def\hISE{\widehat{\mathsf{ISE}}}

\def\IM{\mathsf{IM}}

\def\Ex{\mathsf{E}}

\def\var{\mathsf{var}}
\def\cov{\mathsf{cov}}
\def\CVaR{\mathsf{CVaR}}

\def\ma{\alpha}
\def\mg{\gamma}
\def\ml{\lambda}
\def\mL{\Lambda}
\def\mt{\theta}

\def\mo{\omega}
\def\mve{\varepsilon}
\def\mdb{\boldsymbol{\delta}}
\def\ms{\sigma}
\def\Ex{\mathsf{E}}

\def\TT{^\top}
\def\e1{\mathsf{e}}
\def\RR{\mathds{R}}

\def\Ab{\mathbf{A}}
\def\Db{\mathbf{D}}
\def\bDb{\overline{\mathbf{D}}}
\def\Eb{\mathbf{E}}

\def\Hb{\mathbf{H}}
\def\Ib{\mathbf{I}}

\def\Kb{\mathbf{K}}

\def\bKb{\overline{\mathbf{K}}}

\def\nub{\boldsymbol{\nu}}

\def\Mb{\mathbf{M}}
\def\bMb{\overline{\mathbf{M}}}

\def\Qb{\mathbf{Q}}
\def\Rb{\mathbf{R}}
\def\Sb{\mathbf{S}}

\def\Xb{\mathbf{X}}

\def\Zb{\mathbf{Z}}
\def\Wb{\mathbf{W}}
\def\mLb{\boldsymbol{\Lambda}}
\def\mOb{\boldsymbol{\Omega}}
\def\Phib{\boldsymbol{\Phi}}

\def\Gammab{\boldsymbol{\Gamma}}

\def\1b{\mathbf{1}}
\def\0b{\mathbf{0}}
\def\bb{\mathbf{b}}
\def\cb{\mathbf{c}}

\def\eb{\mathbf{e}}

\def\hb{\mathbf{h}}
\def\kb{\mathbf{k}}

\def\rb{\mathbf{r}}
\def\tb{\mathbf{t}}
\def\ub{\mathbf{u}}

\def\wb{\mathbf{w}}
\def\xb{\mathbf{x}}
\def\yb{\mathbf{y}}
\def\zb{\mathbf{z}}
\def\alphab{\boldsymbol{\alpha}}
\def\betab{\boldsymbol{\beta}}
\def\gammab{\boldsymbol{\gamma}}
\def\phib{\boldsymbol{\phi}}
\def\taub{\boldsymbol{\tau}}
\def\mveb{\boldsymbol{\varepsilon}}

\def\SSF{\mathscr{F}}

\def\SN{{\mathscr N}}
\def\SO{\mathcal{O}}

\def\SX{{\mathscr X}}

\newcommand{\vsp}{\vspace{0.3cm}}

\numberwithin{equation}{section}

\title{Weighted Leave-One-Out Cross Validation\thanks{This work was partially funded by project ANR INDEX (ANR-18-CE91-0007).}}

\author{Luc Pronzato\thanks{CNRS, Universit\'e C\^ote d'Azur, Laboratoire I3S, Sophia Antipolis, France
  ({\tt luc.pronzato@univ-cotedazur.fr}, {https://www.i3s.unice.fr/\string~pronzato/}, {\tt rendas@univ-cotedazur.fr}).}
\and Maria-Jo\~ao Rendas\footnotemark[2]}

\begin{document}

\maketitle

\begin{abstract}
We present a weighted version of Leave-One-Out (LOO) cross-validation for estimating the Integrated Squared Error (ISE) when approximating an unknown function by a predictor that depends linearly on evaluations of the function over a finite collection of sites. The method relies on the construction of the best linear estimator of the squared prediction error at an arbitrary unsampled site based on squared LOO residuals, assuming that the function is a realization of a Gaussian Process (GP). A theoretical analysis of performance of the ISE estimator is presented, and robustness with respect to the choice of the GP kernel is investigated first analytically, then through numerical examples. Overall, the estimation of ISE is significantly more precise than with  classical, unweighted, LOO cross validation. Application to model selection is briefly considered through examples.
\end{abstract}


{\bf MSCcodes} 65D05, 62G99, 62-08

\section{Introduction}
The paper addresses the characterization of the performance of data-driven model learning. We consider the fairly general setting where a learning dataset collecting the evaluations of an unknown function $f$ at a given set of sites $\Xb_n=\{\xb_1,\ldots,\xb_n\}$ is used to predict the value of $f$ at generic points $\xb$ in some set $\SX$.
When function evaluations are computationally expensive (for example when they involve complex computer simulations) their number is necessarily limited and the selection of appropriate sites $\xb_i$ is crucial, a problem addressed by the experimental design literature. Regardless of the sites chosen and the prediction method used, it is important to assess the quality of the predictions produced by the learned model. The most common measure of performance is the Integrated Squared (prediction) Error (ISE) for a given measure of interest $\mu$ over $\SX$, and this is the criterion  considered in this paper.

Gaussian Process (GP) models are commonly used in computer experiments to formalize prior knowledge about the behavior of the unknown $f$, as they provide access to the full Bayesian machinery: considering $f$ as a realization of a GP, it is possible to encode prior knowledge on $f$, such as its regularity over $\SX$, and function evaluations $f(\xb_i)$, considered as observations $y_i=Y_{\xb_i}$ on a sample path of the GP, can be used to update knowledge about $f$ in a Bayesian way.
Although in computer experiments GP models are traditionally used to predict values of $f$, in this work we use them to infer the performance of a given predictor, by exploiting the property that fourth-order moments of Gaussian variables are directly available. The predictor whose performance is inferred may be itself the Best Linear Unbiased Predictor (BLUP) derived from a GP model, but this is not mandatory and our approach applies to any predictor \emph{linear} in the observations $y_i$ whose weights may depend arbitrarily on $\xb$. Linearity in the observations is essential in order to preserve the Gaussianity of prediction errors, but this covers a wide range of prediction methods (extension to non-linear predictors is theoretically possible, through Taylor series expansion and the calculation of higher-order moments of Gaussian variables, but seems tedious and rather unpractical). Also, the paper focuses on the case of noise-free observations, i.e.\ situations where we directly observe the response of a deterministic simulator, but the presence of observational noise can be taken into account by introducing a nugget effect into the GP model, and the modifications this implies are briefly indicated in Section~\ref{S:noisy-observations} (some simulation results for noisy observations are presented in Section~\ref{S:noisy-observations-supp}).

A classical approach to evaluate the performance of a prediction method without using observations other than those used to  learn the prediction model itself is Cross Validation (CV), and in particular Leave-One-Out Cross Validation (LOOCV). In LOOCV, the value of $f$ at each site $\xb_i$ is predicted by the same method but removing $\xb_i$ from the learning dataset;
the difference with the observed value $y_i$ determines a residual error $\mve_{-i}$, and the empirical average of the $\mve_{-i}^2$ is used to quantify the overall quality of the prediction method. A large number of papers have addressed cross validation, in particular questioning in what sense it can be considered as an estimate of a statistic of the prediction error, see e.g.\ \cite{BatesHT2023}.
In general, the performance analysis considers that LOOCV provides an estimate of the expected squared prediction error, assuming that the learning dataset is an independent and identically distributed (i.i.d.) sample from the joint distribution of the model covariates and outputs.
Additionally, it usually assumes that the available observations are noisy versions of the model output, an assumption that we relax here. We will show that, as it can be anticipated, for a well spread design $\Xb_n$ LOOCV tends to strongly overestimate the actual conditional ISE, given the learning dataset. The estimator we propose, by using the geometry of the design on which the model has been learned to weight the LOO residuals, is able to overcome this drawback. As an unsought feature, our method is able to cope with the problem of covariate shift, of current interest amongst the community of machine learning \cite{sugiyama2007covariate,xu2022estimation} and to which CV is known to be highly sensitive. Covariate shift occurs when the design in the learning dataset is not a sample from the target distribution under which we want to assess the prediction error: the weights of the corrected LOOCV estimator that we propose depend on the design used to learn the model studied, and thus automatically adjust to covariate shift.

The main objective of the paper is to propose an estimate of the ISE that overcomes some of the limitations of the LOOCV methodology by relying on a GP model, not necessarily stationary, for the fitted function $f$ (the possible extension to mixtures of GP models is briefly considered in Section~\ref{S:mixtures} in the supplement). This allows us to estimate the expected squared prediction error, conditioned on the learning dataset, the expectation being now taken with respect to $f$. Our method builds on \cite{PR2022} and relies on the construction of the best linear estimate of the squared prediction error at any $\xb\in \SX$  based on the set of squared LOO residuals $\mve_{-i}^2$.
Integration with respect to $\xb$ (a simple summation when the measure $\mu$ is discrete) then provides the ISE estimate.
The numerical experiments presented show that our method significantly improves the accuracy of performance evaluation compared to the straightforward application of LOOCV. As the resulting performance measure depends on the assumed GP model, the paper and its supplement include extensive numerical analyses that confirm robustness with respect to its choice.

\section{Notation, motivation and paper organization}\label{S:motivation}

\subsection{Notation}

Consider an $n$-point design $\Xb_n=\{\xb_1,\ldots,\xb_n\}$ without repetitions, i.e., $\xb_i\neq\xb_j$ for all $i\neq j$, and let  $\yb_n=[y_1,\ldots,y_n]\TT$, where $y_i=f(\xb_i)$ is the observation at site $\xb_i$\footnote{By default  all vectors are column vectors.}. We denote by $\SSF_n$ the learning dataset: $\SSF_n = \{(\xb_i, y_i)_{i=1}^n\}$ --- a sub-index $n$ will indicate dependency of the corresponding quantity on $\SSF_n$ (sometimes, only on the design $\Xb_n$).

Let $\eta_n(\xb)$ be the (arbitrary) predictor of $f(\xb)$ at an unsampled site $\xb$, learned using $\SSF_n$, whose performance we want to assess, and denote by
\bea
\mve_n(\xb)=f(\xb)-\eta_n(\xb)
\eea
its prediction error at $\xb$. 
Our goal is to estimate, using only the dataset $\SSF_n$, the ISE for some given positive measure of importance $\mu$ over $\SX$:
\be\label{ISE}
\ISE(\eta_n) =\int_\SX \mve_n^2(\xb)\,\mu(d\xb) \,. 
\ee

Without any loss of generality, we assume that $\Xb_n\in\SX^n$.
All predictors $\eta_n$ considered in the paper are linear, being  defined by a weight function $\wb(\cdot,\cdot): (\xb,\Xb_n)\in\SX\times\SX^n \to \wb(\xb,\Xb_n)\in\RR^n$, such that the prediction of $f(\xb)$ based on $\yb_n$ is $\eta_n(\xb)=\wb\TT(\xb,\Xb_n)\yb_n$, for any $n\in\mathds{N}$ and any $\Xb_n$. In the following, we use the simpler notation $\wb_n(\xb)=\wb(\xb,\Xb_n)$ --- note that $\eta_n$ does not necessarily interpolate the data, i.e., we may have $\eta_n(\xb_i)\neq y_i$; the prediction error
is therefore not necessarily null at the $\xb_i$. We always assume that $\wb_n(\cdot)$ is bounded on $\SX$.

The estimator of $\ISE(\eta_n)$ proposed in this paper, $\hISE_{BLP}(\eta_n)$, assumes that $f$ is a realization of a GP $Y_\xb \sim \GP(0,\ms^2 K)$, indexed by $\xb$ in $\SX$.
Here $K$ is a Strictly Positive Definite (SPD) kernel and $Y_\xb \sim \GP(0,\ms^2 K)$ means that $\Ex\{Y_\xb\}=0$ and $\Ex\{Y_\xb Y_{\xb'}\}=\ms^2 K(\xb,\xb')$ for all $\xb$ and $\xb'$ in $\SX$.
Throughout the paper, $\kb_n(\xb)$ is the vector with components $K(\xb,\xb_i)$, $i=1,\ldots,n$, and $\Kb_n$ is the $n\times n$ matrix with $\{\Kb_n\}_{i,j}=K(\xb_i,\xb_j)$, $i,j=1,\ldots,n$. We will denote $\Mb_n = \Kb_n^{-1}$ (and to simplify notation the subindex $n$ will be dropped, i.e., $\Mb\equiv\Mb_n$).

The assumption of a GP model for $f$ is important for at least three reasons: (\textit{i}) it is extremely convenient for evaluating the performance of an arbitrary ISE estimator for a given linear predictor $\eta_n$; (\textit{ii}) it is essential for the derivation of the ISE estimator we propose; (\textit{iii}) it is commonly used to define a predictor $\eta_n$.
We thus need to carefully distinguish between three possibly different GP models in the developments presented below.

We will denote by $\GPmodelg$ the (``true'') data generating model. In a real situation, the data $f(\xb_i)$ are not samples from $\GPmodelg$, but this assumption allows explicit computation of the bias and Mean Squared Error (MSE) of the different ISE estimators considered. When we will need to distinguish it from $\GPmodelg$, the GP model assumed for the construction of our estimator $\hISE_{BLP}(\eta_n)$ will be denoted by $\GPmodel{e}$. Indeed, there is no reason to assume that $\GPmodelg \equiv \GPmodel{e}$, and we will investigate the impact of the possible modeling mismatch $\GPmodelg \neq \GPmodel{e}$. Finally, the predictor $\eta_n$ for which we want to estimate the ISE may itself rely on a certain GP model, which we shall denote by $\GPmodel{p}$, where $K^{(p)}$ differs in general from $K$ and $K^{(e)}$.
When necessary, a superscript $^{(e)}$ or $^{(p)}$ will explicitly indicate the underlying model to which we refer.
An essential feature of our method is that it relies solely on predictions (it does not require estimation of the scaling parameter $\ms^2$), so the choice of kernel is not very critical: indeed, predictions based on GP models are known to be robust to the choice of kernel $K$, which is not the case for the prediction of their accuracy; see, e.g., \cite[Sect.~3.5]{Stein99}; one may also refer to \cite{Bachoc2013, KarvonenWTOS2020, NaslidnykKKM2023} for results on the estimation of $\ms^2$ in a GP model and consequences on model calibration.

That we always suppose a GP with zero mean may appear as a severe restriction, as the hypothesis according to which $f$ is a realization of a centred GP is often  difficult to maintain in practice. However, the presence of a linear trend $\taub\TT\hb(\xb)$, with $\hb(\cdot)=[h_1(\cdot),\ldots,h_n(\cdot)]\TT$ a vector of $p$ functions on $\SX$, has no effect on our ISE estimator when the predictor satisfies $[h_i(\xb_1),\ldots,h_i(\xb_n)]\wb_n(\xb)=h_i(\xb)$ for all $i=1,\ldots,p$, all $n$ and all $\xb$ and $\xb_1,\ldots,\xb_n$. This is the case in particular for universal kriging (and when the trend is a constant mean, i.e., $\hb(\xb) \equiv 1$ for all $\xb$, the condition $\sum_{i=1}^n w_i(\xb)=1$ for all $n$ and all $\xb$ is satisfied by the ordinary kriging predictor). See Section~\ref{S:parameterised-mean} in the supplement for details, including a simple and convenient modification of the estimator for the case when the weights $\wb_n(\xb)$ do not satisfy the condition above.

The kernels used in the examples presented in the paper are all isotropic (i.e., $K(\xb,\xb')$ only depends on $\|\xb-\xb'\|$). However, this is not mandatory and any kernel $K^{(e)}$ can be used to construct $\hISE_{BLP}(\eta_n)$. This means that, independently of the particular predictor whose performance is to be assessed, we may choose the kernel we think is the most appropriate as a GP model for $f$: $K^{(e)}$ can be non isotropic, non stationary, and one may even consider a mixture of GP models (see Section~\ref{S:mixtures} in the supplement), which offers considerable flexibility.

\subsection{Motivation}

To  motivate our work and precise the basic notions that we shall use, in this introductory section we consider the case when $\GPmodel{e}=\!\GPmodel{p}=\!\GPmodelg$ and  $\eta_n\equiv \eta^*_n$, the Best Linear Unbiased Predictor (BLUP) of $f$ given $\SSF_n$; that is, $\eta_n$ is the  simple kriging interpolator (the posterior expectation under the GP model):
\be\label{eta_n}
\eta^*_n(\xb)= \Ex\left\{\left.Y_\xb \right| \SSF_n  \right\} = \kb_n\TT(\xb)\Kb_n^{-1}\yb_n \,.
\ee

The posterior variance of $Y_\xb$ is independent of the observations $\yb_n$:
$\var\{\left.Y_\xb \right| \SSF_n\} = \Ex\{[Y_\xb-\eta^*_n(\xb)]^2|\SSF_n\}=\Ex\{[Y_\xb-\eta^*_n(\xb)]^2|\Xb_n\}$, and direct calculation gives $\var\{\left.Y_\xb \right| \SSF_n\}=\ms^2\,{\rho_n^*}^2(\xb)$, where
\be\label{rho_n^2-a}
{\rho_n^*}^2(\xb) = K(\xb,\xb)- \kb_n\TT(\xb)\Kb_n^{-1}\kb_n(\xb)
\ee
is the (simple) kriging variance. As the learning design $\Xb_n$ is fixed, all expectations are conditioned on $\Xb_n$ but in the following we shall simply write $\Ex\{\cdot\}=\Ex\{\cdot|\Xb_n\}$. The assumption that $\eta_n \equiv \eta_n^*$ for a given GP model will be flagged with an asterisk.

For a given (positive) measure of importance $\mu$ on $\SX$, the Integrated Squared Error $\ISE(\eta_n)$ defined in \eqref{ISE} is a natural criterion for measuring the overall predictive quality of $\eta_n$ over $\SX$. Since $Y_\xb$, and thus $\mve_n(\xb)$, is unknown for $\xb\not\in\Xb_n$, $\ISE(\eta_n)$ is not computable. However, under the GP assumption for $Y_\xb$ its expected value (the Integrated Mean Squared Error) is given by
\be\label{IMSE}
\IMSE(\eta^*_n)= \int_\SX \Ex\{\mve_n^2(\xb)|\SSF_n \} \,\mu(\dd\xb) = \ms^2 \int_\SX {\rho_n^*}^2(\xb)  \,\mu(\dd\xb) \,.
\ee
As the process variance $\ms^2$ only appears as a multiplicative factor in \eqref{IMSE}, the design $\Xb_n$ that minimizes $\IMSE(\eta^*_n)$ is independent of $\ms^2$; see e.g.\ \cite{SacksSW89} for an early reference and \cite{GP-SIAM_2014, GP-CSSC_2016} for methods that avoid repeated computations of integrals when constructing an IMSE-optimal design.
Note that unlike the method proposed in this paper, which further exploits knowledge of $\SSF_n$, direct use of \eqref{IMSE} to quantify the predictive quality of the model learned over a design $\Xb_n$ requires knowledge of $\ms^2$, the process variance\footnote{Numerical investigations show that methods that rely on the estimation of $\ms^2$, e.g.\ by Maximum Likelihood or LOOCV, and estimate $\ISE(\eta_n)$ by $\IMSE(\eta_n)$ perform worse than the one proposed in the paper; see Section~\ref{S:conclusion} for a brief discussion.}.

Several methods have been proposed to assess the performance of predictors using the learning dataset $\SSF_n$, amongst which cross validation  and in particular the  LOOCV criterion \cite{Stone74}:
\be\label{ISE_{LOO}}
\hISE_{LOO}(\eta_n) = \frac1n \, \sum_{i=1}^n \mve_{-i}^2 \,,
\ee
where $\mve_{-i}=y_i-\eta_{n\setminus i}(\xb_i)$, with $\eta_{n\setminus i}$ the predictor constructed without the $i$-th design point $\xb_i$.
As explained in Section~\ref{S:LOOCV-GPmodel}, the actual performance of $\hISE_{LOO}(\eta_n)$ as an estimator of $\ISE(\eta_n)$ strongly depends on the design used to construct $\eta_n$. The ISE estimator that we propose optimally weights the squared LOO residuals $\mve_{-i}^2$ by constructing the Best Linear Predictor (BLP) of $\mve_n^2(\xb)$ at an unsampled site $\xb$ (it minimizes a squared loss and is linear in the $\mve_{-i}^2$), and thereby ensures robust performance with respect to the design configuration. Central to the method is the fact that all moments required to calculate this BLP are directly available, thanks to the Gaussian assumption.

\subsection{Paper organization}
In Section~\ref{S:LOOCV-GPmodel}, we derive the bias, variance and MSE of ISE estimators that are linear in the squared LOO residuals $\mve_{-i}^2$, under the GP model assumption and for a general linear predictor $\eta_n$. Results for $\hISE_{LOO}(\eta_n)$ follow as a particular case. The special case where $\eta_n\equiv\eta_n^*$, the BLUP for the assumed GP model, is considered in Section~\ref{S:LOO-etan=etan*}. In Section~\ref{S:ISE-any-predictor}, we derive our estimator $\hISE_{BLP}(\eta_n)$, the integral of the BLP $\widehat{\mve^2_n}(\xb)$ of the squared prediction error $\mve^2_n(\xb)$ based on the squared LOO residuals $\mve_{-i}^2$. Its mean and MSE are given in Section~\ref{S:E-MSE-no-error} under the assumption that $K^{(e)}=K$ (i.e., in absence of modeling error). In Section~\ref{S:ISE-BLP} we assume that $\eta_n\equiv\eta_n^*$, the BLUP for $\GPmodelg$, and in Section~\ref{S:robustness} we consider the performance of $\hISE_{BLP}(\eta_n)$ for a general linear predictor $\eta_n$ in presence of model misspecification ($K^{(e)} \neq K$). Two limiting behaviors for $K^{(e)}$ are briefly investigated in Section~\ref{S:limits}: the independent limit where the correlation between $Y_\xb$ and $Y_{\xb'}$ is negligible when $\xb\neq\xb'$, the flat limit where conversely $Y_\xb$ and $Y_{\xb'}$ remain strongly correlated when $\|\xb'-\xb\|$ is large.
The BLP $\widehat{\mve^2_n}(\xb)$ of Section~\ref{S:ISE-any-predictor} is biased, and therefore $\hISE_{BLP}(\eta_n)$ is biased too; a bias-corrected version (in absence of model misspecification) is presented in Section~\ref{S:ISE_BLUP}. Modifications implied by the presence of observation noise are briefly discussed in Section~\ref{S:noisy-observations}. A numerical investigation of the performance of the estimators $\hISE_{LOO}(\eta_n)$ and $\hISE_{BLP}(\eta_n)$ is carried out in Section~\ref{S:numerical} where several numerical examples are presented. First, in Section~\ref{S:design-influence} we illustrate the influence of the design $\Xb_n$ on both estimators for a univariate function $f$. Then, in Section~\ref{S:numerical-robustness} we study the robustness of $\hISE_{BLP}(\eta_n)$ with respect to the assumed kernel $K^{(e)}$, considering kernels with different correlation lengths; different predictors $\eta_n$ are also considered: a non-interpolating polynomial (Section~\ref{S:polynomial}) and the BLUP for a GP model $\GPmodel{p}$ (Section~\ref{S:BLP-other-K}). Additional material is provided in Section~\ref{S:different-regularity} of the supplement and involves kernels $K^{(e)}$ with different regularities. While in Sections~\ref{S:design-influence} and \ref{S:numerical-robustness} $Y_\xb\sim\GPmodelg$, in Sections~\ref{S:environmental-model} and \ref{S:piston-model} we consider test-cases from the literature, with $f$ depending on 2 and 4 variables, respectively. Other numerical results are presented in the supplement: average performance characteristics of  $\hISE_{BLP}(\eta_n)$ and $\hISE_{LOO}(\eta_n)$ for GP realizations with $n\in\{10\,d, 20\,d, 50\,d, 100\,d, 200\,d\}$ and $d\in\{4,6,8\}$ in Section~\ref{S:behavior-d-n}; empirical performance for random functions that are not GP realizations in Section~\ref{S:f-not-GP}, including the introduction of observation noise in Section~\ref{S:noisy-observations-supp}. Section~\ref{S:conclusion} concludes and draws some perspectives for this work.

That we propose to estimate the ISE \eqref{ISE} by another integral may seem numerically cumbersome. However, the integrated function is known explicitly, and the integral can be approximated by Quasi-Monte Carlo with small computational cost --- an indication of computational times for a Matlab implementation is provided in Sections~\ref{S:environmental-model}, \ref{S:piston-model} and \ref{S:simulations-m-n-d} in the supplement. In fact, the method proposed does not address the integration problem itself, but rather the estimation $\widehat{\mve_n^2}(\xb)$, in the mean square sense,
of the squared prediction error $\mve_n^2(\xb)$ at any $\xb$. Note, however, that we cannot predict precisely the value of $\mve_n^2(\xb)$ at every $\xb$ (and an example presented in Section~\ref{S:quantiles} shows that the approach is unreliable for estimating a tail characteristic such as a quantile or a conditional value-at-risk of $\widehat{\mve_n^2}(X)$ with $X\sim\mu$, for one realization of $Y_\xb\sim\GP(0,\ms^2\,K)$, as the estimated value $\widehat{\mve_n^2}(X)$ and $\mve_n^2(X)$ have different distributions). Estimation of $\ISE(\eta_n)$ by the integral of $\widehat{\mve_n^2}(\xb)$ can be justified by ergodicity arguments when $K$ is stationary, $K(\xb,\xb')\to 0$ as $\|\xb-\xb'\|\to\infty$, and the support of $\mu$ is large enough.

\section{LOOCV for linear predictors under the GP  assumption}\label{S:ISE_LOO}

For linear predictors, of the form  $\eta(\xb) = \wb_n\TT(\xb) \yb_n$, the vector of LOO residuals $\mveb_{LOO}=(\mve_{-1},\ldots,\mve_{-n})\TT$ is also linear in $\yb_n$ and can be written as
\be\label{Rn}
\mveb_{LOO} =  \Rb_n\TT \yb_n \,,
\ee
where
$\Rb_n = \Ib_n - \Wb_{n\setminus}$, with $\Ib_n$ the $n$-dimensional identity matrix and $\Wb_{n\setminus}$ an $n\times n$ matrix whose diagonal is identically null (as  $\mve_{-i}=y_i-\eta_{n\setminus i}(\xb_i)$ and $\eta_{n\setminus i}(\xb_i)$ does not depend on $y_i$). We assume that $\Rb_n$ has full rank.

The assumption that $f$ is a realization of a GP $Y_\xb \sim \GPmodelg$ allows us to derive the expressions of the first two moments of an arbitrary linear combination $\gammab\TT\mveb_{LOO}^{\odot 2}$ of the squared LOO residuals $\mveb_{LOO}^{\odot 2}=(\mve_{-1}^2,\ldots,\mve_{-n}^2)\TT$, $\gammab\in\RR^n$. (Here and in the following we denote by $\Ab^{\odot 2}$ the Hadamard square of matrix $\Ab$: $\{\Ab^{\odot 2}\}_{ij}=\Ab_{ij}^2$.) As the LOOCV criterion $\hISE_{LOO}(\eta_n)$ corresponds to $\gammab=\1b_n/n$ with $\mathbf{1}_n$ the $n$-dimensional vector with all components equal to 1, see \eqref{ISE_{LOO}}, we directly obtain the bias and MSE of $\hISE_{LOO}(\eta_n)$.

\subsection{Consequences of the GP assumption and notation}\label{S:LOOCV-GPmodel}

Under the GP assumption $Y_\xb\sim\GPmodelg$, the prediction error $\mve_n(\xb)=Y_\xb-\eta_n(\xb)$ has zero mean, $\Ex\{\mve_n(\xb)\}=0$, and we denote
\be\label{rho_n^2-b}
\rho_{n}^2(\xb) = \Ex\{\mve_n^2(\xb)/\ms^2\}= K(\xb,\xb)-2\,\wb_n\TT(\xb)\kb_n(\xb)+\wb_n\TT(\xb)\Kb_n\wb_n(\xb) \,.
\ee
The vector of LOO residuals $\mveb_{LOO}$ is also Gaussian and, for all $\xb\in\SX$,
\[
\Ex\{\mveb_{LOO}\}=\mathbf{0},\qquad  \Ex\{\mveb_{LOO}\mveb_{LOO}\TT\}=\ms^2 \Rb_n\TT\Kb_n\Rb_n,\qquad \Ex\{\mveb_{LOO}\mveb_n(\xb)\}=\ms^2\,\Rb_n\TT\tb_n(\xb)\,,
\]
where
\be\label{tb_n}
\tb_n(\xb)= \Ex\{\yb_n \mve_{n}(\xb)\}/\ms^2 = \kb_n(\xb)-\Kb_n\wb_n(\xb) \,.
\ee
Using the formula for the expectation of the product of squared Gaussian variables $a$ and $b$,
\bea
\Ex\{a^2b^2\} = \Ex\{a^2\}\Ex\{b^2\}+2\, [\Ex\{ab\}]^2 \,,
\eea
we get
$\Ex\{\mve_n^2(\xb)\mve_n^2(\xb')\}=\ms^4\left[\rho_n^2(\xb)\rho_n^2(\xb')+2\, \rho_n^4(\xb,\xb')\right]$,
where $\rho_n^2(\xb)$ is given by \eqref{rho_n^2-b} and
\bea 
\rho_n^2(\xb,\xb')\!=\!\frac{1}{\ms^2}\, \Ex\left\{\mve_n(\xb)\mve_n(\xb')\right\}=K(\xb,\xb')-\wb_n\TT(\xb)\kb_n(\xb')-\wb_n\TT(\xb')\kb_n(\xb)+\wb_n\TT(\xb)\Kb_n\wb_n(\xb').
\eea
In the same manner, we obtain for the (normalized) first two moments of $\mveb_{LOO}^{\odot 2}$,
\be
\ub_n = \Ex\{\mveb_{LOO}^{\odot 2}\}/\ms^2 &=& \diag\left\{(\Rb_n\TT\Kb_n\Rb_n)\right\}, \label{un} \\
\Sb_n = \Ex\{\mveb_{LOO}^{\odot 2} {\mveb_{LOO}^{\odot 2}}\TT \}/\ms^4
&=& \Ex\{\mveb_{LOO}^{\odot 2}/\ms^2\}\Ex\{{\mveb_{LOO}^{\odot 2}}\TT/\ms^2\}+2 \left( \Ex\{\mveb_{LOO}\mveb_{LOO}\TT/\ms^2\} \right)^{\odot 2} \nonumber \\
&=& \ub_n\ub_n\TT + 2\, (\Rb_n\TT\Kb_n \Rb_n)^{\odot 2} \,, \label{Sn}
\ee
together with
\be
\cb_n(\xb) = \Ex\{\mve_n^2(\xb)\mveb_{LOO}^{\odot 2}\}/\ms^4 &=&  \Ex\{\mve_n^2(\xb)/\ms^2\}\Ex\{\mveb_{LOO}^{\odot 2}/\ms^2\} + 2 \left( \Ex\{\mve(\xb)\mveb_{LOO}/\ms^2\} \right)^{\odot 2}\nonumber \\
&=&  \rho_{n}^2(\xb)\ub_n +2\,[\Rb_n\TT\tb_n(\xb)]^{\odot 2} \,. \label{cn}
\ee

\begin{table}
\begin{center}
\caption{Moments of quantities of interest.\label{T:defsLOOCV}}
\begin{tabular}{|c|c|c|c|c|c|c|}
\hline 
& b & $\ms^2$ & ${\mveb_{LOO}^{\odot 2}}\TT$ & $\ms\,\mve_n(\xb)$ & $\mve_n^2(\xb)$ &  $\ISE(\eta_n)$  \\ \hline 
$a$ & $\Ex\{ab\}/\ms^4$ & $*$ & $*$ & $*$ & $*$ & $*$ \\ \hline
$\ms^2$ & $*$ & 1 & $\ub_n\TT$ &  $0$ & $\rho_n^2(\xb) $ & $J_n$ \\ \hline
$\mveb_{LOO}^{\odot 2}$ & $*$ & $\ub_n$ & $\Sb_n$ & $\0b$ & $\cb_n(\xb)$ & $\bb_n$\\ \hline
$\ms\,\mve_n(\xb^\prime)$ & $*$ & 0 & $\0b$ & $\rho_n^2(\xb,\xb^\prime)$ & 0 & 0  \\ \hline
$\mve_n^2(\xb^\prime)$  & $*$ & $\rho_n^2(\xb^\prime)$ & $\cb_n\TT(\xb^\prime)$ & 0 & $\rho_n^2(\xb)\rho_n^2(\xb^\prime)+2\, \rho_n^4(\xb,\xb^\prime)$ &  \\ \hline
$\ISE(\eta_n)$ & $*$ & $J_n$ & $\bb_n\TT$ & 0 & & $J_n^2+2\,V_n$\\ \hline
\end{tabular}
\end{center}
\end{table}

With these definitions (summarized in Table~\ref{T:defsLOOCV}), we can compute the bias, variance and MSE of any linear combination $\hISE(\eta_n)=\gammab\TT\mveb_{LOO}^{\odot 2}$, $\gammab\in\RR^n$:
\be
\Bias\{\hISE(\eta_n)\} &=& \Ex\left\{\hISE(\eta_n) - \ISE(\eta_n) \right\} =  \ms^2 \, \gammab\TT \ub_n - \Ex\left\{\ISE(\eta_n) \right\}\,, \label{Bias-gamma} \\
\var\{\hISE(\eta_n)\} &=& \Ex\left\{\hISE^2(\eta_n)\right\} - \left[\Ex\left\{\hISE(\eta_n)\right\}\right]^2 =  \ms^4 \gammab\TT (\Sb_n-\ub_n\ub_n\TT) \gammab \nonumber \\
&& \hspace{5.4cm} = 2\,\ms^4 \gammab\TT (\Rb_n\TT\Kb_n \Rb_n)^{\odot 2} \gammab \,, \label{var-gamma}  \\
\MSE\{\hISE(\eta_n)\} &=& \Ex \left\{ \left( \hISE(\eta_n) - \ISE(\eta_n) \right)^2 \right\}  \nonumber \\
&=&  \ms^4 \gammab\TT \Sb_n \gammab  - 2\,\ms^4 \gammab\TT \bb_n + \var\left\{\ISE(\eta_n) \right\} + \left[\Ex\left\{\ISE(\eta_n) \right\}\right]^2 \,, \label{MSE-gamma}
\ee
with
\be
\Ex\left\{\ISE(\eta_n) \right\} &=& \IMSE(\eta_n) =  \ms^2 J_n\,,  \label{IMSE_arbitrary} \\
\var\left\{\ISE(\eta_n) \right\} &=& 2\,\ms^4\, V_n \,, \label{varISE}
\ee
where we have introduced
\be
\bb_n &=& \int_\SX \cb_n(\xb)\,\mu(\dd\xb) \,, \label{bn} \\
J_n &=& \int_\SX \rho_{n}^2(\xb)\,\mu(\dd\xb)\,, \label{Jn}\\
V_n & = & \int_{\SX^2} \rho_n^4(\xb,\xb^\prime)\mu(\dd\xb)\mu(\dd\xb^\prime) \,. \label{Vn} 
\ee

We thus obtain for the LOOCV estimator \eqref{ISE_{LOO}} (for which $\gammab=\1b_n/n$)
\be
\Bias\{\hISE_{LOO}(\eta_n)\} &=&   \frac{\ms^2}{n} \mathbf{1}_n\TT \ub_n - \IMSE(\eta_n)\,, \label{Bias-LOO} \\
\MSE\{\hISE_{LOO}(\eta_n)\} &=&  \ms^4\left[\frac{\mathbf{1}_n\TT \Sb_n \mathbf{1}_n}{n^2}  - 2\,\frac{\mathbf{1}_n\TT \bb_n}{n} + J_n^2+2\,V_n \right]\,. \label{MSE-LOO}
\ee

\subsection{BLUP for the assumed GP model}\label{S:LOO-etan=etan*}
We consider here the special case where $\eta_n$ is the BLUP $\eta^*_n$  for the GP from which $f$ is drawn, i.e., $\GPmodel{p} \equiv \GPmodelg$, a situation where simpler expressions can be found since $\wb_n(\xb)=\Kb_n^{-1}\kb_n(\xb)$, see \eqref{eta_n}. Indeed, it implies $\tb_n(\cdot) \equiv 0$, see \eqref{tb_n}, and $\rho_n^2(\xb)$ defined by \eqref{rho_n^2-b} equals ${\rho_n^*}^2(\xb)$ given by \eqref{rho_n^2-a}.

Since the LOO residual $\mve_{-i}$ is computed using the BLUP $\eta^*_n$ that leaves out $(\xb_i,y_i)$, i.e.,
\bea
\mve_{-i} = y_i-\eta^*_{n\setminus i}(\xb_i), \qquad i=1,\ldots,n \,,
\eea
where $\eta^*_{n\setminus i}(\xb)=\kb_{n\setminus i}\TT(\xb)\Kb_{n\setminus i}^{-1}\yb_{n\setminus i}$ (with obvious notation), straightforward use of the block-matrix inversion formula gives
\bea
\Rb_n = \Mb\Db_n\,,
\eea
where $\Mb=\Kb_n^{-1}$ and $\Db_n=\diag\{1/\Mb_{ii}, i=1,\ldots,n\}$, with
\bea
\Mb_{ii}=\left[K(\xb_i,\xb_i)-\kb_{n\setminus i}\TT(\xb_i)\Kb_{n\setminus i}^{-1}\kb_{n\setminus i}\TT(\xb_i)\right]^{-1}=1/[{\rho_{n\setminus i}^*}^2(\xb_i)]
\eea
the $i$-th diagonal element of $\Mb$; see \cite{Dubrule83} (one may also refer to \cite{GinsbourgerS2021} for the extension to multiple-fold CV). Note that in this case $\Rb_n$ always has full rank.

To highlight the difference with previous section, we insert an asterisk in superscript for $\cb_n$, $\ub_n$ and $\Sb_n$ and write
\be
\cb_n^*(\xb) &=& \ub_n^* \, {\rho_n^*}^2(\xb)\,,  \nonumber \\
\Sb_n^* &=& \ub_n^*{\ub_n^*}\TT + 2\, \Db_n^{\odot 2} \Mb^{\odot 2} \Db_n^{\odot 2}\,, \label{Sn^*}
\ee
with
\be\label{un^*}
\ub_n^* &=& (1/\Mb_{11},\ldots,1/\Mb_{nn})\TT \,.
\ee
The LOOCV criterion is then equal to
\be \label{ISE_{LOO}_gp}
\hISE_{LOO}(\eta^*_n) = \frac1n \, \sum_{i=1}^n \mve_{-i}^2 = \frac1n\, \yb_n\TT \Mb\Db_n^2\Mb\yb_n \,,
\ee
and its expectation for a given design $\Xb_n$ is
\begin{align*}
\Ex\left\{\hISE_{LOO}(\eta^*_n)\right\}
&=\frac{\ms^2}{n}\, \tr\left( \Mb\Db_n^2 \Mb \Kb_{n}\right) = \frac{\ms^2}{n}\, \sum_{i=1}^n \Mb_{ii} {\Db_n}_{ii}^2\\
&=\frac{\ms^2}{n}\,\tr\left(\Db_n \right) = \frac{\ms^2}{n}\, \sum_{i=1}^n \frac{1}{\Mb_{ii}} = \frac{\ms^2}{n}\, \sum_{i=1}^n {\rho_{n\setminus i}^*}^2(\xb_i)\,.
\end{align*}

To shed some light on the issues involved in using \eqref{ISE_{LOO}} as a measure of prediction accuracy, let us consider two extreme situations: (\textit{i}) $\Xb_n$ is such that each design point has another one in its vicinity; (\textit{ii}) $\Xb_n$ is such that all design points are far enough from each other to have a negligible correlation between $Y_{\xb_i}$ and $Y_{\xb_j}$ for $i\neq j$. In the first case, since when $\xb_i$ is dropped there is another design point in its neighborhood, we have $\ub_n \simeq 0$, and thus $\hISE_{LOO}(\eta_n)$ is overoptimistic, with a negative bias close to $-\IMSE(\eta_n)$, see \eqref{Bias-LOO}. The behavior in case (\textit{ii}) depends on $\eta_n$, but for any reasonable predictor such that on average the accuracy of $\eta_n(\xb)$ improves when $\xb$ is closer to a design point $\xb_i$ (and thus the correlation between $Y_{\xb}$ and $Y_{\xb_i}$ increases), on average $\mve_{-i}^2$ will be larger than $\mve_n^2(\xb)$: $\hISE_{LOO}(\eta_n)$ will thus tend to be pessimistic, with a positive bias.
One may refer to Section~\ref{S:design-influence} for an illustrative example.

The actual performance of $\hISE_{LOO}(\eta_n)$ therefore strongly depends on the design configuration: its attractive features pointed out in the literature are in fact valid on average, for designs $\Xb_n$ with $\xb_i \stackrel{\scriptsize\rm{i.i.d.}}{\sim} \mu$. Note that when the $\xb_i$ are designed to ensure precise prediction of $f$ over $\SX$, they are usually space-filling and thus far from resembling an i.i.d.\ sample, see, e.g., \cite{PM_SC_2012}. This situation approaches case (\textit{ii}) and we will see in Sections~\ref{S:numerical-robustness} to \ref{S:piston-model} that indeed $\hISE_{LOO}(\eta_n)$ tends to overestimate $\ISE(\eta_n)$.

\section{Best linear estimation of the ISE based on squared LOO residuals}\label{S:ISE_BLP}

The LOOCV estimator \eqref{ISE_{LOO}} has two antagonist distinctive features: (\textit{i})
it is free of any modeling assumption, and  can thus be used to infer the predictive quality of any predictor $\eta_n$;  (\textit{ii}) it is agnostic to the geometry of the design $\Xb_n$, and is thus unable to capture its impact on the expected errors in regression problems, being thus highly sensitive to  the covariate shift problem \cite{sugiyama2007covariate}. On the one hand, the estimator we propose in this paper loses the universality of feature (\textit{i}) as it is strongly grounded on the GP assumption for $f$ and is derived for linear predictors only. On the other hand, its parameters are tuned to the design geometry, so that the impact of this geometry on the estimated error is correctly accounted for. A key advantage over LOOCV is thus that we can choose the design $\Xb_n$ by focusing solely on the accuracy of $\eta_n$, without worrying about the possible impact of the choice of $\Xb_n$ on the precision of the estimate of $\ISE(\eta_n)$.

Our estimator, $\hISE_{BLP}(\eta_n)$, relies on the assumption that $Y_\xb \sim \GPmodel{e}$ for some some SPD kernel $K^{(e)}$. In Sections~\ref{S:ISE-any-predictor}
to \ref{S:ISE-BLP} we assume that $\GPmodel{e}\equiv\GPmodelg$ and drop the superscript $^{(e)}$, but in Section~\ref{S:robustness} we investigate the performance (bias and MSE) of $\hISE_{BLP}(\eta_n)$ in the case where the data generating model $\GPmodelg$ differs from the assumed model $\GPmodel{e}$; see also Sections~\ref{S:different-regularity} and \ref{S:behavior-d-n} of the supplement.

\subsection{Best linear estimation of $\mve_n^2(\xb)$}\label{S:ISE-any-predictor}

We consider estimation of the squared prediction error $\mve_n^2(\xb)$ of a linear predictor $\eta_n$ at a generic point $\xb\in\SX$, based on the $n$ observed squared LOO residuals $\mveb_{LOO}^{\odot 2}$, assuming that $Y_\xb\sim\GPmodelg$. The linear estimator that minimizes $\Ex\{[\mve_n^2(\xb)-\betab\TT\mveb_{LOO}^{\odot 2}]^2\}$ with respect to $\betab\in\RR^n$ is
\be
\widehat{\mve_n^2}_{BLP}(\xb) = \widehat\betab\TT\!\!(\xb) \mveb_{LOO}^{\odot 2}, \qquad \mbox{with} \qquad \widehat\betab(\xb) = \Sb_n^{-1} \cb_n(\xb) \,, \label{beta_BLP}
\ee
where $\Sb_n=\Ex\{\mveb_{LOO}^{\odot 2}{\mveb_{LOO}^{\odot 2}}\TT\}/\ms^4$ and $\cb_n(\xb)=\Ex\{\mve_n^2(\xb)\mveb_{LOO}^{\odot 2}\}/\ms^4$ are respectively given by \eqref{Sn} and \eqref{cn}. The assumption that $\Rb_n$ has full rank implies that $\Sb_n$ is invertible.

One may notice that when $\wb_n(\xb_i)=\eb_i$, the $i$-th canonical basis vector, for all $i=1,\ldots,n$ (which is the case for example when $\eta_n$ is a kriging predictor for a kernel $K^{(p)}$), then $\rho_n^2(\xb_i)=0$ and $\tb_n(\xb_i)=\0b$ for all $i$, see \eqref{rho_n^2-b} and \eqref{tb_n}, and therefore $\cb_n(\xb_i)=\0b$, see \eqref{cn}, implying that $\widehat{\mve_n^2}_{BLP}(\xb_i)=0$ for all $i$. This is however not necessarily the case for an arbitrary predictor $\eta_n$.

The estimate of $\ISE(\eta_n)$ proposed in this paper is
\be\label{estimate-ISE-general}
\hISE_{BLP}(\eta_n) &=& \int_\SX \widehat{\mve_n^2}_{BLP}(\xb)\,\mu(\dd\xb) = {\mveb_{LOO}^{\odot 2}}\TT\Sb_n^{-1} \int_\SX \cb_n(\xb)\,\mu(\dd\xb) = {\mveb_{LOO}^{\odot 2}}\TT\Sb_n^{-1}\bb_n \,, 
\ee
with $\bb_n$ given by \eqref{bn}. When $\mu$ is approximated by a discrete measure on $N$ points $\xb^{(i)}$, $i=1,\ldots,N$, the complexity of the evaluation of $\hISE_{BLP}(\eta_n)$ is of the order $\SO(N n^3)$ ($\SO(n^3)$ for the evaluation of each $\widehat{\mve_n^2}_{BLP}(\xb^{(i)})$). As the minimization of $\MSE\{\hISE(\eta_n)\}$ with respect to $\gammab$ in \eqref{MSE-gamma} yields $\widehat\gammab_{BLP}=\Sb_n^{-1}\bb_n=\int_\SX \widehat\betab(\xb)\,\mu(\dd\xb)$,
$\hISE_{BLP}(\eta_n)$ is the best estimator of $\ISE(\eta_n)$ that is linear in the $\mve_{-i}^2$.
Note that there is no guarantee that $\widehat{\mve_n^2}_{BLP}(\xb) = \widehat\betab\TT\!\!(\xb) \mveb_{LOO}^{\odot 2}$ be positive. We keep this $\widehat{\mve_n^2}_{BLP}(\xb)$ in our analysis, but use ${\widehat{\mve_n^2}_{BLP}}^+(\xb)=\max\{\widehat{\mve_n^2}_{BLP}(\xb),0\}$ in the numerical implementation that generated the examples provided in Section~\ref{S:numerical} and in the supplement:
${\widehat{\mve_n^2}_{BLP}}^+(\xb)$ minimizes $\Ex\{[\mve_n^2(\xb)-\betab\TT\mveb_{LOO}^{\odot 2}]^2\}$ with respect to $\betab\in\RR^n$ under the constraint $\betab\TT\mveb_{LOO}^{\odot 2}\geq 0$.

\subsection{Mean and MSE of the best linear ISE estimator (no modeling error)}\label{S:E-MSE-no-error}
The assumption $Y_\xb\sim\GP(0,\ms^2K)$ allows us to compute the statistical moments of $\hISE_{BLP}(\eta_n)$, in particular its bias and MSE. Substituting $\gammab= \widehat\gammab_{BLP}=\Sb_n^{-1}\bb_n$ in \eqref{Bias-gamma} and \eqref{MSE-gamma} we get $\Ex\{\hISE_{BLP}(\eta_n)\} = \ms^2 \bb_n\TT \Sb_n^{-1} \ub_n$
and thus the bias of $\hISE_{BLP}(\eta_n)$ is
\bea
\Bias \{\hISE_{BLP}(\eta_n) \} = \ms^2 \bb_n\TT \Sb_n^{-1} \ub_n-\ms^2 J_n\, .
\eea
Its mean squared error is
\be
\MSE\{\hISE_{BLP}(\eta_n)\} &=& \ms^4(J_n^2+2\,V_n) - \ms^4 \bb_n\TT\Sb_n^{-1}\bb_n \,, \label{MSE-BLP}
\ee
where $\ub_n$, $\Sb_n$, $\bb_n$, $J_n$ and $V_n$ are respectively given by \eqref{un}, \eqref{Sn}, \eqref{bn}, \eqref{Jn} and \eqref{Vn}. Notice that $\MSE\{\hISE_{BLP}(\eta_n)\}< \ms^4(J_n^2+2\,V_n) = \Ex\{\ISE^2(\eta_n)\}$, the MSE of the trivial estimator $\hISE(\eta_n)=0$ --- which is not necessarily the case for $\MSE\{\hISE_{LOO}(\eta_n)\}$, see \eqref{MSE-LOO}.
Direct comparison with \eqref{MSE-LOO} gives
\bea
\MSE\{\hISE_{LOO}(\eta_n)\} - \MSE\{\hISE_{BLP}(\eta_n)\} = \ms^4\, \left(\1b_n/n-\Sb_n^{-1}\bb_n\right)\TT\Sb_n\left(\1b_n/n-\Sb_n^{-1}\bb_n\right)  \geq 0
\eea
and Section~\ref{S:numerical} will highlight the superiority of $\hISE_{BLP}(\eta_n)$ over $\hISE_{LOO}(\eta_n)$ in various situations involving model misspecification; see also Sections~\ref{S:different-regularity} to \ref{S:f-not-GP} in the supplement.

\subsection{BLUP for the assumed GP model}\label{S:ISE-BLP}

As in Section~\ref{S:LOO-etan=etan*}, assume now that $\eta_n\equiv \eta^*_n$ given by \eqref{eta_n} (with again $Y_\xb\sim\GP(0,\ms^2K)$). With the notation of Section~\ref{S:LOO-etan=etan*} we get $\bb_n^* = \int_\SX \cb_n^*(\xb)\,\mu(\dd\xb)=J_n^* \ub_n^*$ with $J_n^*= \int_\SX {\rho_n^*}^2(\xb)  \,\mu(\dd\xb)$, and our linear estimator has the simple form
\be\label{ISE-BLP1}
\hISE_{BLP}(\eta^*_n)={\mveb_{LOO}^{\odot 2}}\TT \widehat\gammab_{BLP}^*  = J_n^*\, {\mveb_{LOO}^{\odot 2}}\TT{\Sb_n^*}^{-1}\ub_n^* \,,
\ee
where $\Sb_n^*$ and $\ub_n^*$ are given by \eqref{Sn^*} and \eqref{un^*}.
Simple algebraic manipulations yield
\bea
\Bias \{\hISE_{BLP}(\eta_n^*) \} &=& \ms^2 J_n^* \left( {\ub_n^*}\TT {\Sb_n^*}^{-1} \ub_n^*-1 \right) = - \frac{\ms^2 J_n^*}{1+{\ub_n^*}\TT \Qb_n^{-1} \ub_n^*} \,,
\eea
where $\Qb_n=2 (\Db_n\Mb\Db_n)^{\odot 2}$, showing that $\hISE_{BLP}(\eta_n^*)$ is negatively biased. As the numerical results presented in Section~\ref{S:numerical} show, this is the case in most situations of interest, and we present a bias-corrected version in Section~\ref{S:ISE_BLUP}.
We also get from \eqref{MSE-BLP}:
\bea
\MSE\{\hISE_{BLP}(\eta_n^*)\} &=& \var\{\ISE(\eta_n^*)\} + \IMSE^2(\eta_n^*) \left( 1- {\ub_n^*}\TT {\Sb_n^*}^{-1} \ub_n^* \right) \\
&=& \var\{\ISE(\eta_n^*)\} + \IMSE^2(\eta_n^*) \,\frac{1}{1+{\ub_n^*}\TT \Qb_n^{-1} \ub_n^*} \,.
\eea

\subsection{Best linear ISE estimation with model misspecification}\label{S:robustness}

In this section, we return to the general framework of Section~\ref{S:ISE-any-predictor}, where $\eta_n(\xb)=\wb\TT(\xb,\Xb_n)\yb_n$ is a given arbitrary linear predictor, but we estimate $\ISE(\eta_n)$ assuming the misspecified model $Y_\xb\sim\GPmodel{e}$ when in fact $Y_\xb\sim\GPmodelg$, $K\neq K^{(e)}$. We thus add the superscript $^{(e)}$ to the notation of Section~\ref{S:ISE-any-predictor}.
We have now
$\widehat\gammab_{BLP}= {\Sb_n^{(e)}}^{-1}{\bb_n^{(e)}}$ and
$\hISE_{BLP}(\eta_n) =  {\mveb_{LOO}^{\odot 2}}\TT{\Sb_n^{(e)}}^{-1} {\bb_n^{(e)}}$.
Substituting $\widehat\gammab_{BLP}$ for $\gammab$ in \eqref{Bias-gamma} and \eqref{MSE-gamma} we obtain
\be
\hspace{-1cm} \Ex\{\hISE_{BLP}(\eta_n)\} &=& \ms^2\, \ub_n\TT{\Sb_n^{(e)}}^{-1} {\bb_n^{(e)}} \,,  \label{E_0} \\
\hspace{-1cm} \MSE\{\hISE_{BLP}(\eta_n)\} &=& \ms^4\left[{\bb_n^{(e)}}\TT{\Sb_n^{(e)}}^{-1} \Sb_n {\Sb_n^{(e)}}^{-1} {\bb_n^{(e)}}  
- 2\, {\bb_n^{(e)}}\TT{\Sb_n^{(e)}}^{-1}\bb_n + J_n^2+2\,V_n \right]\,,  \label{MSE_0}
\ee
where $\ub_n, \Sb_n, \bb_n, J_n$ and $V_n$ are defined in Section~\ref{S:LOOCV-GPmodel} (with the superscript $^{(e)}$ when the kernel $K^{(e)}$ is substituted for $K$).

As one may expect, $\MSE\{\hISE_{BLP}(\eta_n)\}$ is minimum when using the oracle ISE estimator based on the true model $\GPmodelg$. Indeed, denoting $\hISE_{BLP}^{(oracle)}(\eta_n)$ the estimator that uses $K$ instead of $K^{(e)}$, by direct calculation with \eqref{MSE_0} we get
\be
&& \MSE\{\hISE_{BLP}(\eta_n)\} - \MSE\{\hISE_{BLP}^{(oracle)}(\eta_n)\} \nonumber\\
&& \hspace{2cm} =  \ms^4 \left({\Sb_n^{(e)}}^{-1}\bb_n^{(e)}-\Sb_n^{-1}\bb_n\right)\TT\Sb_n \left({\Sb_n^{(e)}}^{-1}\bb_n^{(e)}-\Sb_n^{-1}\bb_n\right) \geq 0 \,. \label{ineq-MSEs}
\ee

\subsection{Independent and flat limits}\label{S:limits}

Here we assume that $K^{(e)}$ is translation invariant, with $K^{(e)}(\xb,\xb')=K_\mt(\xb,\xb')=\Psi[\mt(\xb-\xb')]$,
for some function $\Psi$ defined on $\RR^+$ such that $\Psi(\0b_d)=1$ and $\Psi(\zb)$ tending to zero when $\|\zb\|\to+\infty$. In particular, $K_\mt$ may be isotropic, with $K_\mt(\xb,\xb')=\psi(\mt\|\xb-\xb'\|)$ and $\mt$ acting like the inverse of a correlation length, with $\psi(0)=1$ and $\psi(r)\to 0$ as $r\to+\infty$. All kernels used in the examples of Section~\ref{S:numerical} have this property.
We assume that $\mu(\Xb_n)=0$.
We call independent limit the case $\mt\to+\infty$ and flat limit the case $\mt\to 0$. There is no limiting behaviors to consider for $\hISE_{LOO}(\eta_n)$ as it does not use $K^{(e)}$, and we thus only consider the case of $\hISE_{BLP}(\eta_n)$ defined by \eqref{estimate-ISE-general}.

\paragraph{Independent limit}
For a fixed design $\Xb_n$, as $\mt\to+\infty$ direct calculation gives $\Kb_n^{(e)}\to\Ib_n$, ${\ub_n^{(e)}}\to{\ub_n^{(e)}}(\infty)=
\diag\left\{(\Rb_n\TT\Rb_n)\right\}$,
and we get
\bea
J_n^{(e)} &\underset{\mt\to +\infty}{\to}& J_n^{(e)}(\infty) = 1+ \int_\SX \|\wb_n(\xb)\|^2\, \mu(\dd\xb) \,, \\
\bb_n^{(e)} &\underset{\mt\to +\infty}{\to}& {\bb_n^{(e)}}(\infty) = J_n^{(e)}(\infty)\,{\ub_n^{(e)}}(\infty) +
2\, \diag\left\{(\Rb_n\TT I(\wb)\Rb_n)\right\} \,, \\
\Sb_n^{(e)} &\underset{\mt\to +\infty}{\to}& {\Sb_n^{(e)}}(\infty) = {\ub_n^{(e)}}(\infty){\ub_n^{(e)}}(\infty)\TT +2\,(\Rb_n\TT\Rb_n)^{\odot 2} \,,
\eea
with $I(\wb)=\int_\SX \wb_n(\xb)\wb_n\TT(\xb)\,\mu(\dd\xb)$. Therefore,
\bea
\hISE_{BLP}(\eta_n) \underset{\mt\to +\infty}{\to} {\mveb_{LOO}^{\odot 2}}\TT {\Sb_n^{(e)}}^{-1}(\infty) {\bb_n^{(e)}}(\infty) \ \mbox{ and } \
\Ex\{\hISE_{BLP}(\eta_n)\} \underset{\mt\to +\infty}{\to}  \ms^2\, \ub_n\TT{\Sb_n^{(e)}}(\infty)^{-1} {\bb_n^{(e)}}(\infty)\,,
\eea
and, similarly, the independent limit for $\MSE\{\hISE_{BLP}(\eta_n)\}$ is obtained by substituting ${\Sb_n^{(e)}}(\infty)$ and ${\bb_n^{(e)}}(\infty)$ for $\Sb_n^{(e)}$ and $\bb_n^{(e)}$ in \eqref{MSE_0}.

\paragraph{Flat limit} We let now $\mt$ tend to zero in $K^{(e)}(\xb,\xb')=\Psi[\mt(\xb-\xb')]$.
When we take $K=K^{(e)}$, i.e., when the data are also generated by $\GPmodel{e}$, a careful analysis (which is beyond the scope of this paper) based on \cite{BarthelmeATU2023, BarthelmeU2021} shows the existence of a flat limit for the weights $\widehat\gammab_{BLP}^* = J_n^*\, {\Sb_n^*}^{-1}\ub_n^*$ of the estimator
$\hISE_{BLP}(\eta_n^*)$.
Then, since $\Ex\{\mveb_{LOO}^{\odot 2}\}=\ms^2\ub_n^*$ and $\{\ub_n^*\}_i={\rho_{n\setminus i}^*}^2(\xb_i)$, see Section~\ref{S:LOO-etan=etan*}, for a fixed $\ms_e^2$, $\Ex\{\hISE_{BLP}(\eta_n^*)\}\underset{\mt\to 0}{\to} 0$ (and similarly $\MSE\{\hISE_{BLP}(\eta_n^*)\}\underset{\mt\to 0}{\to} 0$).

As explained below, the situation is different in the (more meaningful) situation where $K\neq K^{(e)}$, $K$ is fixed and $\mt\to 0$ in $K^{(e)}$. Studying the precise behavior of $\hISE_{BLP}(\eta_n)$ when $\mt\to 0$ would require developments beyond the scope of this paper; we nevertheless list some basic facts that explain certain features observed in the examples in Section~\ref{S:numerical}.

As $\mt \to 0$, we have
$\Kb_n^{(e)}\to \1b_n\1b_n\TT$ and $\kb_n^{(e)}(\xb)\to \1b_n$ for all $\xb$. Therefore,
\eqref{rho_n^2-b} gives ${\rho_n^{(e)}}^2(\xb) \underset{\mt\to 0}{\to} [1-\wb_n\TT(\xb) \1b_n]^2$ and thus $J_n^{(e)} \underset{\mt\to 0}{\to} J_n^{(e)}(0) = \int_\SX [1-\wb_n\TT(\xb)\1b_n]^2\, \mu(\dd\xb)$, and \eqref{tb_n} yields $\tb_n^{(e)}(\xb)\underset{\mt\to 0}{\to}[1-\wb_n\TT(\xb) \1b_n] \1b_n$. We also have
\bea
\ub_n^{(e)} = \diag\left\{(\Rb_n\TT\Kb_n^{(e)}\Rb_n)\right\} \underset{\mt\to 0}{\to} \ub_n^{(e)}(0) = (\Rb_n\TT\1b_n)^{\odot 2} \,,
\eea
so that \eqref{cn} gives
$\cb_n^{(e)}(\xb) \underset{\mt\to 0}{\to} 3\,[1-\wb_n\TT(\xb) \1b_n]^2\,\ub_n^{(e)}(0)$ for any $\xb\in\SX$, and therefore
\bea
\bb_n^{(e)} \underset{\mt\to 0}{\to} \bb_n^{(e)}(0) = 3\, J_n^{(e)}(0) \ub_n^{(e)}(0) \,. 
\eea
As $\Rb_n\TT\Kb_n^{(e)}\Rb_n \underset{\mt\to 0}{\to} \Rb_n\TT\1b_n \1b_n\TT\Rb_n$, $\Sb_n^{(e)}\underset{\mt\to 0}{\to} \Sb_n^{(e)}(0) = 3\, \ub_n^{(e)}(0)[\ub_n^{(e)}(0)]\TT$, a rank-one matrix.
The singularity of $\Sb_n^{(e)}(0)$ prevents the existence of a flat-limit for $\hISE_{BLP}(\eta_n)$ when $\mt$ tends to zero, explaining why we encounter numerical difficulties for evaluating $\hISE_{BLP}(\eta_n)$ for (very) small $\mt$.

When the predictor $\eta_n$ is such that $\wb_n\TT(\cdot)\1b_n \equiv 1$ for any $n$ and any $\Xb_n$, we have $J_n^{(e)}(0)=0$ and thus $\bb_n^{(e)}(0)=\0b_n$, and moreover the matrix $\Rb_n$ in \eqref{Rn} satisfies $\Rb_n\TT\1b_n=\0b_n$ and thus $\ub_n^{(e)}(0)=\0b_n$. Therefore, $\Sb_n^{(e)}$ and $\bb_n^{(e)}$ respectively tend to the null matrix and null vector when $\mt_{\rm BLP}\to 0$, and we may expect the numerical difficulties to be less pronounced for predictors with this property; see the numerical example in Section~\ref{S:BLP-other-K} for an illustration.

\subsection{Best linear unbiased estimation of the ISE}\label{S:ISE_BLUP}
Assuming that $Y_\xb\sim\GPmodelg$, we can easily correct the bias of the linear estimator of $\mve_n^2(\xb)$ derived in Section~\ref{S:ISE-any-predictor}: we minimize $\Ex\{[\mve_n^2(\xb)-\betab\TT\mveb_{LOO}^{\odot 2}]^2\}$ with respect to $\betab\in\RR^n$ under the constraint $\betab\TT \Ex\{\mveb_{LOO}^{\odot 2}\}=\Ex\{\mve_n^2(\xb)\}$. Since $\Ex\{\mve_n^2(\xb)\}=\ms^2 \rho_n^2(\xb)$ and $\Ex\{\mveb_{LOO}^{\odot 2}\}=\ms^2\ub_n$, see \eqref{rho_n^2-b} and \eqref{un}, the constraint is $\betab\TT\ub_n=\rho_n^2(\xb)$, which does not depend on $\ms^2$. The optimal solution to this convex minimization problem is
\be
\widehat\betab_U(\xb) = \Sb_n^{-1} \left[\cb_n(\xb) + \frac{\rho_n^2(\xb)-\ub_n\TT\Sb_n^{-1}\cb_n(\xb)}{\ub_n\TT\Sb_n^{-1}\ub_n}\,\ub_n\right] \,. \label{beta_BLUP}
\ee
The unbiased version of the linear estimate of $\ISE(\eta_n)$ is then
$\hISE_{BLUP}(\eta_n) = {\widehat\gammab_{BLUP}}\TT \mveb_{LOO}^{\odot 2}$,
with $\widehat\gammab_{BLUP}=\int_\SX \widehat\betab_U(\xb)\,\mu(\dd\xb)=\Sb_n^{-1}\bb_n+(J_n-\ub_n\TT\Sb_n^{-1}\bb_n)\Sb_n^{-1}\ub_n/(\ub_n\TT\Sb_n^{-1}\ub_n)$ (which we also directly obtain by minimization of $\MSE\{\hISE(\eta_n)\}$ in \eqref{MSE-gamma} under the constraint $\gammab\TT\ub_n=J_n$, see \eqref{Bias-gamma} and \eqref{IMSE_arbitrary}). Its bias, variance and MSE are respectively given by \eqref{Bias-gamma}, \eqref{var-gamma} and \eqref{MSE-gamma} with $\widehat\gammab_{BLUP}$ substituted for $\gammab$. Note that $\hISE_{BLUP}(\eta_n)$ is unbiased only in absence of model misspecification. Cases where $\eta_n\equiv \eta^*_n$ and where a misspecified kernel $K^{(e)}\neq K$ is assumed can be considered similarly to Sections~\ref{S:ISE-BLP} and \ref{S:robustness}. A numerical illustration of the performance of $\hISE_{BLUP}(\eta_n)$ is given in Sections~\ref{S:BLP-other-K} (see Figure~\ref{F:ISE=M1-M52-theta5_d2_n100_M0-M32-theta10_M2-M32-unbiased}) and \ref{S:behavior-d-n}, \ref{S:f-not-GP} in the supplement, indicating a moderate improvement over $\hISE_{BLP}(\eta_n)$ (in particular because bias cancellation is only effective in the absence of modeling error).

\subsection{Introduction of a nugget effect for noisy observations}\label{S:noisy-observations}

Statistical modeling of physical systems usually relies on observations corrupted by noise. Suppose that we observe $y(\xb_i)=f(\xb_i)+\zeta_i$ at the $n$ design points $\xb_i$, where the measurement errors $\zeta_i$ are i.i.d.\ random variables. Assuming that these errors are normal, the developments above for the construction of the ISE estimate $\hISE_{BLP}(\eta_n)$ remain valid provided we use now a GP model with nugget effect: we assume that $Y_\xb \sim \GP(0,\ms^2 K'_r)$ with $K'_r$ defined by $K'_r(\xb,\xb')=K(\xb,\xb')+r\,\delta_{\xb,\xb'}$, where $\delta_{\xb,\xb'}=1$ when $\xb=\xb'$ and is zero otherwise. The implementation of the method requires knowledge of the nugget effect $r$. Estimating $r$ from the data $\SSF_n$ is a possible option (see, e.g., \cite[Sect.~4]{KarvonenO2023}) that we do not pursue here: it raises several issues, notably of robustness (note that both $\ms^2$ and $r$ need to be estimated), which are worth investigating further. However, the numerical results presented in Section~\ref{S:noisy-observations-supp} in the supplement indicate that the performance of $\hISE_{BLP}(\eta_n)$ remains noticeably superior to that of LOOCV even when $r$ is severely misspecified.

\section{Numerical experiments}\label{S:numerical}
\subsection{Influence of the design $\Xb_n$}\label{S:design-influence}

This simple example illustrates the discussion at the end of Section~\ref{S:LOOCV-GPmodel}. Here, the function $f$ depends on a single variable $x\in\SX=[0,1]$ and is a realization of a GP: $Y_x\sim\GPmodelg$ with $\ms=1$ and $K(x,x')=K_{3/2,\mt_0}(\xb,\xb')=\psi_{3/2,\mt_0}(|x-x'|)$, where $\psi_{3/2,\mt}$ corresponds to the isotropic Mat\'ern 3/2 kernel,
\be\label{Matern32}
\psi_{3/2,\mt}(r) = (1+\sqrt{3}\,\mt\, r) \exp(-\sqrt{3}\,\mt\, r) \,.
\ee
The predictor $\eta_n$ is the BLUP for the model $\GPmodel{p}$, with $K^{(p)}(x,x')=K_{5/2,\mt_p}(\xb,\xb')=\psi_{5/2,\mt_p}(|x-x'|)$ corresponding to the isotropic Mat\'ern 5/2 kernel
\be\label{Matern52}
\psi_{5/2,\mt}(r) &=&  \left[1+\sqrt{5}\,\mt\, r + (5/3)\, \mt^2\, r^2\right] \exp(-\sqrt{5}\,\mt\, r) \,.
\ee
We take $\mt_0=5$ and $\mt_p=2$ (the predictor thus assumes extra regularity and smoothness). We compare the estimators $\hISE_{LOO}(\eta_n)$ and $\hISE_{BLP}(\eta_n)$ for a particular realization of $Y_x$ and for a family of $n$-point designs $\Xb_n(\delta)$, with $n=10$, ranging from designs composed of 5 pairs of neighboring points to designs well spread over $\SX$:
$\Xb_n(\delta)=\{0,0.2,0.4,0.6,0.8\}\cup\{\delta,0.2+\delta,0.4+\delta,0.6+\delta,0.8+\delta\}$, $\delta\in[0.005,0.1]$ (so that $\delta=\min_{i\neq j}|x_i-x_j|$).
The estimator $\hISE_{BLP}(\eta_n)$ uses the true model $\GPmodelg$.

\begin{figure}[ht]
\centering
\includegraphics[width=0.49\textwidth]{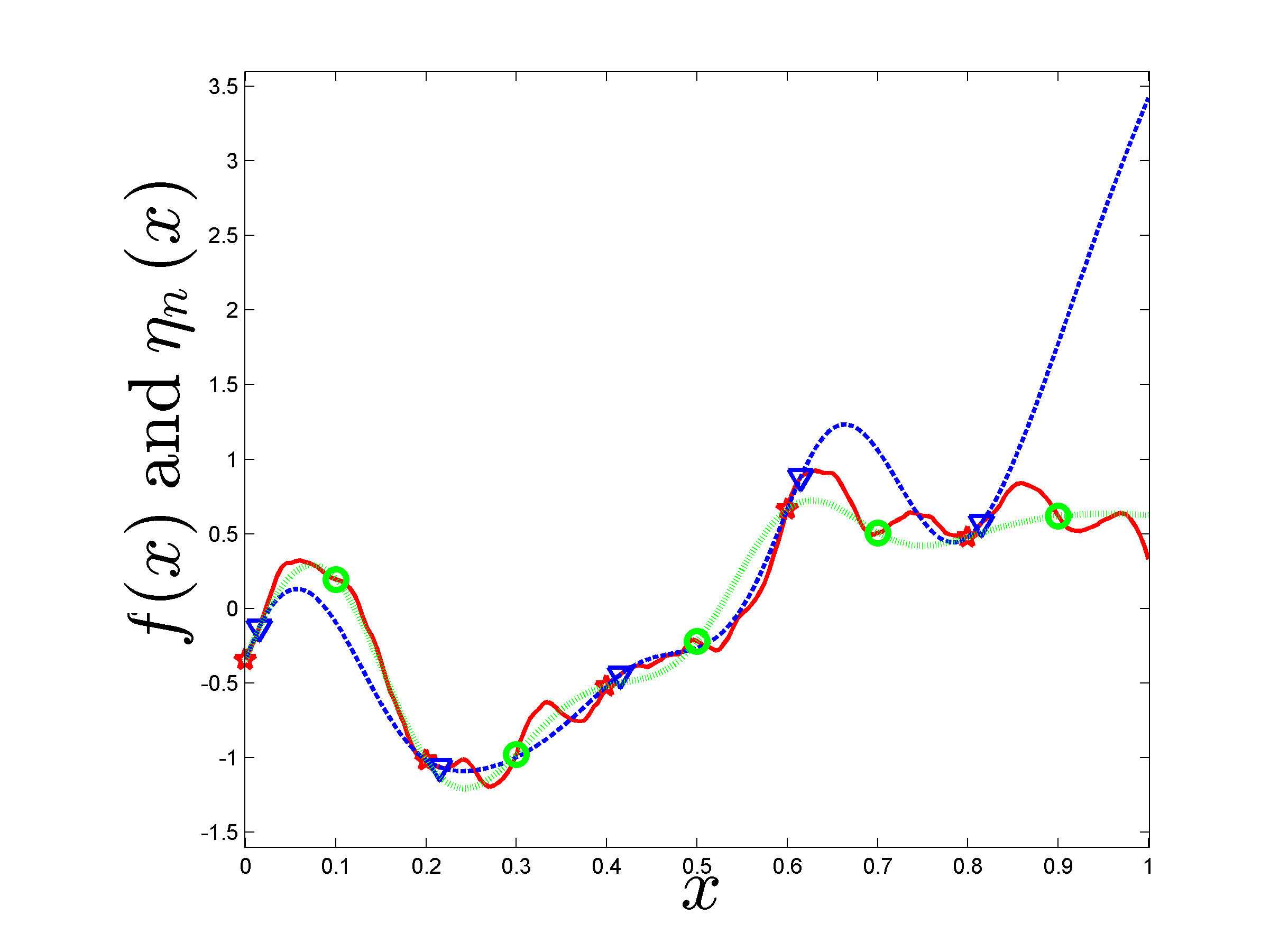}
\includegraphics[width=0.49\textwidth]{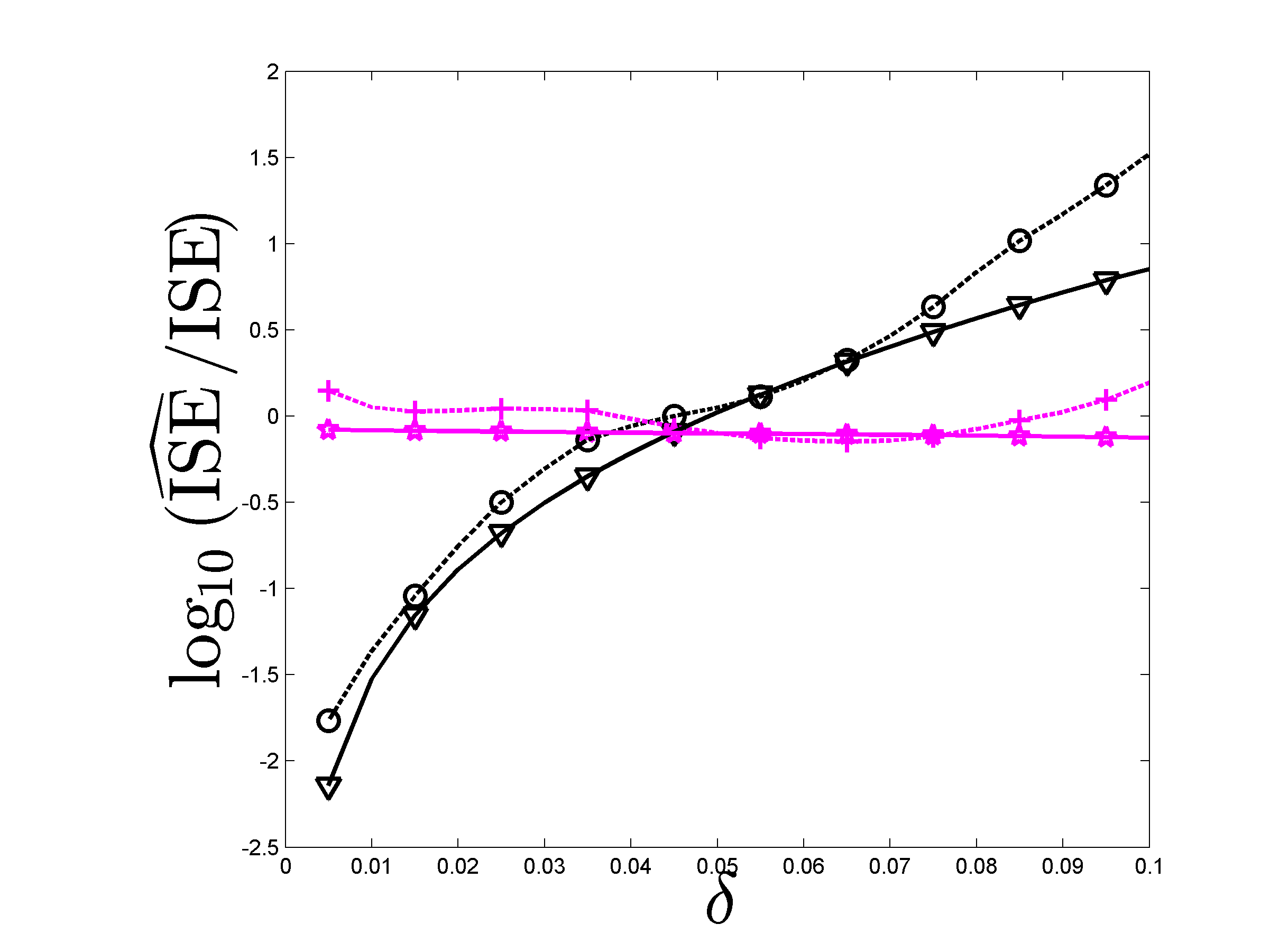}
\caption{\small Left: $f(x)$ ({\color{red} ---}) and $\eta_n(x)$ for the designs $\Xb_n(0.015)$ ({\color{blue} $\cdots$} with {\color{red} $\star$} and
{\color{blue} $\triangledown$}) and $\Xb_n(0.1)$ ({\color{green} $\cdots$} with {\color{red} $\star$} and {\color{green} $\circ$}). Right: $\log_{10}[\hISE_{LOO}(\eta_n)/\ISE(\eta_n)]$ (- - - with $\circ$) and
$\log_{10}[\hISE_{BLP}(\eta_n)/\ISE(\eta_n)]$ ({\color{magenta} - - -} with {\color{magenta} $+$});
$\log_{10}[\Ex\{\hISE_{LOO}(\eta_n)\}/\IMSE(\eta_n)]$ (--- with $\triangledown$) and
$\log_{10}[\Ex\{\hISE_{BLP}(\eta_n)\}/\IMSE(\eta_n)]$ ({\color{magenta} ---} with {\color{magenta} $\star$}) as functions of $\delta$.
}
\label{F:design-influence}
\end{figure}

The left panel of Figure~\ref{F:design-influence} shows the realization of $Y_x$ defining $f(x)$ (red solid line) and two predictions $\eta_n(x)$ corresponding to $\Xb_n(0.015)$ (triangles and dotted line in blue) and $\Xb_n(0.1)$ (circles and dotted line in green); the design points $\{0,0.2,0.4,0.6,0.8\}$ (present in $\Xb_n(0.015)$ and $\Xb_n(0.1)$) are indicated by red stars. The right panel shows the evolution of the ratios $\hISE_{LOO}(\eta_n)/\ISE(\eta_n)$ (black dotted line with $\circ$) and
$\hISE_{BLP}(\eta_n)/\ISE(\eta_n)$ (magenta dotted line with $+$), in log scale, as functions of $\delta$, for the particular realization of the left panel. The solid line curves, black with $\triangledown$ and magenta with $\star$, are respectively for $\Ex\{\hISE_{LOO}(\eta_n)\}/\IMSE(\eta_n)$ and $\Ex\{\hISE_{BLP}(\eta_n)\}/\IMSE(\eta_n)$.

When $\delta$ is small, prediction at a removed point $x_i$ is accurate due to the presence of another design point nearby, and the LOO error $\mve_{-i}$ is significantly smaller than a typical error $\mve_n(x)$ for $x\in\SX$. As a consequence, $\hISE_{LOO}(\eta_n)$ strongly underestimates $\ISE(\eta_n)$. Conversely, for designs corresponding to large $\delta$, removing one $x_i$ leaves a big hole in $\Xb_n$ and prediction at this $x_i$ is inaccurate: $\mve_{-i}$ is thus significantly larger than a typical $\mve_n(x)$ and  $\hISE_{LOO}(\eta_n)$ overestimates $\ISE(\eta_n)$. On the opposite, $\hISE_{BLP}(\eta_n)$ gives an acceptable estimation of $\ISE(\eta_n)$ for all values of $\delta$ considered.

\begin{figure}[ht]
\centering
\includegraphics[width=0.5\textwidth]{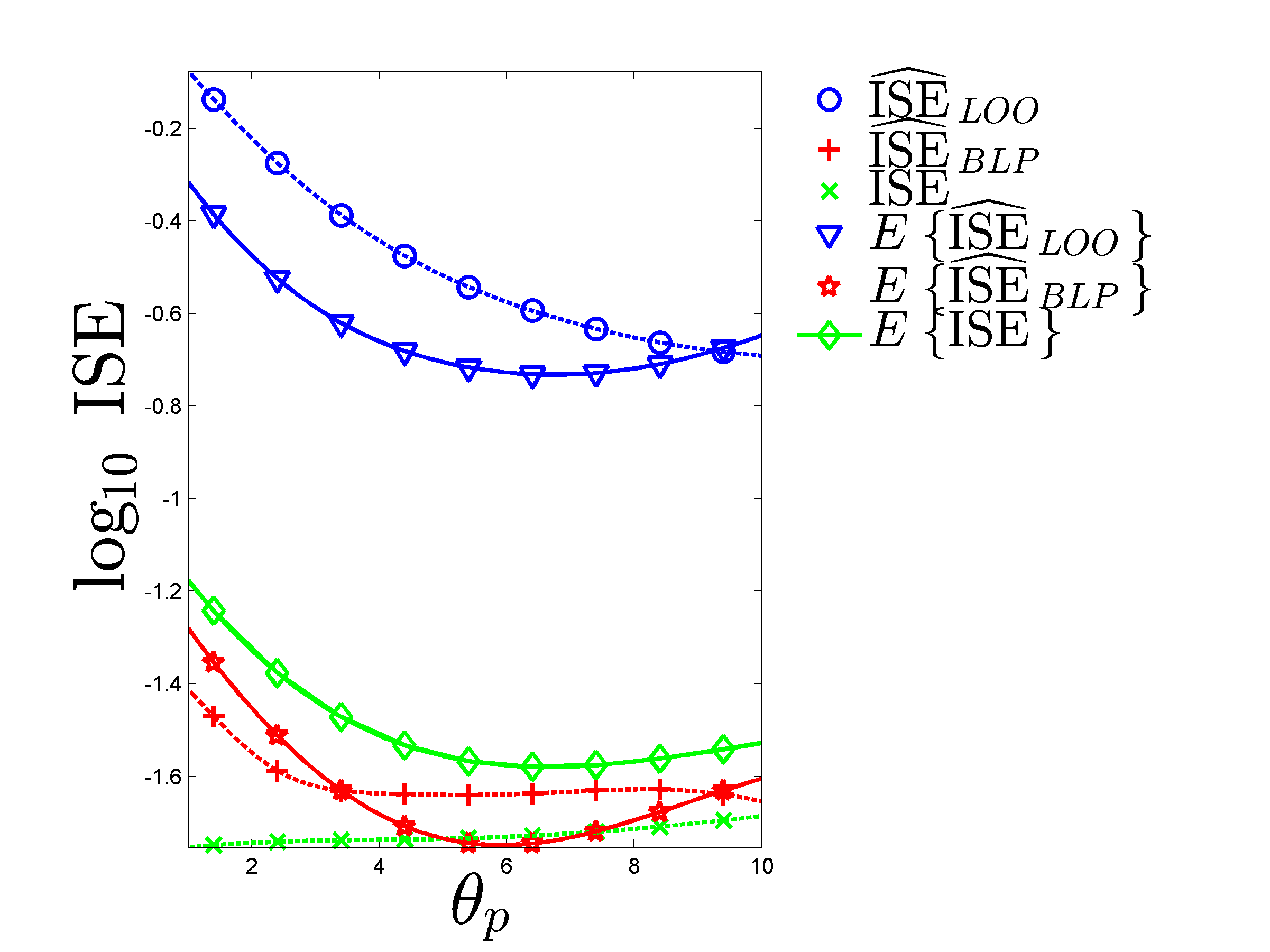}

\caption{\small $\log_{10}[\hISE_{LOO}(\eta_n)]$, $\log_{10}[\hISE_{BLP}(\eta_n)]$ and $\log_{10}[\ISE(\eta_n)]$ for the particular realization on the left panel of Figure~\ref{F:design-influence}, and $\log_{10}[\Ex\{\hISE_{LOO}(\eta_n)\}]$, $\log_{10}[\Ex\{\hISE_{BLP}(\eta_n)\}]$ and $\log_{10}[\Ex\{\ISE(\eta_n)\}]$, as functions of $\mt_p\in[1,10]$.
}
\label{F:predictor}
\end{figure}

Figure~\ref{F:predictor} is for the design $\Xb_n(0.1)=\{0,0.1,0.2,\ldots,0.9\}$ (the best among all $\Xb_n(\delta)$ in terms of $\IMSE(\eta_n)$) and shows (in log scale) $\hISE_{LOO}(\eta_n)$, $\hISE_{BLP}(\eta_n)$ and $\ISE(\eta_n)$ for the realization on the left panel of Figure~\ref{F:design-influence}, together with their expected values, when $\mt_p$, the range parameter in the kernel $K_{3/2,\mt_p}$ of the predictor, varies in $[1,10]$.
Estimation of $\ISE(\eta_n)$ by $\hISE_{BLP}(\eta_n)$ is much more accurate than with $\hISE_{LOO}(\eta_n)$ for all predictors considered. Note that the small negative bias of
$\hISE_{BLP}(\eta_n)$ is not very sensitive to the smoothness of $\eta_n$ for this example. Also note that $\hISE_{LOO}(\eta_n)$ and $\hISE_{BLP}(\eta_n)$ are both minimum for $\mt_p=10$ whereas $\ISE(\eta_n)$, the true ISE, is minimum for $\mt_p=1$ (however, $\Ex\{\ISE(\eta_n)\}$, $\Ex\{\hISE_{LOO}(\eta_n)\}$ and $\Ex\{\hISE_{BLP}(\eta_n)\}$ are respectively minimum for $\mt_p \simeq 6.6$, 6.7 and 5.9).

\subsection{Robustness to the choice of $K^{(e)}$}\label{S:numerical-robustness}

Here we consider numerical examples of construction of $\hISE_{BLP}(\eta_n)$ involving different predictors $\eta_n$, different data generating models $\GPmodelg$ and different assumed models $\GPmodel{e}$  with $K^{(e)}\neq K$, and compare the performances of $\hISE_{BLP}(\eta_n)$ and $\hISE_{LOO}(\eta_n)$. We always use isotropic kernels. In particular, $K^{(e)}(\xb,\xb')=\psi^{(e)}(\mt_{\rm BLP}\|\xb-\xb'\|)$ and we study the influence of the choice of the range parameter $\mt_{\rm BLP}$, the inverse of a correlation length. The notation $\mt_{\rm BLP}$ is to highlight the fact that $\mt_{\rm BLP}$ only influences the estimation of $\ISE(\eta_n)$ by $\hISE_{BLP}(\eta_n)$.

We take $\SX=[0,1]^2$ and $\Xb_n$ ($n=100$) is the $10\times 10$ regular grid with coordinates $(i-1)/9$, $i=1,\ldots,10$; $\mu$ is the empirical measure on the first $2^{10}$ points of the low-discrepancy Sobol' sequence in $\SX$ (which means that we consider a Quasi-Monte Carlo approximation of the ISE for the uniform measure on $\SX$). We generate $n$ observations $\yb_n$ for the design $\Xb_n$ and the model $\GPmodelg$, with $\ms^2=1$ (its value is irrelevant as it simple acts as a scaling factor) and $K(\xb,\xb')=K_{3/2,\mt_0}(\xb,\xb')=\psi_{3/2,\mt_0}(\|\xb-\xb'\|)$, the Mat\'ern 3/2 kernel given by \eqref{Matern32}, where we set $\mt_0=10$. With $\mt_0$ fixed, the expected ISE for this model, $\Ex\{\ISE(\eta_n)\}=\IMSE(\eta_n)$, only depends on $\Xb_n$, see \eqref{IMSE_arbitrary}.

In the first example below, $\eta_n$ corresponds to a polynomial model that is not an interpolator.

\subsubsection{Prediction with a non-interpolating polynomial model}\label{S:polynomial}

The predictor $\eta_n$ is obtained by polynomial model fitting: we (wrongly) assume that the data $\yb_n$ are given by
\bea 
y_i=\phib\TT(\xb_i)\alphab+\delta_i \,,
\eea
where the error vector $\mdb=(\delta_1,\ldots,\delta_n)\TT$ is normally distributed $\SN(0,\mg^2\Ib_n)$ and where each component $\phi_\ell(\xb)$ of $\phib(\xb)=[\phi_1(\xb),\ldots,\phi_m(\xb)]\TT$ is a multivariate polynomial in the $d$ components of $\xb$; see Section~\ref{S:Appendix-polynomial} in the supplement. The predictor is $\eta_n(\xb)= \phib\TT(\xb)\widehat\alphab$, with $\widehat\alphab$ the posterior mean of the model parameters and $\phib(\xb)$ a vector of polynomial functions of $\xb$. As this model (wrongly) assumes the presence of i.i.d.\ observation errors with positive variance $\mg^2$, $\eta_n$ is not an interpolator.

We take $\mg^2=0.1$; $\phib(\xb)$ has dimension $m=n/2=50$ and each of its components has the form $\varphi_{\ell_1}(x_1)\varphi_{\ell_2}(x_2)$, where $\varphi_{i}(\cdot)$ denotes the Legendre polynomial of degree $i$, orthonormal for the uniform measure on $[0,1]$. The indices $\ell_1$ and $\ell_2$ take the values
\bea
\left\{\left[\!\!\!
    \begin{array}{c}
      \ell_1 \\
      \ell_2 \\
    \end{array}
  \!\!\!\right],\, \ell=1,\ldots,50\right\}
= \left\{\begin{array}{l}0     0     1     1     0     2     1     2     0     3     2     1     3     0     4     2     3     1 4     0     5     3     2     4     1     5     0     6     2     5     3     4     1     6     0     7 3     5     2     6     4     1     7     0     8     4     5     3     6     2 \\
0     1     0     1     2     0     2     1     3     0     2     3     1     4     0     3     2     4 1     5     0     3     4     2     5     1     6     0     5     2     4     3     6     1     7     0
5     3     6     2     4     7     1     8     0     5     4     6     3     7
\end{array}\right\} \,,
\eea
and the polynomial model is thus of total degree 9. We set a vague prior on the parameters $\alphab$, assuming that $\alphab\sim\SN(\0b_m,\mLb)$ with $\mLb=\diag\{\mL_1,\ldots,\mL_m\}$ where $\mL_\ell= \ml_{\ell_1}\ml_{\ell_2}$, $\ell=1,\ldots,50$, with $\ell_1,\ell_2$ as indicated above and $\ml_k=10^3\times 2^{-k}$, $k\geq 0$ (the prior on $\alphab$ is therefore not very informative).

We construct $\hISE_{BLP}(\eta_n)$ using the model $\GPmodel{e}$, where $K^{(e)}=K_{3/2,\mt_{\rm BLP}}$ ($K^{(e)}$ thus coincides with $K$ for $\mt_{\rm BLP}= \mt_0 = 10$). For each value of $\mt_{\rm BLP}$ we calculate $\Ex\{\hISE_{BLP}(\eta_n)\}$ and $\MSE\{\hISE_{BLP}(\eta_n)\}$, respectively given by \eqref{E_0} and \eqref{MSE_0}.

The procedure is repeated $M=100$ times, with a different vector $\yb_n[i]$ and thus a different predictor $\eta_n[i]$ each time, and we compute the empirical means $\widetilde\Ex\{\ISE(\eta_n)\}$ and $\widetilde\Ex\{\hISE_{BLP}(\eta_n)\}$ of the $\ISE(\eta_n[i])$ and $\hISE_{BLP}(\eta_n[i])$, respectively, $i=1,\ldots,M$, together with the empirical standard deviations. We also compute $\widetilde\MSE\{\hISE_{BLP}(\eta_n)\}$, given by the empirical mean of the squared errors $[\ISE(\eta_n[i])-\hISE_{BLP}(\eta_n[i])]^2$, and use their empirical standard deviation to build confidence intervals.

\begin{figure}[ht]
\centering
\includegraphics[width=0.49\textwidth]{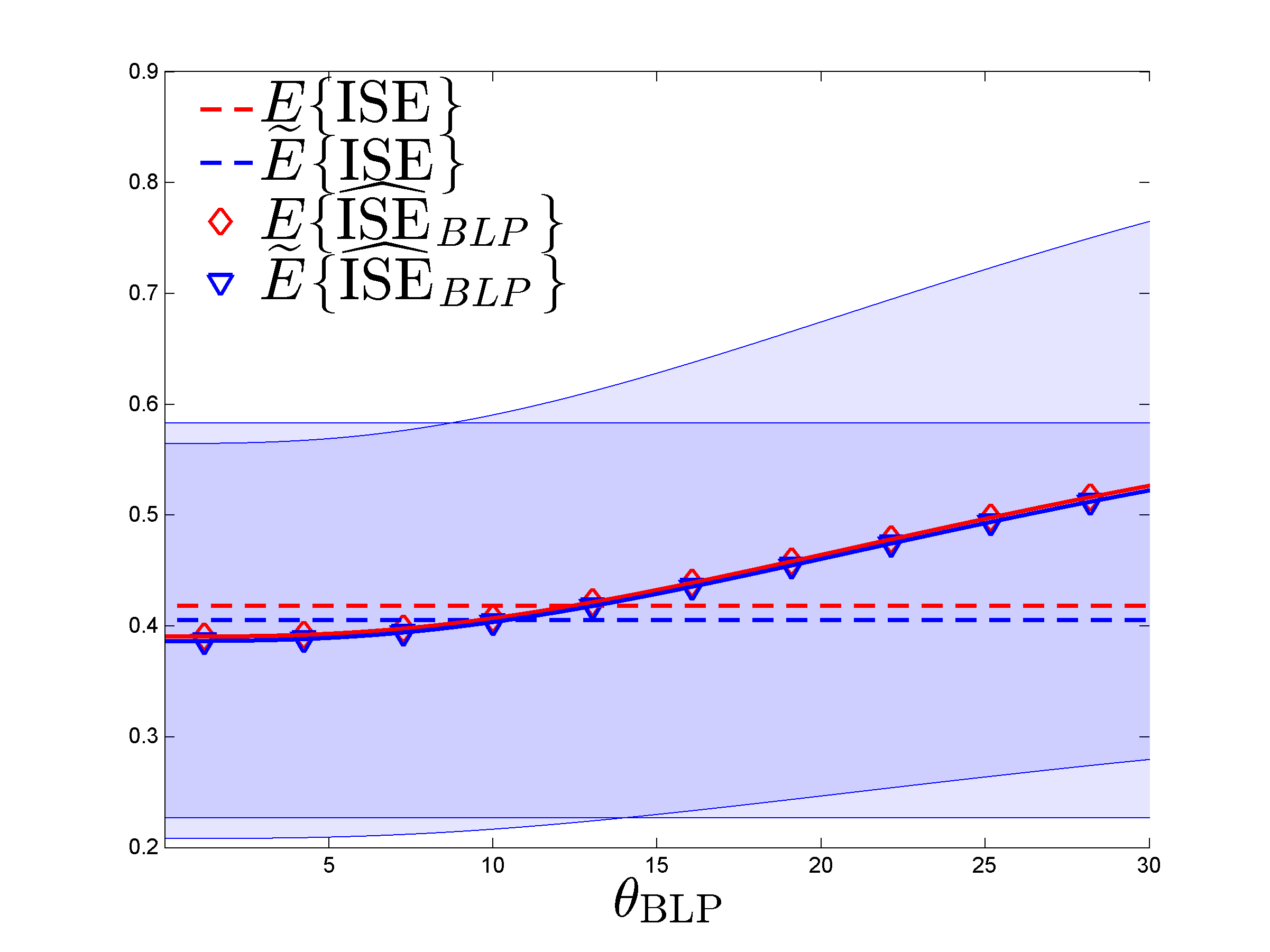}
\includegraphics[width=0.49\textwidth]{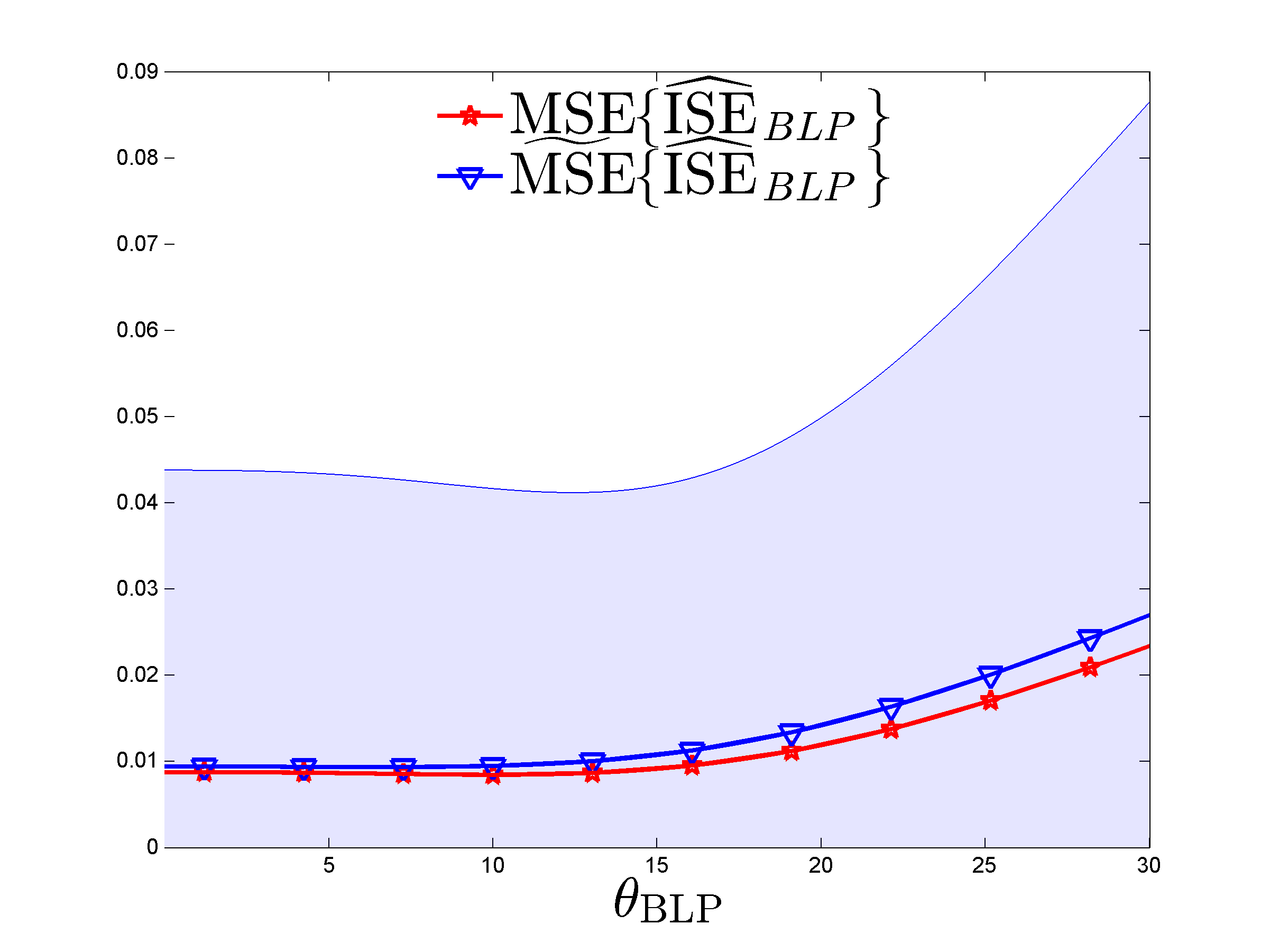}
\caption{\small Estimation of $\ISE(\eta_n)$ by $\hISE_{BLP}(\eta_n)$ when $\eta_n$ is a (non-interpolating) polynomial of total degree 9 with 100 design points forming a regular grid in $[0,1]^2$; $Y_\xb\sim\GP(0,K_{3/2,10})$, $K^{(e)}=K_{3/2,\mt_{\rm BLP}}$, $\mt_{\rm BLP}\in[0.001,30]$.} \label{F:ISE=M1-poly_d2_n100_m50_M0-M32-theta10_M2-M32}
\end{figure}

The left panel of Figure~\ref{F:ISE=M1-poly_d2_n100_m50_M0-M32-theta10_M2-M32} shows $\Ex\{\ISE(\eta_n)\}$, $\Ex\{\hISE_{BLP}(\eta_n)\}$, $\widetilde\Ex\{\ISE(\eta_n)\}$ and $\widetilde\Ex\{\hISE_{BLP}(\eta_n)\}$ as functions of $\mt_{\rm BLP}$; the confidence bands on $\widetilde\Ex\{\ISE(\eta_n)\}$ and $\widetilde\Ex\{\hISE_{BLP}(\eta_n)\}$ (two standard deviations) are colored. The right panel shows $\MSE\{\hISE_{BLP}(\eta_n)\}$ and $\widetilde\MSE\{\hISE_{BLP}(\eta_n)\}$ as functions of $\mt_{\rm BLP}$, and the confidence band on $\widetilde\MSE\{\hISE_{BLP}(\eta_n)\}$ (two standard deviations, truncated to positive values) is colored.
Notice the good agreement between empirical and exact values for the mean and MSE of $\hISE_{BLP}(\eta_n)$.
Although not clearly visible on the plot, $\MSE\{\hISE_{BLP}(\eta_n)\}$ is minimum for $\mt_{\rm BLP}=\mt_0=10$, in agreement with \eqref{ineq-MSEs}.

On the same data set, the estimator $\hISE_{LOO}(\eta_n)$ has an empirical mean and standard deviation of approximately 3.6 and 2.5, respectively; its (exact) expected value is about 3.373.
These values are well outside the range shown on Figure~\ref{F:ISE=M1-poly_d2_n100_m50_M0-M32-theta10_M2-M32}-left.
Conversely, $\hISE_{BLP}(\eta_n)$ has a small negative bias for $\mt_{\rm BLP}$ around $\mt_0$ or smaller, and its positive bias becomes significant only when $\mt_{\rm BLP}$ is much larger than $\mt_0$.

Figure~\ref{F:ISE=M1-poly_d2_n100_m50_M0-M32-theta10_M2-M32} shows that $\Ex\{\hISE_{BLP}(\eta_n)\}$ and $\MSE\{\hISE_{BLP}(\eta_n)\}$ are increasing functions of $\mt_{\rm BLP}$ for $\mt_{\rm BLP}$ large enough; the independent limits ($\mt_{\rm BLP} \to +\infty$), obtained from the calculations of Section~\ref{S:limits}, are given in Table~\ref{Tb:indep-lims-ex2} and are in good agreement with the values obtained numerically for large $\mt_{\rm BLP}$ ($\mt_{\rm BLP} > 100$, say). These values indicate that $\hISE_{BLP}(\eta_n)$ gives a much more precise estimation of $\ISE(\eta_n)$ than $\hISE_{LOO}(\eta_n)$ for all $\mt_{\rm BLP} \gtrsim 10^{-3}$. Note that $\Ex\{\ISE^2(\eta_n)\}$ (first column of Table~\ref{Tb:indep-lims-ex2}) is the MSE of the trivial estimator $\hISE(\eta_n)=0$ and is much smaller than $\MSE\{\hISE_{LOO}(\eta_n)\}$ (second column).

\begin{table}[ht]
\caption{\small $\Ex\{\ISE(\eta_n)\}$ and $\Ex\{\ISE^2(\eta_n)\}$ (first column); $\Ex\{\hISE_{LOO}(\eta_n)\}$ and $\MSE\{\hISE_{LOO}(\eta_n)\}$ (second column); independent limits ($\mt_{\rm BLP}\to\infty$) for $\Ex\{\hISE_{BLP}(\eta_n)\}$ and $\MSE\{\hISE_{BLP}(\eta_n)\}$ (third column). The independent limits are identical for all choices of $K^{(e)}$ considered in the examples of Sections~\ref{S:BLP-other-K} and \ref{S:different-regularity} in the supplement.}
\begin{center}
{\small
				\begin{tabular}{|l|l| c|c|c|}
					\hline	
                  & & $\ISE(\eta_n)$ & $\hISE_{LOO}(\eta_n)$ & $\hISE_{BLP}(\eta_n)$ ($\mt_{\rm BLP}\to\infty$) \\	\hline	
           Ex.~of Section~\ref{S:polynomial} & $\Ex$  & 0.418 & 3.373 &  0.672 \\ \hline
                          & $\MSE$ & 0.181 & 12.785 & 0.082  \\ \hline
           Ex.~of Sections~\ref{S:BLP-other-K} and \ref{S:different-regularity} & $\Ex$ & 0.187   & 0.731 & 0.478 \\ \hline
                                      & $\MSE$& 0.035   & 0.338 & 0.103 \\ \hline
\end{tabular}
}
\end{center}
\label{Tb:indep-lims-ex2}
\end{table}

As the prior on the model parameters is rather vague, the construction almost coincides with Least-Squares regression. Since the linear model contains an intercept ($\phi_1(\xb)=1$ for all $\xb$), the prediction $\wb_n(\xb)\TT\1b_n$ associated with $\yb_n=\1b_n$ is almost one for all $\xb$: we have $|\wb_n(\xb)\TT\1b_n-1|<5\cdot 10^{-8}$ over $\SX$. Hence, in agreement with the flat-limit discussion in Section~\ref{S:limits}, small values of $\mt_{\rm BLP}$ do not cause severe numerical difficulties, and in Figure~\ref{F:ISE=M1-poly_d2_n100_m50_M0-M32-theta10_M2-M32} we could use values as small as $\mt_{\rm BLP}= 10^{-3}$. When the prior on $\alphab$ is more informative, the range for $\mt_{\rm BLP}$ should be restricted to larger values: for example, when $\ml_k=50\times 2^{-k}$, $\Ex\{\hISE_{BLP}(\eta_n)\}$ and $\MSE\{\hISE_{BLP}(\eta_n)\}$ behave qualitatively like in Figure~\ref{F:ISE=M1-poly_d2_n100_m50_M0-M32-theta10_M2-M32} when $\mt_{\rm BLP} \gtrsim 0.015$, but numerical instability appears for smaller $\mt_{\rm BLP}$.

\subsubsection{Linear prediction with a GP model}\label{S:BLP-other-K}

The predictor $\eta_n$ is now the BLUP for the GP model $\GPmodel{p}$, $\eta_n(\xb)={\kb_n^{(p)}}\TT(\xb){\Kb_n^{(p)}}^{-1}\yb_n$, where $K^{(p)}(\xb,\xb')=K_{5/2,\mt_p}(\xb,\xb')=\psi_{5/2,\mt_p}(\|\xb-\xb'\|)$, see \eqref{Matern52}, with $\mt_p=5$;
$\Xb_n$, $\SX$ and $\mu$ are like in Section~\ref{S:polynomial}
and the data are still generated with $\GPmodelg$ where $\ms^2=1$ and
$K=K_{3/2,\mt_0}$ with $\mt_0=10$, see \eqref{Matern32}.

\begin{figure}[ht]
\centering
\includegraphics[width=0.49\textwidth]{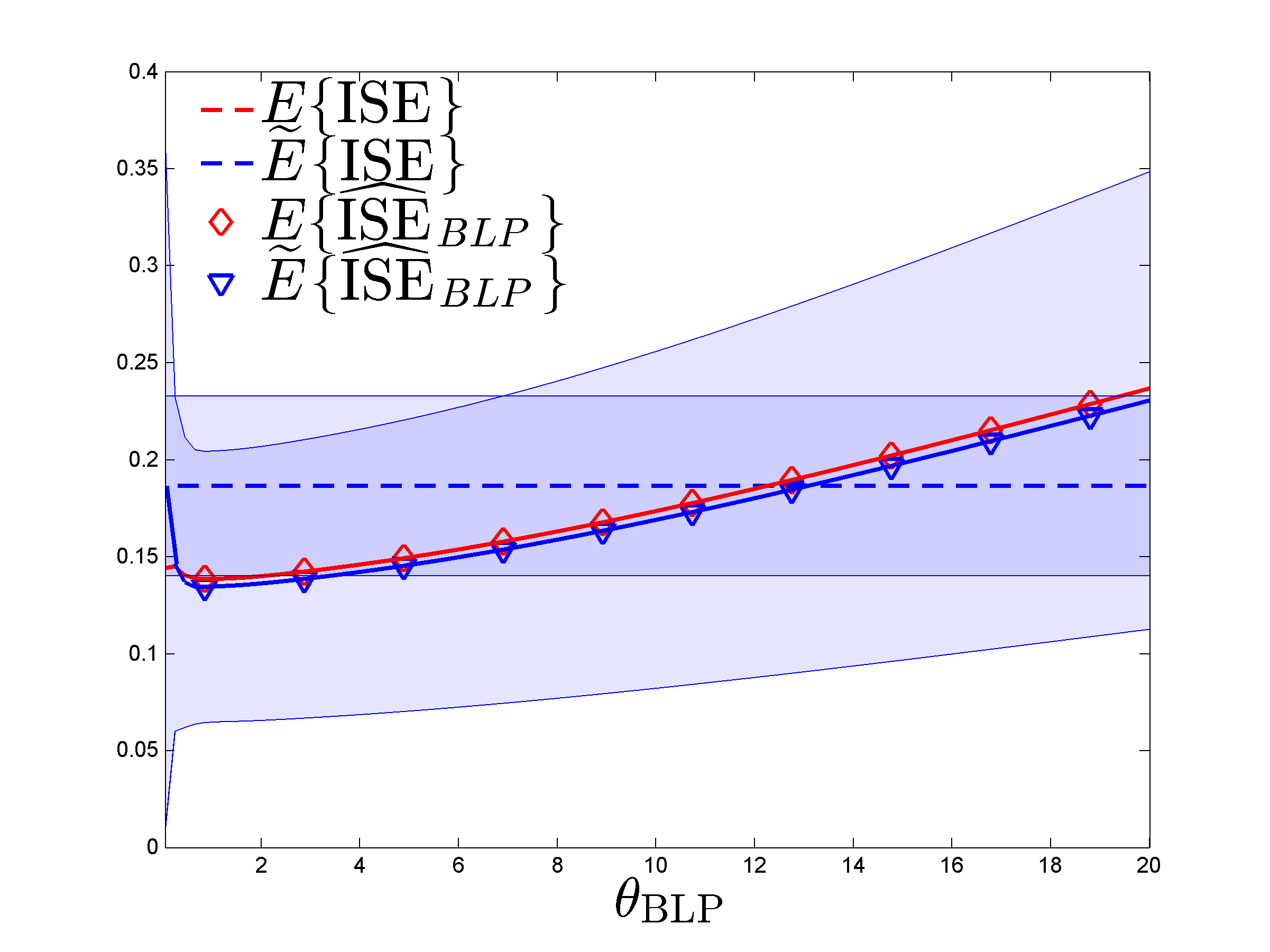}
\includegraphics[width=0.49\textwidth]{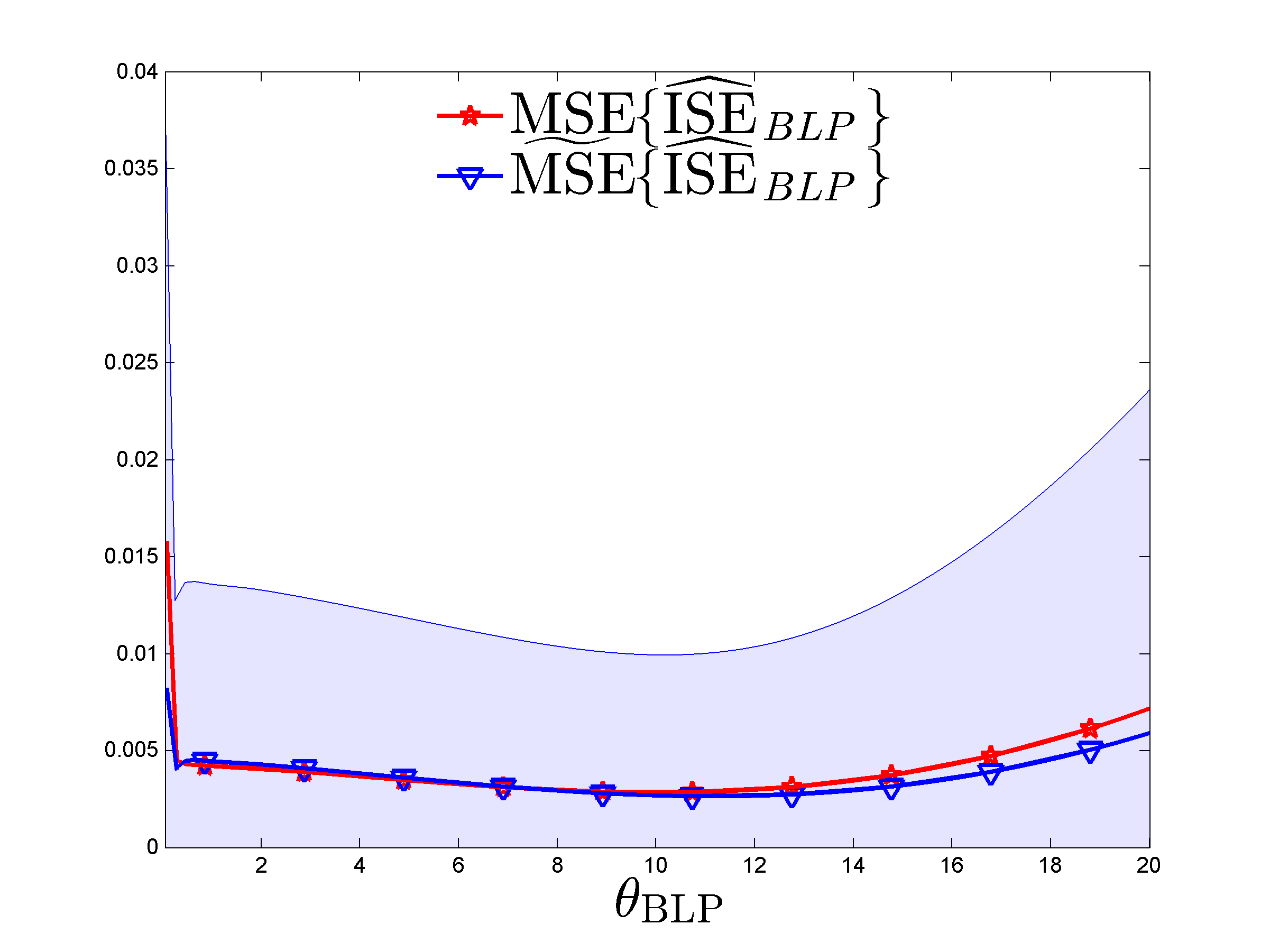}
\caption{\small Estimation of $\ISE(\eta_n)$ by $\hISE_{BLP}(\eta_n)$ when $\eta_n$ is the BLUP (simple-kriging predictor) for the model $\GP(0,K_{5/2,5})$ on $[0,1]^2$; $Y_\xb\sim\GP(0,K_{3/2,10})$, $K^{(e)}=K_{3/2,\mt_{\rm BLP}}$, $\mt_{\rm BLP}\in[0.05,20]$ ($\Xb_n$ is a regular grid of 100 design points).} \label{F:ISE=M1-M52-theta5_d2_n100_M0-M32-theta10_M2-M32}
\end{figure}

We construct $\hISE_{BLP}(\eta_n)$ for the model $\GPmodel{e}$, using $K^{(e)}=K_{3/2,\mt_{\rm BLP}}$ with $\mt_{\rm BLP}\neq \mt_0$, i.e., $K^{(e)}$ and $K$ have the same regularity but different correlation lengths.
Figure~\ref{F:ISE=M1-M52-theta5_d2_n100_M0-M32-theta10_M2-M32} presents the same information as Figure~\ref{F:ISE=M1-poly_d2_n100_m50_M0-M32-theta10_M2-M32} in this setting. Comparison of the two figures shows that predictions by the BLUP for the model $\GPmodel{p}$ are significantly more precise than with the polynomial model of Section~\ref{S:polynomial}. Here, $\widetilde\Ex\{\ISE(\eta_n)\}$ and $\Ex\{\ISE(\eta_n)\}$ are practically confounded on the left panel; on the right panel, $\MSE\{\hISE_{BLP}(\eta_n)\}$ is again minimum for $\mt_{\rm BLP}=\mt_0=10$. In view of the values of $\Ex\{\hISE_{LOO}(\eta_n)\}$ and $\MSE\{\hISE_{LOO}(\eta_n)\}$ indicated in Table~\ref{Tb:indep-lims-ex2}, $\hISE_{BLP}(\eta_n)$ performs significantly better than $\hISE_{LOO}(\eta_n)$ for the whole range of values of $\mt_{\rm BLP}$ considered. Note that the plots are for $\mt_{\rm BLP}\geq 0.05$ and the numerical difficulties caused by the singularity of the flat limit $\Sb_n^{(e)}(0)$ of the matrix $\Sb_n^{(e)}$ are already apparent for $\mt_{\rm BLP}$ close to $0.05$; see Section~\ref{S:limits}.

\begin{figure}[ht]
\centering
\includegraphics[width=0.49\textwidth]{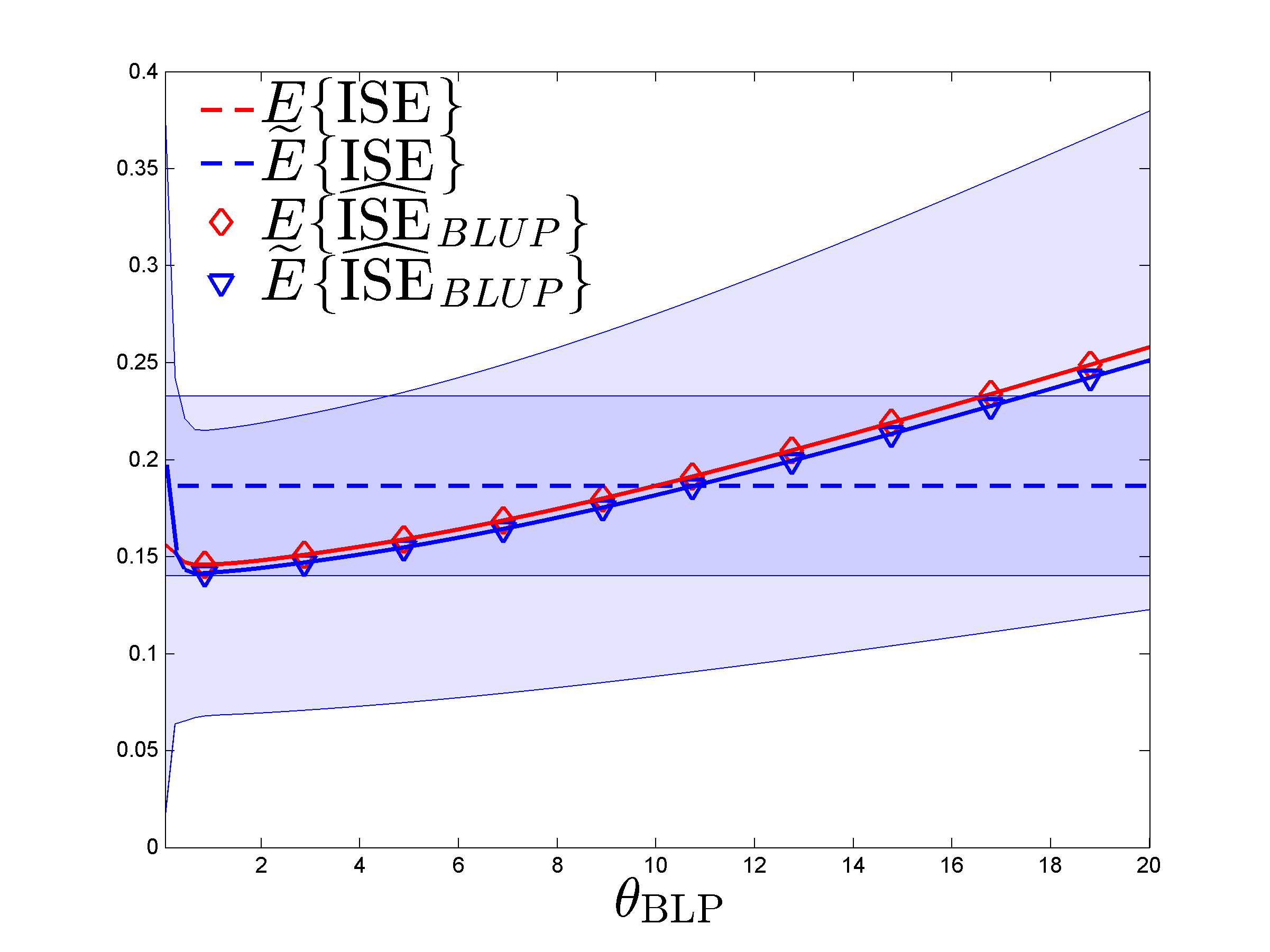}
\includegraphics[width=0.49\textwidth]{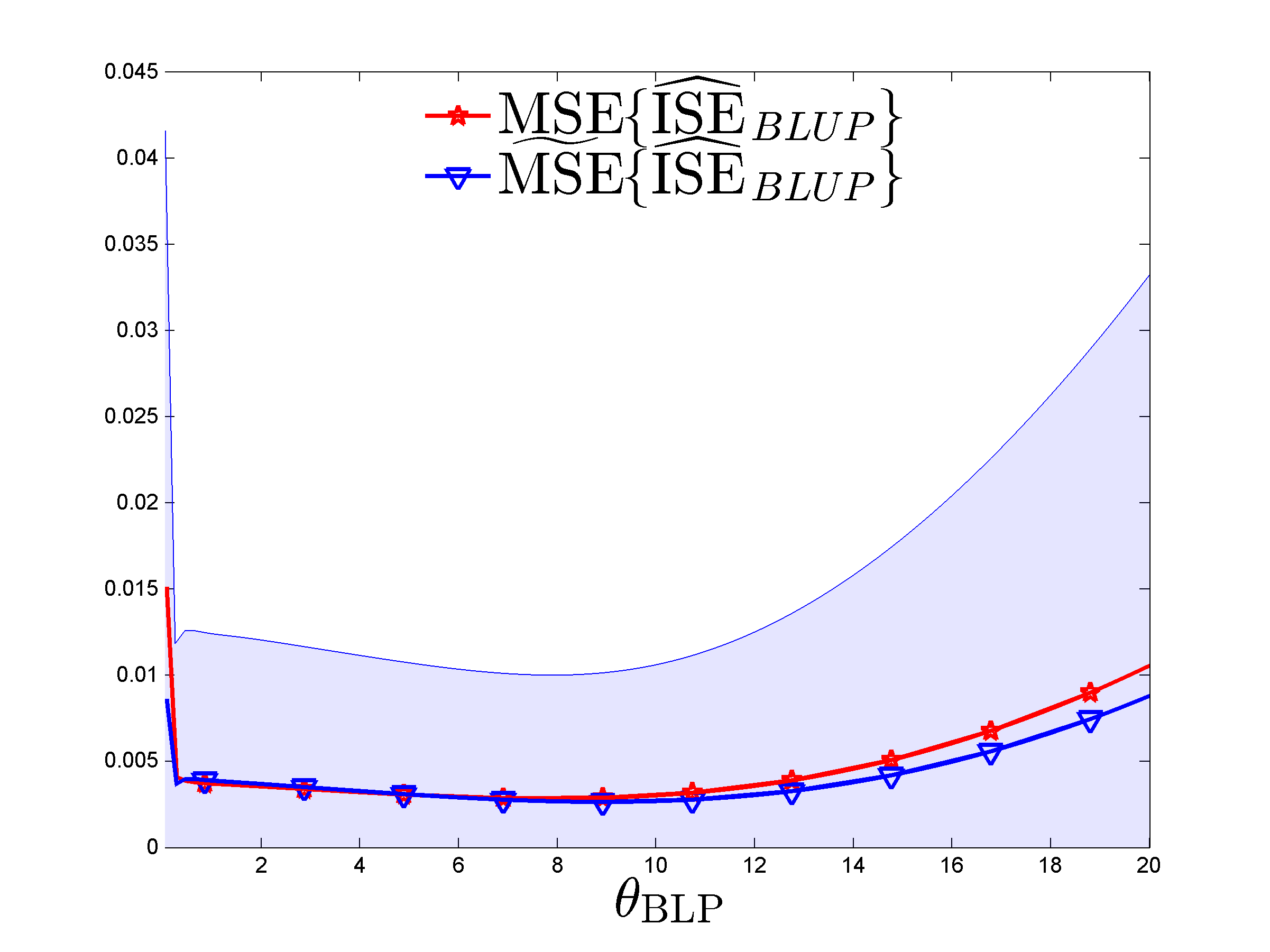}
\caption{\small Same as Figure~\ref{F:ISE=M1-M52-theta5_d2_n100_M0-M32-theta10_M2-M32} but for the unbiased estimator $\hISE_{BLUP}(\eta_n)$ of Section~\ref{S:ISE_BLUP}.} \label{F:ISE=M1-M52-theta5_d2_n100_M0-M32-theta10_M2-M32-unbiased}
\end{figure}

Figure~\ref{F:ISE=M1-M52-theta5_d2_n100_M0-M32-theta10_M2-M32-unbiased} presents the same information as Figure~\ref{F:ISE=M1-M52-theta5_d2_n100_M0-M32-theta10_M2-M32} for the unbiased estimator $\hISE_{BLUP}(\eta_n)$ of Section~\ref{S:ISE_BLUP}. The left panel shows that $\hISE_{BLUP}(\eta_n)$ is indeed unbiased when the model is correct (i.e., for $\mt_{\rm BLP}=10$), but remains biased otherwise (and is more biased than $\hISE_{BLP}(\eta_n)$ for large $\mt_{\rm BLP}$). Its MSE (right panel) is significantly larger (respectively, slightly smaller) than that of $\hISE_{BLP}(\eta_n)$ for large (respectively, small) $\mt_{\rm BLP}$.

\begin{figure}[ht]
\centering
\includegraphics[width=0.49\textwidth]{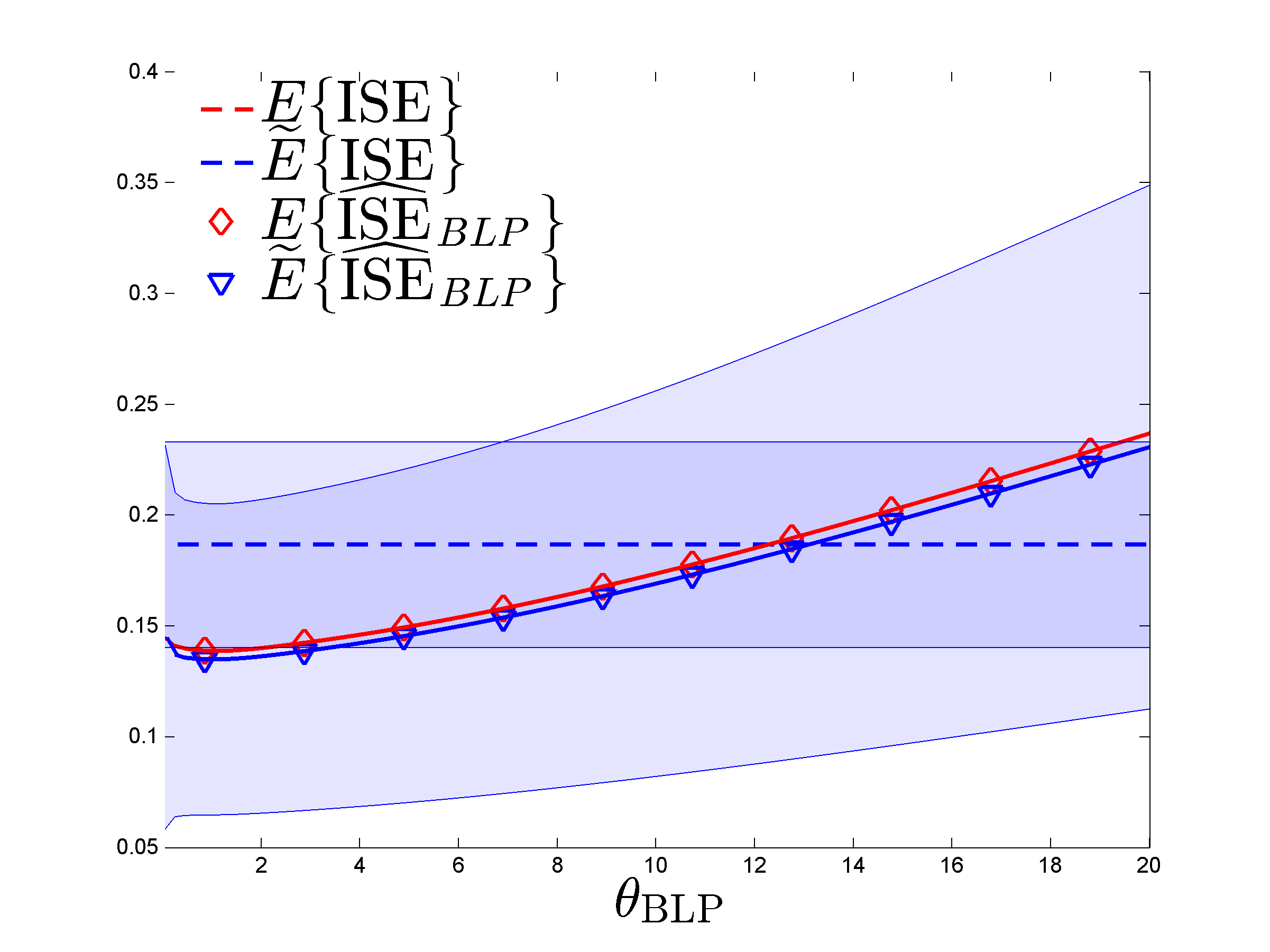}
\includegraphics[width=0.49\textwidth]{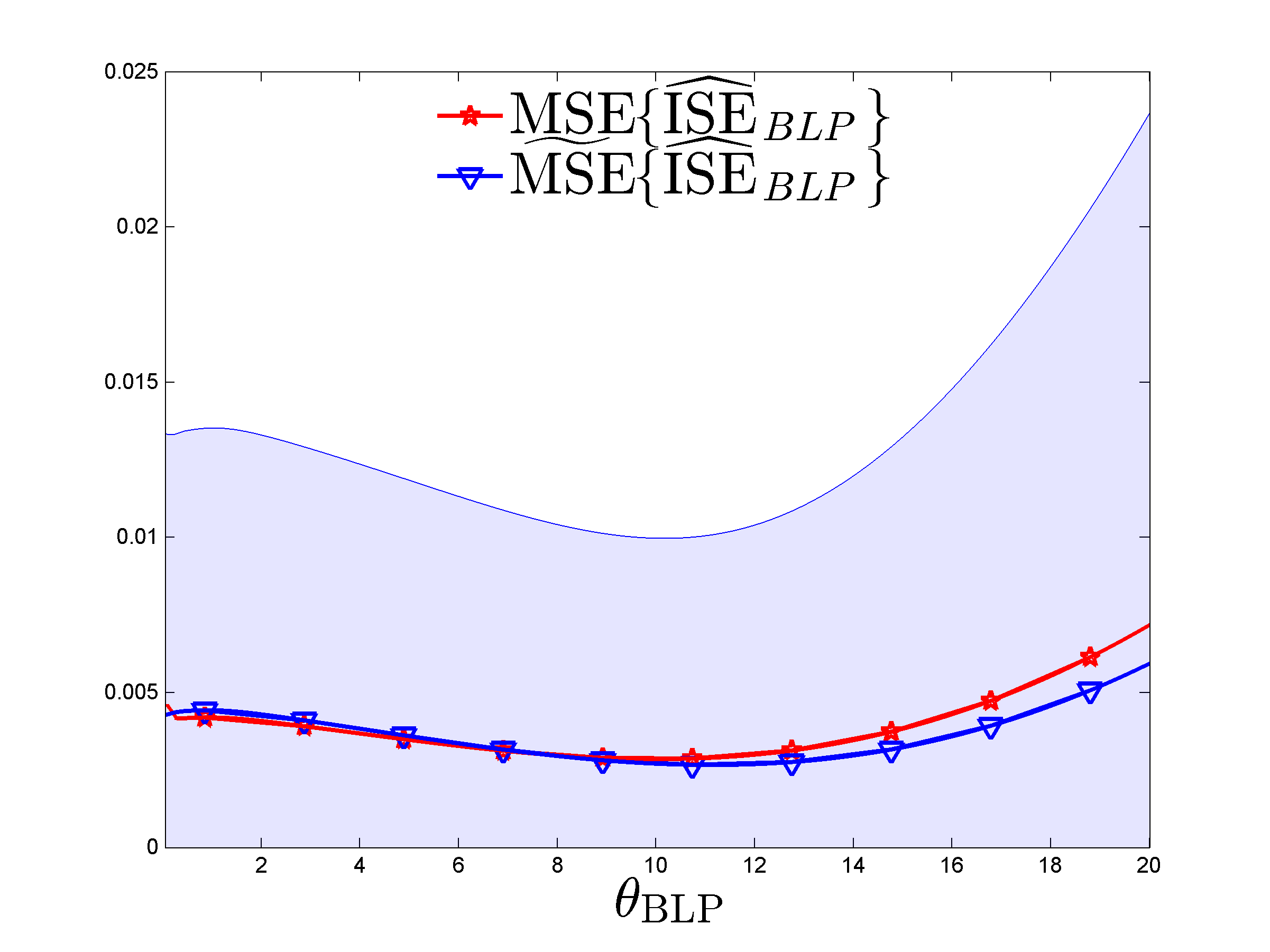}
\caption{\small
Same as Figure~\ref{F:ISE=M1-M52-theta5_d2_n100_M0-M32-theta10_M2-M32} but for the ordinary-kriging predictor $\widehat\eta_n$ for the model $\GP(0,K_{5/2,5})$.} \label{F:ISE_OK=M1-M52-theta5_d2_n100_M0-M32-theta10_M2-M32}
\end{figure}

The behavior of $\hISE_{BLP}(\eta_n)$ is slightly different for small $\mt_{\rm BLP}$ when $\eta_n$ is the ordinary-kriging predictor $\widehat\eta_n$ for the model $\GPmodel{p}$. Here, the vector of weights $\widehat\wb_n(\xb)$ minimizes $\rho_n^2(\xb)$ given by \eqref{rho_n^2-b} for $K=K^{(p)}$ under the constraint $\widehat\wb_n\TT(\xb)\1b_n=1$, and is thus solution of
\bea
\left(
  \begin{array}{cc}
    \Kb_n^{(p)} & \1b_n \\
    \1b_n\TT & 0 \\
  \end{array}
\right) \left(
          \begin{array}{c}
            \widehat\wb_n(\xb) \\
            \ml_n \\
          \end{array}
        \right) = \left(
          \begin{array}{c}
            \kb_n^{(p)}(\xb) \\
            1 \\
          \end{array}
        \right)\,,
\eea
where $\ml_n$ is the Lagrange coefficient for the constraint.
Denoting by $\bKb_n$ the $(n+1)\times(n+1)$ matrix on the left-hand side and $\bMb=\bKb_n^{-1}$, using block-matrix inversion we get $\mve_{-i}=\bMb_{i,1:n}\yb_n/\bMb_{i,i}$ for $i=1,\ldots,n$, and $\Rb_n=\bMb\,\bDb_n$ in \eqref{Rn}, with $\bDb_n=\diag\{1/\bMb_{i,i},\, i=1,\ldots,n\}$.
As $\widehat\wb_n\TT(\xb)\1b_n=1$, $\Rb_n\TT\1b_n=\0b_n$ and $\Sb_n^{(e)}$ and $\bb_n^{(e)}(\xb)$ respectively tend to the null matrix and null vector when $\mt_{\rm BLP}\to 0$.
In agreement with the flat-limit discussion in Section~\ref{S:limits}, when $\mt_{\rm BLP}$ is small we observe a more stable behavior for $\hISE_{BLP}(\eta_n)$
on Figure~\ref{F:ISE_OK=M1-M52-theta5_d2_n100_M0-M32-theta10_M2-M32} for the ordinary kriging predictor $\widehat\eta_n$ than on Figure~\ref{F:ISE=M1-M52-theta5_d2_n100_M0-M32-theta10_M2-M32} for the BLUP $\eta_n$.

In Section~\ref{S:different-regularity} of the supplement, we consider the situation where $K^{(e)}$ and $K$ have different regularities. We still use $K=K_{3/2,10}$ and $\eta_n$ is the simple-kriging predictor for the model $\GP(0,K_{5/2,5})$, but the construction of $\hISE_{BLP}(\eta_n)$ relies on  $K^{(e)}(\xb,\xb')=\psi^{(e)}(\|\xb-\xb'\|)$, where we consider different $\psi^{(e)}$: $\psi_{1/2,\mt_{\rm BLP}}(r)=\exp(-\mt_{\rm BLP}\, r)$, $\psi_{5/2,\mt_{\rm BLP}}(r)$ given by \eqref{Matern52}, $\psi_{\IM,\mt_{\rm BLP}}(r)=(1+\mt_{\rm BLP}^2\, r^2)^{-1}$ and $\psi_{\infty,\mt_{\rm BLP}}(r)=\exp(-\mt_{\rm BLP}^2\, r^2)$, corresponding respectively to the Mat\'ern 1/2, Mat\'ern 5/2, inverse multiquadric and Gaussian kernel.
Our conclusion is that the choice of $K^{(e)}$ is not crucial, provided it is regular enough (possibly more regular than $K$) and $\mt_{\rm BLP}$ is not excessively small.

\subsection{An environmental model} \label{S:environmental-model}

This example uses the model of \cite{BliznyukRSRWM2008} that describes the pollutant spill caused by a chemical accident. We use the implementation given at \url{https://www.sfu.ca/~ssurjano/environ.html}, with parameters set at the values $M = 10$, $D = 0.07$, $L = 1.505$ and $\tau = 30.1525$, as on the figure shown there. The space-time design variables are taken in $\SX=[0,3]\times[1,60]$, which we renormalize to $[0,1]^2$. The function varies approximately between 0 and 70 over $\SX$, with a rather sharp peak at the center of $\SX$.

As the function is fixed, we use random designs to provide a statistical comparison between methods operating in various conditions. We generate random $n$-point designs in $[0,1]^2$, with $n=200$, using the relaxed greedy-packing algorithm of \cite{PZ2022}. The construction uses $\xb_1=(1/2,1/2)\TT$ and then $\xb_{k+1}=\ma_k \xb_i +(1-\ma_k) \xb^*$ for $k\geq 1$, where $\xb^*\in\Arg\max_{\xb\in\SX} \min_{\xb_j\in\Xb_k} \|\xb-\xb_j\|$, $\xb_i\in\{\xb_\ell\in\Xb_k:\, \|\xb^*-\xb_\ell\|=\min_{\xb_j\in\Xb_k} \|\xb^*-\xb_j\|\}$, and the $\ma_k$ are independently uniformly distributed in $[0,a]$, $0\leq a<1$. To make the method implementable, we select $\xb^*$ within a finite subset $\SX_N$ of $\SX$: $\SX_N$ corresponds to the first $N=2^{12}$ Sobol' points in $[0,1]^2$. The same points are used to approximate integrals; i.e., $\mu$ is the uniform measure on $\SX_N$. From \cite[Th.~3.6]{PZ2022}, the packing (respectively, covering) efficiencies of such $\Xb_n$ with respect to optimal packing (respectively, covering) designs in $\SX_N$ equal at least $(1-a)/2$ for any $n \leq N$. We take $a=0.2$, which yields efficiencies at least 40\%.

The predictor for which we estimate the ISE is intentionally not well adapted to this situation: $\eta_n$ is the BLUP for the model $\GPmodel{p}$ with $K^{(p)}=K_{3/2,\mt_p}$, see \eqref{Matern32}, i.e., $\eta_n$ is the simple-kriging predictor for that model. (The ordinary-kriging predictor would be would be a better choice, as the mean of $f$ over $\SX_N$ is about 9.5.) The estimator $\hISE_{BLP}(\eta_n)$ uses the kernel $K^{(e)}(\xb,\xb')=\psi_{5/2,\mt_{\rm BLP}}(\|\xb-\xb'\|)$, see \eqref{Matern52}, and different values of $\mt_{\rm BLP}$ are considered.

\begin{figure}[ht]
\centering
\includegraphics[width=0.49\textwidth]{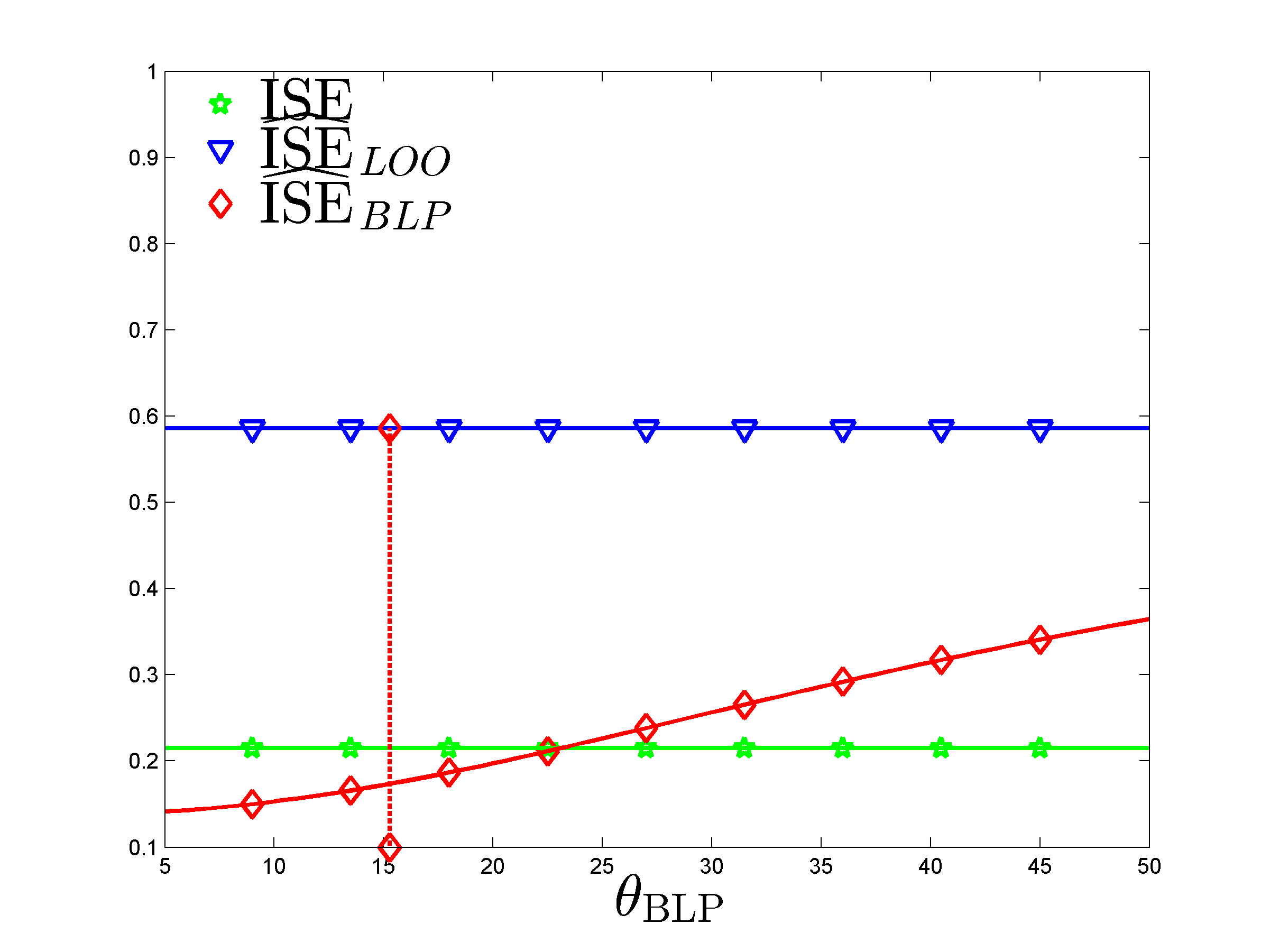}
\includegraphics[width=0.49\textwidth]{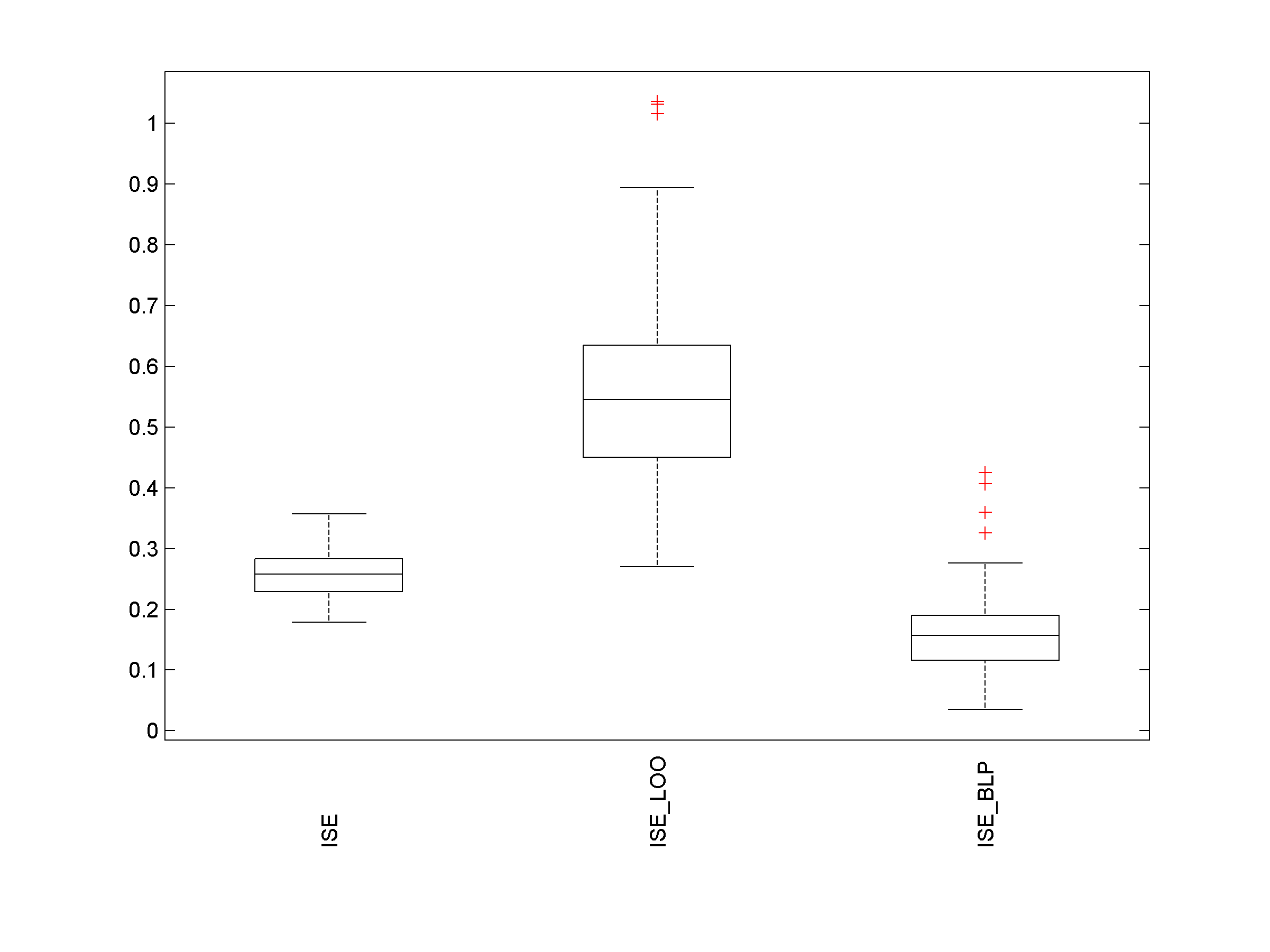}
\caption{\small Environmental model: $\eta_n$ is the BLUP for the model $\GP(0,\ms^2\, K_{3/2,1})$ and $K^{(e)}(\xb,\xb')=\psi_{5/2,\mt_{\rm BLP}}(\|\xb-\xb'\|)$. Left: $\ISE(\eta_n)/\mo_n$, $\hISE_{LOO}(\eta_n)/\mo_n$ and $\hISE_{BLP}(\eta_n)/\mo_n$ as functions of $\mt_{\rm BLP}$, with $\mo_n$ the empirical variance of the $f(\xb_i)$, for one random design having packing and covering efficiencies at least 40\%; the value $\widehat\mt_{LOO}$ that minimizes $\hISE_{LOO}(\eta_n)$ with $\eta_n$ the BLUP for the model $\GP(0,\ms^2 K_{5/2,\mt})$ is indicated by a vertical line.
Right: boxplots of $\ISE(\eta_n)/\mo_n$, $\hISE_{LOO}(\eta_n)/\mo_n$ and  $\hISE_{BLP}(\eta_n)/\mo_n$, for 100 random designs having packing and covering efficiencies at least 40\% ($\mt_{\rm BLP}=\widehat\mt_{LOO}$ in $\hISE_{BLP}(\eta_n)$).
} \label{F:ISEs_varyn_envmodel}
\end{figure}

\vsp
We first set $\mt_p=1$ in $K^{(p)}$, which provides smooth predictions $\eta_n$ for designs $\Xb_n$ well spread over $\SX$. The left panel of Figure~\ref{F:ISEs_varyn_envmodel} is for a single (typical) design $\Xb_n$ and shows $\hISE_{BLP}(\eta_n)/\mo_n$ as a function of $\mt_{\rm BLP}$, with $\mo_n$ the empirical variance of the $f(\xb_i)$, $\xb_i\in\Xb_n$. Values of $\ISE(\eta_n)/\mo_n$ and $\hISE_{LOO}(\eta_n)/\mo_n$ (not depending on $\mt_{\rm BLP}$) are also shown: $\hISE_{LOO}(\eta_n)$ provides a severe overestimation of $\ISE(\eta_n)$;
$\hISE_{BLP}(\eta_n)$ is significantly more accurate for the range considered for $\mt_{\rm BLP}$. The vertical line indicates the value $\widehat\mt_{LOO}$, the LOOCV estimator of $\mt$ that minimizes $\hISE_{LOO}(\eta_n^*)$, with $\eta_n^*$ the BLUP for the model $\GP(0,\ms^2 K_{5/2,\mt})$; see \eqref{ISE_{LOO}_gp}.

Here, $(1/n)\sum_{i=1}^n y_i \simeq 9.59$ suggesting the use of a GP model with nonzero mean for the construction of $\hISE_{BLP}(\eta_n)$.
However, the weights $\wb_n(\xb)$ of the predictor $\eta_n$ satisfy $\int_\SX [1-\wb_n\TT(\xb)\1b_n]^2\,\mu(\dd\xb) \simeq 2 \cdot 10^{-8}$, and due to the robustness of $\hISE_{BLP}(\eta_n)$ to the presence of a non-zero constant trend when $\wb_n\TT(\xb)\1b_n \approx 1$ for all $\xb$ (see Section~\ref{S:A-OK} in the supplement), the correction of Section~\ref{S:parameterised-mean} is not required.

On the right panel of Figure~\ref{F:ISEs_varyn_envmodel}, to confirm that the results above are not due to a particularly favorable choice of $\Xb_n$ we consider 100 random designs (all with packing and covering efficiencies at least 40\%). We have seen in Section~\ref{S:numerical-robustness} that the estimator $\hISE_{BLP}(\eta_n)$ is not very sensitive to the choice of $K^{(e)}$, suggesting that the precise data fitting of a GP model $\GPmodel{e}$ is not needed: we use only the isotropic kernel $K^{(e)}(\xb,\xb')=\psi_{5/2,\mt_{\rm BLP}}(\|\xb-\xb'\|)$, although other, non-isotropic, kernels may be more suitable; $\mt_{\rm BLP}$ is chosen by LOOCV estimation rather than maximum likelihood due  the superior robustness of LOOCV to model misspecification, see, e.g., \cite{Bachoc2013}, and we take $\mt_{\rm BLP}=\min\{\max\{\widehat\mt_{LOO},5\},50\}$ (note that $\widehat\mt_{LOO}$ is different for each $\Xb_n$) to compute $\hISE_{BLP}(\eta_n)$.
The maximum of $\int_\SX [1-\wb_n\TT(\xb)\1b_n]^2\,\mu(\dd\xb)$ over the 100 designs considered is less than $6\cdot 10^{-8}$ and we ignore again the presence of a nonzero mean. The figure presents boxplots of the normalized $\ISE(\eta_n)$ (divided by the variance $\mo_n$ of the observations $f(\xb_i)$, $\xb_i\in\Xb_n$) and of the normalized estimates $\hISE_{LOO}(\eta_n)$ and $\hISE_{BLP}(\eta_n)$. Note the much better performance of $\hISE_{LOO}(\eta_n)$ (although significantly worse than that of $\hISE_{BLP}(\eta_n)$) compared to the example of Section~\ref{S:polynomial}.
The computational time\footnote{Computations are in Matlab, on a PC with a clock speed of 2.5 GHz and 32 GB RAM.} for the construction of $\eta_n(\xb)$ for a given $\Xb_n$ and all $\xb\in\SX_N$ is about 0.04 s and the calculation of $\hISE_{BLP}(\eta_n)$ takes about 0.09 s (average values over 100 repetitions).

\begin{figure}[ht]
\centering
\includegraphics[width=0.49\textwidth]{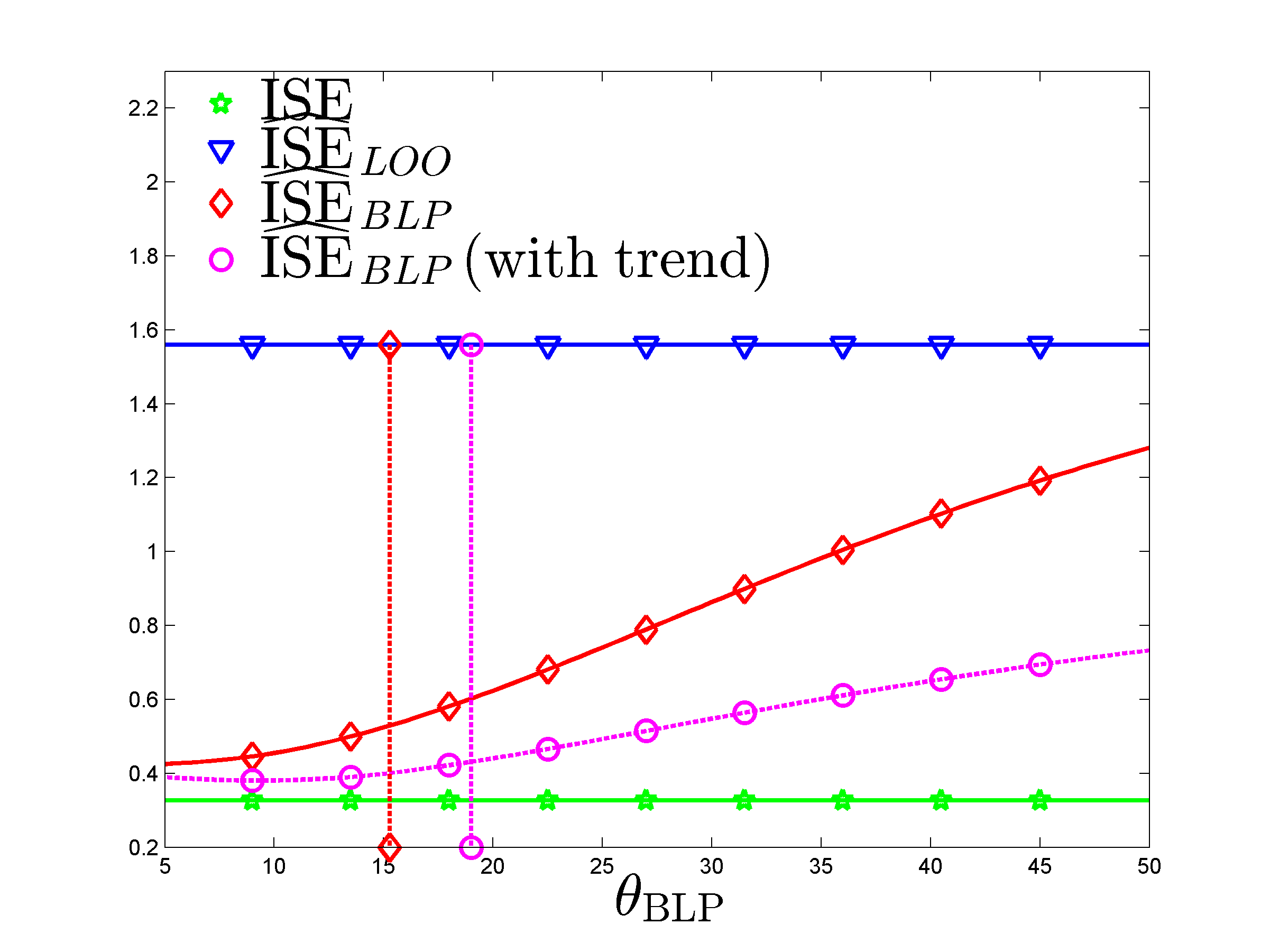}
\includegraphics[width=0.49\textwidth]{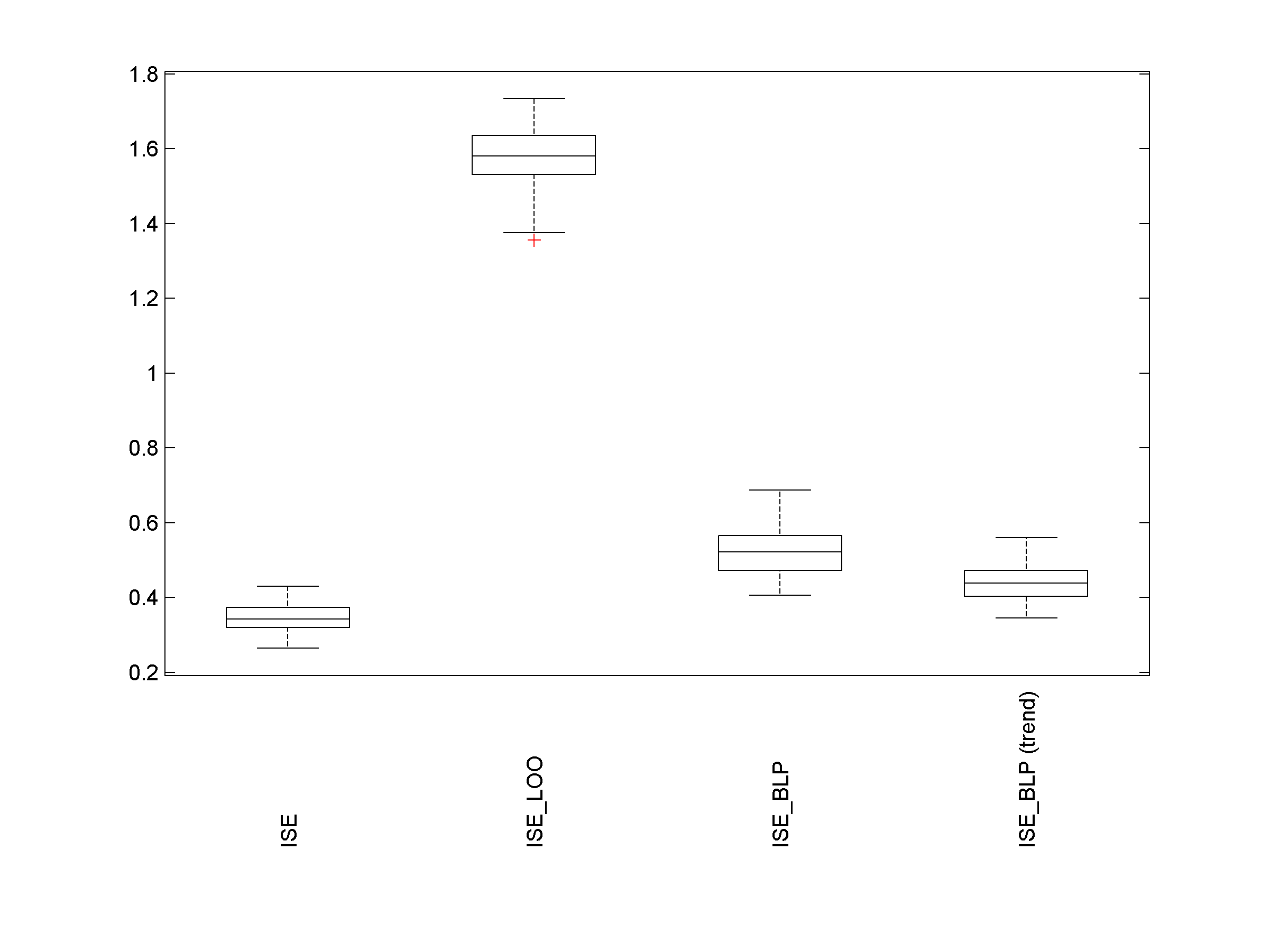}
\caption{\small Same as in Figure~\ref{F:ISEs_varyn_envmodel}, but $\eta_n$ is the BLUP for the model $\GP(0,\ms_p^2\, K_{3/2,\mt_p})$  with $\mt_p=1.5546/[2\,\PR(\Xb_n)]$. Two versions of $\hISE_{BLP}(\eta_n)$ are considered: one that ignores the nonzero trend, the other that uses the method in Section~\ref{S:parameterised-mean} of the supplement. On the left panel, the vertical lines indicate the values of $\widehat\mt_{LOO}$ for the models $\GP(0,K_{5/2,\mt})$ ({\color{red} $\lozenge$- -
 -$\lozenge$}) and $\GP(\tau,K_{5/2,\mt})$ ({\color{magenta} $\circ$- -
 -$\circ$}).} \label{F:ISEs_varyn_envmodel-poor-model}
\end{figure}

\vsp
We use now a predictor with $\mt_p=1.5546/[2\,\PR(\Xb_n)]$, with $\PR(\Xb_n)=(1/2)\,\min_{\xb_i\neq\xb_j}\|\xb_i-\xb_j\|$ the packing radius of $\Xb_n$ (so that $\psi_{3/2,\mt_p}[2\,\PR(\Xb_n)]\simeq 0.25$, corresponding to a model with rather weak correlation); the shorter correlation length induces a slightly inflated value of $\ISE(\eta_n)$ compared to previous case with $\mt_p=1$.
Figure~\ref{F:ISEs_varyn_envmodel-poor-model} is the counterpart of Figure~\ref{F:ISEs_varyn_envmodel} for this new situation.
On the left panel, the design is the same as in the left panel of Figure~\ref{F:ISEs_varyn_envmodel}, we have $\mt_p \simeq 32.8$,
$\int_\SX [1-\wb_n\TT(\xb)\1b_n]^2\,\mu(\dd\xb) \simeq 0.09$ and the performance of $\hISE_{BLP}(\eta_n)$ deteriorates compared with Figure~\ref{F:ISEs_varyn_envmodel} when the nonzero mean is ignored (red curve with diamonds).
For all values of $\mt_{\rm BLP}$ considered, $\hISE_{BLP}(\eta_n)$ becomes significantly closer to $\ISE(\eta_n)$ when the trend is taken into account via the approach in Section~\ref{S:parameterised-mean} of the supplement (magenta curve with circles): we assume the model $\GP(\tau,\ms^2\, K_{5/2,\mt_{\rm BLP}})$, estimate $\ISE_0(\eta_n)$ for centered data as indicated in \eqref{ISE_0},
and then add $I(\widehat\tau^n)$; see Section~\ref{S:A-OK}.
The vertical lines indicate the values of $\widehat\mt_{LOO}$ for the two models $\GP(0,\ms^2 K_{5/2,\mt})$ and $\GP(\tau,\ms^2 K_{5/2,\mt})$, i.e., with and without zero mean: in the former case, $\widehat\mt_{LOO}$ minimizes $\hISE_{LOO}(\eta_n^*)$ with $\eta_n^*$ the BLUP (the simple kriging predictor) for the model $\GP(0,\ms^2 K_{5/2,\mt})$; in the second case $\widehat\mt_{LOO}$ minimizes $\hISE_{LOO}(\widehat\eta_n)$ with  $\widehat\eta_n$ the ordinary kriging predictor for the model $\GP(\tau,\ms^2 K_{5/2,\mt})$ --- see Section~\ref{S:BLP-other-K} for the expression of the LOO residuals $\mve_{-i}$ in this model.

For the right panel of Figure~\ref{F:ISEs_varyn_envmodel-poor-model} we use the same 100 random designs as on the right panel of Figure~\ref{F:ISEs_varyn_envmodel} (which gives  $\mt_p=1.5546/[2\,\PR(\Xb_n)]\in(31,35)$ for the predictor $\eta_n$).
As before, we use  $\mt_{\rm BLP}=\min\{\max\{\widehat\mt_{LOO},5\},50\}$ to compute $\hISE_{BLP}(\eta_n)$, where $\widehat\mt_{LOO}$ minimizes either $\hISE_{LOO}(\eta_n^*)$ or $\hISE_{LOO}(\widehat\eta_n)$ depending whether we assume a GP with zero mean or not.

Here $0.068 < \int_\SX [1-\wb_n\TT(\xb)\1b_n]^2\,\mu(\dd\xb) < 0.112$, which is not negligible contrary to previous case with $\mt_p=1$. When the trend is ignored (third boxplot), $\hISE_{BLP}(\eta_n)$ performs already much better than $\hISE_{LOO}(\eta_n)$ (second boxplot); performance is further  improved when we apply the correction proposed in Section~\ref{S:parameterised-mean} of the supplement to account for the nonzero mean of $Y_\xb$, see the fourth boxplot. As the left panel of Figure~\ref{F:ISEs_varyn_envmodel-poor-model} suggests, better performance could be obtained by choosing a smaller $\mt_{\rm BLP}$. However, optimization with respect to $\mt_{\rm BLP}$ is not feasible in a real practical situation as $\ISE(\eta_n)$ is unknown.

\vsp
We conclude this section by a quick consideration of the problem of model selection. We first highlight that a precise estimator of $\ISE(\eta_n)$ is not an indispensable tool for selecting a predictor from a given class. Indeed, numerical experiments indicate that although $\hISE_{LOO}(\eta_n)$ is often a poor estimate of $\ISE(\eta_n)$, showing an important positive bias, the predictor that minimizes this estimate has often a small ISE: it is the stability of the precision of the ISE estimate when $\eta_n$ varies in the class considered that is important, not the absolute precision itself. Hence, although the better performance of $\hISE_{BLP}(\eta_n)$ as an estimator of $\ISE(\eta_n)$ is an invitation to use
$\hISE_{BLP}(\eta_n)$ for model selection, the gain may be marginal.

\begin{figure}[ht]
\centering
\includegraphics[width=0.59\textwidth]{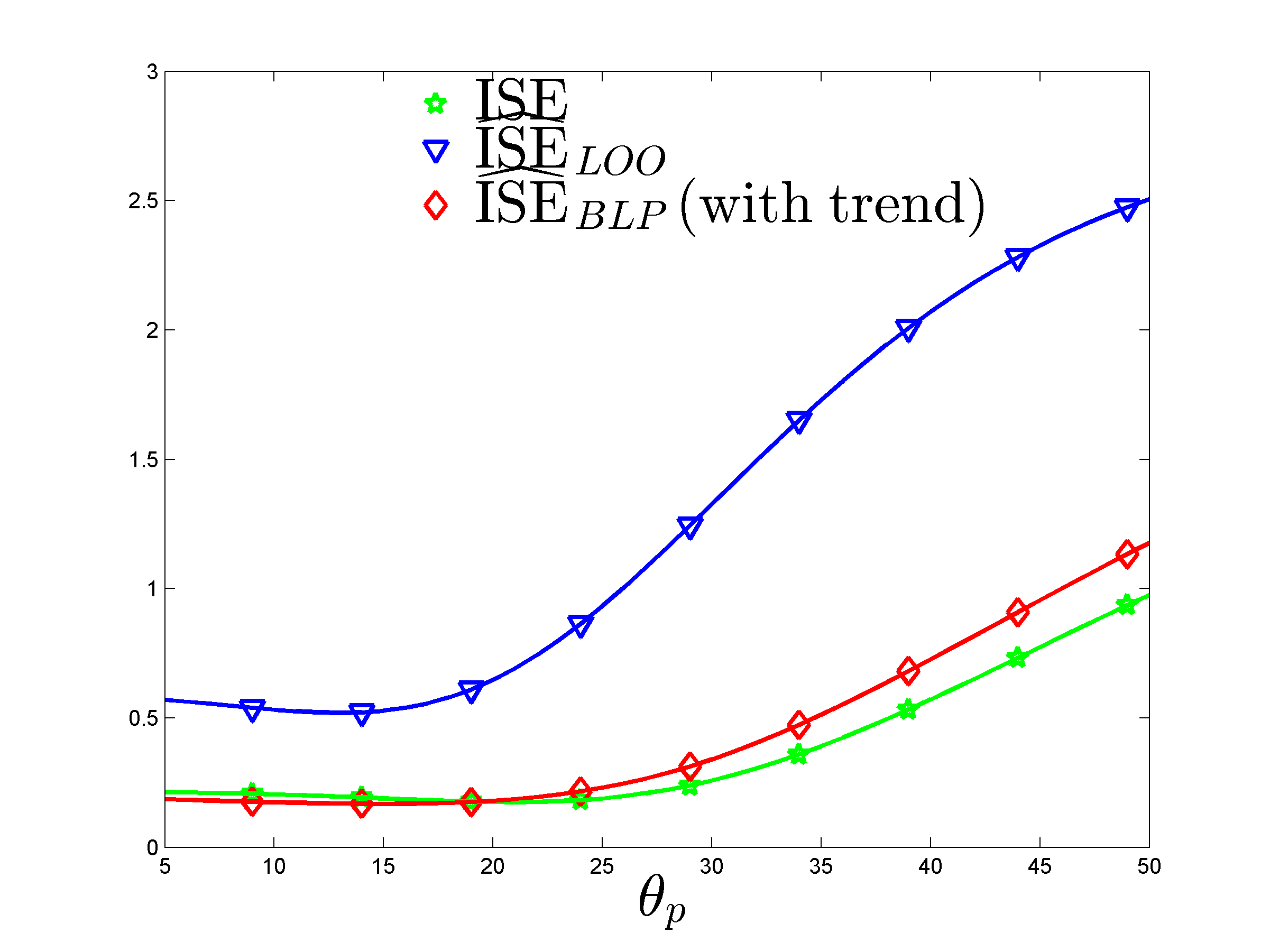}

\caption{\small $\ISE(\eta_n)/\mo_n$, $\hISE_{LOO}(\eta_n)/\mo_n$ and $\hISE_{BLP}(\eta_n)/\mo_n$, for same design as on the left panels of Figs.~\ref{F:ISEs_varyn_envmodel} and \ref{F:ISEs_varyn_envmodel-poor-model},  as functions of $\mt_p\in[5,50]$ (in $\hISE_{BLP}(\eta_n)$, $\mt_{\rm BLP}=\widehat\mt_{LOO}$ for the model $\GP(\tau,K_{5/2,\mt})$).
}
\label{F:predictor-influence-environmental}
\end{figure}

As an illustration, Figure~\ref{F:predictor-influence-environmental} shows, for the same design $\Xb_n$ as on the left panels of Figures~\ref{F:ISEs_varyn_envmodel} and \ref{F:ISEs_varyn_envmodel-poor-model}, the evolution of $\ISE(\eta_n)/\mo_n$, $\hISE_{LOO}(\eta_n)/\mo_n$ and $\hISE_{BLP}(\eta_n)/\mo_n$ as functions of $\mt_p$ when $\eta_n$ is the BLUP for the model $\GP(0,\ms^2_p\,K_{3/2,\mt_p})$ ($\mo_n$ is the variance of the $f(\xb_i)$ for $\xb_i\in\Xb_n$ and $\hISE_{BLP}(\eta_n)$ uses the correction of Section~\ref{S:parameterised-mean} of the supplement, with $\mt_{\rm BLP}=\widehat\mt_{LOO}$ for the model $\GP(\tau,\ms^2 K_{5/2,\mt})$). The estimation of $\ISE(\eta_n)$ by $\hISE_{BLP}(\eta_n)$ is much more precise than by $\hISE_{LOO}(\eta_n)$ for all values of $\mt_p$ considered, but the optimal (minimizing) $\mt_p$
for $\hISE_{LOO}(\eta_n)$ and $\hISE_{BLP}(\eta_n)$
are rather close, with only a slight advantage to the latter (the optimal $\mt_p$ being closer to the value minimizing the true ISE, $\ISE(\eta_n)$).

This is confirmed by the results obtained for 100 random designs.
We select $\eta_n[\mt_p]$, the BLUP for the model $\GP(0,\ms^2\,K_{3/2,\mt_p})$, among the 46 predictors associated with $\mt_p=5,6,\ldots,50$, by minimization of $\hISE_{LOO}(\eta_n[\mt_p])$ or $\hISE_{BLP}(\eta_n[\mt_p])$. The first row of Table~\ref{Tb:selection} gives, for both estimators, and also for the selection based on the oracle $\ISE(\eta_n[\mt_p])$, the empirical mean of $\ISE(\eta_n[\mt_p^{(i)}])/\mo_n$ over the 100 designs, with $\mt_p^{(i)}$ the value associated with the smallest estimated ISE for the $i$-th design. To appreciate the significance of the numerical values in the table, we also computed the true ISE for the trivial predictor given by $\overline{\eta_n}= \yb_n\TT\1b_n/n$, i.e., the empirical mean of the observations, and we indicate in the table (last column) the value
$(1/100) \sum_{i=1}^{100} \ISE(\overline{\eta_n})/\mo_n$. The estimator $\hISE_{BLP}(\eta_n)$ uses the trend-correction approach of Section~\ref{S:parameterised-mean} in the supplement for the model $\GP(\tau,\ms^2\, K_{5/2,\mt_{\rm BLP}})$ and $\mt_{\rm BLP}=\widehat\mt_{LOO}$ for this model.

We can see that $\hISE_{BLP}(\eta_n)$ performs slightly better than $\hISE_{LOO}(\eta_n)$ --- but $\hISE_{LOO}(\eta_n)$ performs surprisingly well if we consider its strong overestimation of the true ISE (it turns out that both estimators very rarely select the same model as the oracle that uses $\ISE(\eta_n[\mt_p])$; see also Figure~\ref{F:design-influence} of Section~\ref{S:design-influence} for another illustration).

\begin{table}[htp]
\caption{\small $(1/100) \sum_{i=1}^{100} \ISE(\eta_n[\mt_p^{(i)}])/\mo_n$.}
\begin{center}
				\begin{tabular}{|l|l|l|l|l|}
					\hline	
ISE estimator   & oracle $\ISE(\eta_n)$  & $\hISE_{LOO}(\eta_n)$ & $\hISE_{BLP}(\eta_n)$ & $\ISE(\overline{\eta_n})$  \\	
\hline
Ex.\ of Section~\ref{S:environmental-model} & 0.197 & 0.238 & 0.224 & 0.775 \\
\hline
Ex.\ of Section~\ref{S:piston-model} & 4.13 $\cdot 10^{-3}$ & 4.47 $\cdot 10^{-3}$ & 4.23 $\cdot 10^{-3}$ & 0.537\\
\hline
				\end{tabular}
\end{center}
\label{Tb:selection}
\end{table}

\subsection{The piston model} \label{S:piston-model}

The example concerns a simplified version of a 7-dimensional piston model that describes the motion of a piston within a cylinder, see  {\url{https://www.sfu.ca/~ssurjano/piston.html}}, with the seven design variables
$x_1=M\in[30,60]$, $x_2=S\in[0.005, 0.020]$, $x_3=V_0 \in[0.002, 0.010]$, $x_4=k \in [1000, 5000]$, $x_5=P_0 \in [90000, 110000]$, $x_6=T_a \in [290, 296]$ and $x_7=T_0 \in [340, 360]$.
As the screening analysis in \cite{Moon2010} indicates that only the first four variables have a significant influence on the model response, we  consider a 4-dimensional reduced version of the model, where the input variables $\xb_i$ for $i=5,6,7$ are set to the mid-point of the above intervals. The variables $\xb=(x_1,\ldots,x_4)$ are renormalized in $\SX=[0,1]^4$, we replace $\SX$ by the finite set $\SX_N$ given by the first $N=2^{16}$ Sobol' points in $\SX$ and take $\mu$ equal to the empirical measure on $\SX_N$. We then generate random $n$-point designs in $\SX$ (with $n=50$), using the same greedy-packing algorithm as in Section~\ref{S:environmental-model}, all having packing and covering efficiencies at least 40\%. The predictor $\eta_n$ is again the BLUP for the model $\GP(0,\ms_p^2\, K_{3/2,\mt_p})$; as the function $f$ is fairly smooth, we take $\mt_p=1$.

\begin{center}
\begin{figure}[ht]
\centering
\includegraphics[width=0.49\textwidth]{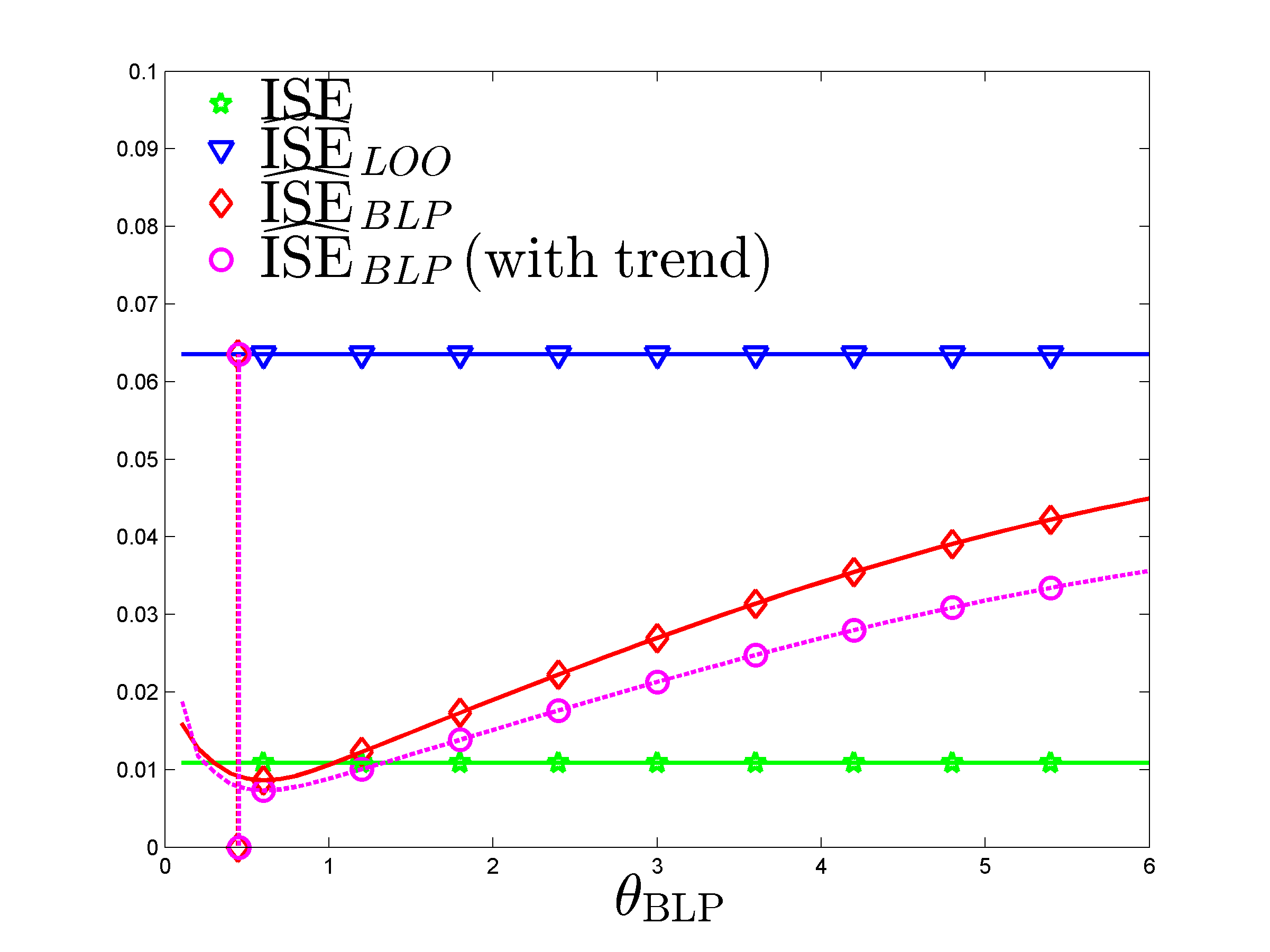}
\includegraphics[width=0.49\textwidth]{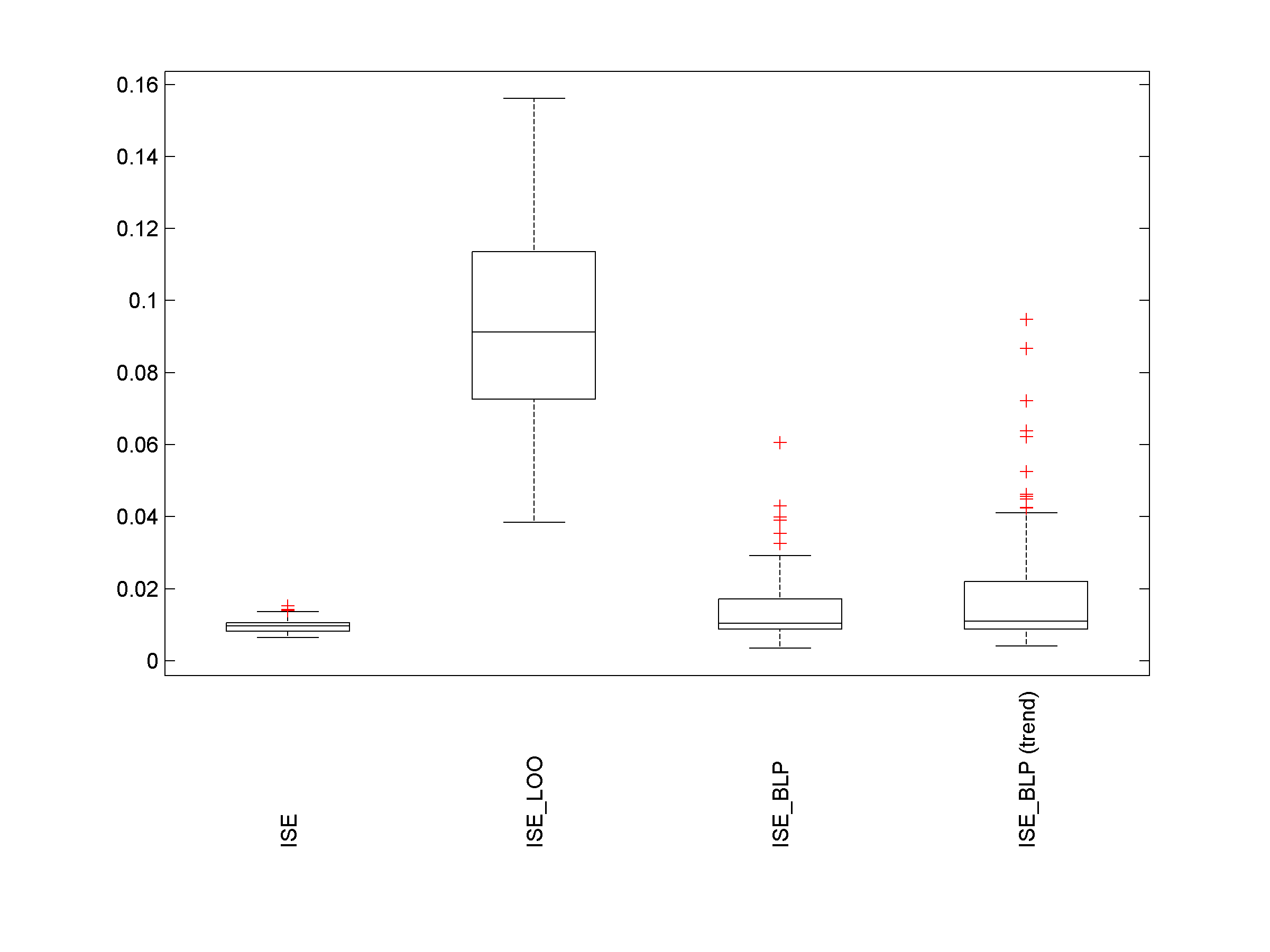}
\caption{\small Same as in Figure~\ref{F:ISEs_varyn_envmodel-poor-model} for the piston model, with $\eta_n$ the BLUP for $\GP(0,\ms^2\, K_{3/2,1})$. On the left panel, the values of $\widehat\mt_{LOO}$ for the models $\GP(0,K_{5/2,\mt})$ ({\color{red} $\lozenge$- -
 -$\lozenge$}) and $\GP(\tau,K_{5/2,\mt})$ ({\color{magenta} $\circ$- -
 -$\circ$}), indicated by vertical lines, are practically confounded.} \label{F:ISEs_piston-model}
\end{figure}
\end{center}

The left panel of Figure~\ref{F:ISEs_piston-model} presents $\hISE_{BLP}(\eta_n)$ as functions of $\mt_{\rm BLP}$, together with $\ISE(\eta_n)$ and $\hISE_{LOO}(\eta_n)$, for a single design $\Xb_n$.
For the red curve with diamonds, the model assumed is $\GPmodel{e}$ with $K^{(e)}=K_{5/2,\mt_{\rm BLP}}$. We have $\int_\SX [1-\wb_n\TT(\xb)\1b_n]^2\,\mu(\dd\xb) \simeq 0.11\cdot 10^{-3}$. For large values of $\mt_{\rm BLP}$ performance slightly improves when we use the model $\GP(\tau,\ms^2\, K_{5/2,\mt_{\rm BLP}})$ with the correction of Section~\ref{S:parameterised-mean} in the supplement (magenta curve with circles), but the reverse is true for small $\mt_{\rm BLP}$, in particular for $\mt_{\rm BLP}=\widehat\mt_{LOO}$.

This is confirmed by the right panel of Figure~\ref{F:ISEs_piston-model}, which displays boxplots obtained for 100 random designs. We have  $0.07\cdot 10^{-3} < \int_\SX [1-\wb_n\TT(\xb)\1b_n]^2\,\mu(\dd\xb) < 0.14\cdot 10^{-3}$, and the trend-correction of Section~\ref{S:parameterised-mean} is not quite necessary: on the opposite, $\hISE_{BLP}(\eta_n)$ shows a significantly stronger variability for the model $\GP(\tau,\ms^2\, K_{5/2,\mt_{\rm BLP}})$ which accounts for the presence of a nonzero mean than for the model $\GP(0,\ms^2\, K_{5/2,\mt_{\rm BLP}})$ ($\hISE_{BLP}(\eta_n)$ uses $\mt_{\rm BLP}=\widehat\mt_{LOO}$ for the model considered). Nevertheless, both estimators perform much better than $\hISE_{LOO}(\eta_n)$. Note the negligible variability of $\ISE(\eta_n)$ across designs due to the strong regularity of $f$.

For a given $\Xb_n$, the computational time of the construction of $\eta_n(\xb)$ for all $\xb\in\SX_N$ is about 0.18 s and the calculation of $\hISE_{BLP}(\eta_n)$ takes about 0.35 s (average values over 100 repetitions).

The performance of $\hISE_{LOO}(\eta_n)$ and $\hISE_{BLP}(\eta_n)$ (based on the model $\GP(0,K_{5/2,\widehat\mt_{LOO}})$) for model selection is summarized in the second row of Table~\ref{Tb:selection}. As in Section~\ref{S:environmental-model}, we select $\eta_n[\mt_p]$ in a finite family: $\eta_n[\mt_p]$ is the BLUP for the model $\GP(0,\ms^2\,K_{3/2,\mt_p})$ with $\mt_p\in\{0.01, 0.02,\ldots,0.5\}$ (50 elements). As for the example of Section~\ref{S:environmental-model}, the predictors selected with $\hISE_{BLP}(\eta_n)$ have, on average, a slightly smaller ISE than those selected with $\hISE_{LOO}(\eta_n)$.

\section{Conclusions and further developments}\label{S:conclusion}

The paper proposes a method that set weights on LOO squared residuals when estimating the ISE of a linear predictor $\eta_n$. The resulting ISE estimator $\hISE_{BLP}(\eta_n)$ is more precise than usual (unweighted) LOOCV, sometimes considerably so. The dependence of the weights on the sampling design gives the estimator a certain robustness to the design configuration, unlike LOOCV. On the downside, the method is not as universal as LOOCV: it is limited to ISE estimation for linear predictors and relies on a GP model (or a mixture of GP models) for the function that the predictor approximates. The numerical examples presented indicate reasonable robustness with respect to the choice of the kernel $K^{(e)}$ of the assumed GP model.

Here we have only considered LOO residuals, but the results in \cite{GinsbourgerS2021} open the way to extension to multiple-fold CV. There, the $i$-th LOO residual $\mve_{-i}$ is replaced by a vector of residuals $\mveb_{I_i}$ at the design points $\Xb_{I_i}=\{\xb_j, \, j\in I_i\}$ with $I_i\subset\{1,\ldots,n\}$, for which only the other points in $\Xb_n\setminus\Xb_{I_i}$ are used for prediction.
The weighted ISE estimator $\hISE_{BLP}(\eta_n)$ for such multiple-fold CV would rely on the construction of the best linear estimator of $\mve_n^2(\xb)$ based on squared residuals $\mveb_{I_i}^{\odot 2}$ for all $I_i$ considered. Under a GP model assumption, the $\mveb_{I_i}$ are Gaussian, and the expressions given in \cite{GinsbourgerS2021} can be used to calculate the expectations needed to compute $\hISE_{BLP}(\eta_n)$, following the same lines as in Section~\ref{S:ISE-any-predictor}. When the sets $I_i$ form a partition of $\{1,\ldots,n\}$, the concatenation of the $\mveb_{I_i}$ forms a vector of length $n$, which entails not major changes compared with the developments in Section~\ref{S:ISE_BLP}; however, when the concatenation forms a vector of length $m>n$, the associated matrix $\Sb_m$ is singular and some adaptation becomes necessary.

One may note that when $Y_\xb\sim\GPmodelg$, $\Ex\{\ISE(\eta_n)\}=\ms^2 J_n$, see \eqref{IMSE_arbitrary}. This observation prompts us to estimate $\ISE(\eta_n)$ by $\widehat\ms^2 J_n$, with $\widehat\ms^2$ an estimator of the process variance $\ms^2$. In particular, assuming that the data are generated with the model $\GPmodel{e}$, we may use the maximum-likelihood estimator
$\widehat \ms^2_{ML} = (1/n) \yb_n\TT{\Kb_n^{(e)}}^{-1}\yb_n$, or the LOO estimator,
\bea
\widehat \ms^2_{LOO} = \frac1n \, \sum_{i=1}^n \Mb_{ii}^{(e)} \mve_{-i}^2 = \frac1n\, \yb_n\TT  \left(\sum_{i=1}^n \frac{\Mb_{\cdot i}^{(e)}\Mb_{i \cdot}^{(e)}}{\Mb_{ii}^{(e)}} \right) \yb_n = \frac1n\, \yb_n\TT \Mb^{(e)}\Db_n^{(e)}\Mb^{(e)}\yb_n \,,
\eea
where $\Mb$ and $\Db_n$ are defined in Section~\ref{S:LOO-etan=etan*}; see, e.g., \cite{Cressie93, Bachoc2013, KarvonenWTOS2020}. Moreover, since $\Ex\{\ISE(\eta_n^*)\}=\ms^2 J_n^*$ when $Y_\xb\sim\GPmodelg$, exploitation of the expression \eqref{ISE-BLP1} for $\hISE_{BLP}(\eta_n^*)$ suggests that we could also estimate $\ms^2$ by the best linear estimator based on squared LOO residuals, $\widehat\ms^2_{BLP}=\hISE_{BLP}(\eta_n^*)/{J_n^{(e)}}^*$ (or the unbiased version $\widehat\ms^2_{BLUP}=\hISE_{BLUP}(\eta_n^*)/{J_n^{(e)}}^*$). For lack of space, we have not reported here the numerical results obtained with the ISE estimators $\widehat\ms^2_{ML} J_n$, $\widehat\ms^2_{LOO} J_n$ and $\widehat\ms^2_{BLP} J_n$, nor have we presented a comparative study of the performances of $\widehat\ms^2_{ML}$, $\widehat\ms^2_{LOO}$ and $\widehat\ms^2_{BLP}$ as estimators of $\ms^2$ (one may refer to \cite{Bachoc2013} for a comparison between $\widehat\ms^2_{ML}$ and $\widehat\ms^2_{LOO}$) and we content ourselves with delivering the raw conclusion of our observations: $\widehat\ms^2_{BLP}$ is often a valid alternative to $\widehat\ms^2_{ML}$ and $\widehat\ms^2_{LOO}$ (in the same way as $\hISE_{BLP}(\eta_n)$ is a valid alternative to  $\hISE_{LOO}(\eta_n)$), but the associated ISE estimators are generally not competitive compared with $\hISE_{BLP}(\eta_n)$, even if they sometimes perform significantly better than $\hISE_{LOO}(\eta_n)$.

Finally, we have presented some preliminary, but promising, results concerning the application of our ISE estimator $\hISE_{BLP}(\eta_n)$ to model selection, in particular to the selection of a GP model when $\eta_n$ is the BLUP for $\GPmodel{p}$.
In that case, if we take $K^{(p)}$ and $K^{(e)}$ in the same family, the method can be iterated, following the same fixed-point principle as for iteratively reweighted least-squares (see, e.g., \cite{Green84}): the first kernel $K_1$ can be initialized through selection by LOOCV; then, at each iteration $j\geq 1$, model selection by minimization of the estimator $\hISE_{BLP}(\eta_n[K^{(p)}])$ constructed with $K^{(e)}=K_j$, yields the kernel $K_{j+1}$ to be used to calculate $\hISE_{BLP}(\eta_n[K^{(p)}])$ at next iteration. We do not expand on this iterative approach here, although it would be worth exploring further.

The limitations of the method are those inherent in the use of GP models for function approximation. In situations where the predictor $\eta_n$ under consideration performs well enough, finding an appropriate GP model for $f$ seems to be a feasible task, making $\hISE_{BLP}(\eta_n)$ a useful tool for estimating $\ISE(\eta_n)$. However, there are situations where $\eta_n$ performs poorly and where it is difficult to find a suitable GP model for $f$; in particular, the design $\Xb_n$ may be too sparse to detect the variability of $f$ (which can happen especially when $d$ is large).
If $\eta_n$ is very inaccurate but has sufficient variability, it is possible that the LOO residuals are large enough for the inaccuracy to be detected by the high value of $\hISE_{LOO}(\eta_n)$ (and possibly of $\hISE_{BLP}(\eta_n)$). However, if $\eta_n$ is much smoother than $f$, it may produce very small LOO residuals and $\hISE_{LOO}(\eta_n)$ may then severely underestimate $\ISE(\eta_n)$ --- and $\hISE_{BLP}(\eta_n)$ will not do any better, the principle of LOOCV itself being ineffective. A simple illustrative example (with $d=1$) is presented in Section~\ref{S:poor-designs} of the supplement. In this case, only the use of an independent test set can help reveal the poor performance of $\eta_n$ (and, as proposed in \cite{FekhariIMPR2022, PR2022}, ISE estimation can rely on the BLP of the squared errors $\mve_n^2(\xb)$ based on the squared test residuals). The Matlab code of a function that calculates $\hISE_{LOO}(\eta_n)$, $\hISE_{BLP}(\eta_n)$ and $\hISE_{BLUP}(\eta_n)$ for a linear predictor $\eta_n$ is given in Section~\ref{S:Matlab-code} of the supplement.




%


\bibliographystyle{siamplain}

\appendix
\section*{Appendix: supplementary material}

\section{A polynomial regression model}\label{S:Appendix-polynomial}

The model assumes that the data $\yb_n$ are given
\bea 
y_i=\phib\TT(\xb_i)\alphab+\delta_i \,,
\eea
where the error vector $\mdb=(\delta_1,\ldots,\delta_n)\TT$ is normally distributed $\SN(0,\mOb_n)$ and where each component $\phi_\ell(\xb)$ of $\phib(\xb)=[\phi_1(\xb),\ldots,\phi_m(\xb)]\TT$ is a multivariate polynomial in the $d$ components of $\xb$. We set a normal prior on $\alphab$, and assume that $\alphab\sim\SN(\0b_m,\mLb)$ with $\mLb=\diag\{\mL_1,\ldots,\mL_m\}$. The posterior mean of $\alphab$ under these assumptions is
$\widehat\alphab=(\Phib\TT\mOb_n^{-1}\Phib+\mLb^{-1})^{-1} \Phib\TT\mOb_n^{-1}\yb_n$, where $\Phib$ is the $n\times m$ matrix with $i$-th row equal to $\phib\TT(\xb_i)$, $i=1,\ldots,n$. The prediction at any given $\xb$ is then
\bea
\eta_n(\xb)= \phib\TT(\xb)\widehat\alphab = \phib\TT(\xb) (\Phib\TT\mOb_n^{-1}\Phib+\mLb^{-1})^{-1} \Phib\TT\mOb_n^{-1}\yb_n \,.
\eea
Straightforward matrix manipulation shows that
\bea
\eta_n(\xb)= \phib\TT(\xb)\mLb\Phib\TT(\Phib\mLb\Phib\TT+\mOb_n)^{-1}\yb_n \,.
\eea
Take $\mOb_n=\mg^2\,\Ib_n$, with $\Ib_n$ the $n$-dimensional identity matrix. Then, denoting $\Kb_n^{(p)}=\Phib\mLb\Phib\TT+\mg^2\,\Ib_n$ and $\kb_n^{(p)}(\xb)=\Phib\mLb\phib(\xb)$, we get $\eta_n(\xb)=[\kb_n^{(p)}(\xb)]\TT{\Kb_n^{(p)}}^{-1}\yb_n$; that is, $\eta_n$ is the BLUP for a GP model with a kernel with nugget effect, defined by $K^{(p)}(\xb,\xb')=K^\phi(\xb,\xb') + \mg^2 \delta_{\xb,\xb'}$, where $\delta_{\xb,\xb'}=1$ when $\xb=\xb'$ and is zero otherwise and
\be\label{tensor-poly}
K^\phi(\xb,\xb') = \sum_{\ell=1}^m \mL_\ell \phi_\ell(\xb)\phi_\ell(\xb')\,.
\ee
Unless $\mg^2=0$ (and $m\geq n$), the predictor $\eta_n$ is not an interpolator. The position of $i$ in $\{1,\ldots,n\}$ is irrelevant to compute the LOO error $\mve_{-i}=y_i-\eta_{n\setminus i}(\xb_i)$, and one may thus consider the case $i=n$. The $i$-th row of ${\Kb_n^{(p)}}^{-1}=(\Kb_n^\phi+\mg^2\Ib_n)^{-1}$ then equals
\bea
\left\{(\Kb_n^\phi+\mg^2\Ib_n)^{-1}\right\}_{i\cdot} = \left(
                                                          \begin{array}{cc}
                                                            -\frac{[\kb_{n\setminus i}^{(p)}(\xb_i)]\TT(\Kb_{n\setminus i}^\phi+\mg^2\Ib_{n-1})^{-1}}{A_i} & \frac{1}{A_i} \\
                                                          \end{array}
                                                        \right)
\eea
with $A_i=K^{(p)}(\xb_i,\xb_i)-[\kb_{n\setminus i}^{(p)}(\xb_i)]\TT(\Kb_{n\setminus i}^\phi+\mg^2\Ib_{n-1})^{-1}\kb_{n\setminus i}^{(p)}(\xb_i)$.
We can then identify the $i$-th row $\rb_i\TT$ of the matrix $\Rb_n$ for the construction of $\mve_{-i}$ in \eqref{Rn}:
\bea
\mve_{-i}=y_i-\eta_{n\setminus i}(\xb_i) &=& y_i-[\kb_{n\setminus i}^{(p)}(\xb_i)]\TT(\Kb_{n\setminus i}^\phi+\mg^2\Ib_{n-1})^{-1} \yb_{n\setminus i} \nonumber\\
&=& \frac{\left\{(\Kb_n^{(p)})^{-1}\right\}_{i\cdot} \yb_n}{\left\{(\Kb_n^{(p)})^{-1}\right\}_{ii}} = \rb_i\TT\yb_n\,. 
\eea

In the example of Section~\ref{S:polynomial}, the polynomial model is constructed by tensorization of univariate polynomials. The index $\ell$ of a component $\phi_\ell(\xb)$ of $\phib(\xb)$ is in fact a multiindex $\underline{\ell}=\{\ell_1,\ldots,\ell_d\}$, with $\phi_{\underline{\ell}}(\xb)=\prod_{i=1}^d \varphi_{\ell_i}(x_i)$ for $\xb=(x_1,\ldots,x_d)\TT$. The degree of the $\varphi_k$ increases with $k$; a scalar $\ml_k$ is attached to each of them, and $\mL_{\underline{\ell}}=\prod_{i=1}^d \ml_{\ell_i}$ with $\ml_k$ decreasing with $k$ in order to give more importance to lower degree polynomials. Only the terms corresponding to the $m$ largest $\mL_{\underline{\ell}}$ is kept to form the kernel \eqref{tensor-poly}. The construction used below relies on Legendre polynomials, orthonormal for the uniform measure on $[0,1]$: $\varphi_0(x)=1$, $\varphi_1(x)=\sqrt{3}(2x-1)$, $\varphi_2=\sqrt{5}(6x^2-6x+1)$, $\varphi_3(x)=\sqrt{7}(20 x^3-30 x^2+12 x-1)$\ldots As $\varphi_k$ has degree $k$, setting $\ml_k=t^{-k}$ for some $t>1$ gives $\mL_{\underline{\ell}}=t^{-\sum_{i=1}^d \ell_i}$ and thus implies that the terms selected in \eqref{tensor-poly} are among those with lowest total degree. One may refer to \cite{P-RESS2019} for implementation details.

\section{Robustness of $\widehat\ISE_{BLP}(\eta_n)$ to the choice of $K^{(e)}$: $K^{(e)}$ and $K$ have different regularities}\label{S:different-regularity}

This is a continuation of Section~\ref{S:BLP-other-K}.
We still use $K=K_{3/2,10}$ and $\eta_n$ is the simple-kriging predictor for the model $\GP(0,K_{5/2,5})$, but the construction of $\widehat\ISE_{BLP}(\eta_n)$ relies on  $K^{(e)}(\xb,\xb')=\psi^{(e)}(\|\xb-\xb'\|)$, where we consider different $\psi^{(e)}$:
\be
\psi_{1/2,\mt_{\rm BLP}}(r) &=& \exp(-\mt_{\rm BLP}\, r) \,, \nonumber \\
\psi_{5/2,\mt_{\rm BLP}}(r) &=&  \left[1+\sqrt{5}\,\mt_{\rm BLP}\, r + (5/3)\, \mt_{\rm BLP}^2\, r^2\right] \exp(-\sqrt{5}\,\mt_{\rm BLP}\, r)\,, \label{psi52} \\
\psi_{\IM,\mt_{\rm BLP}}(r) &=& (1+\mt_{\rm BLP}^2\, r^2)^{-1}\,, \label{psiIM} \\
\psi_{\infty,\mt_{\rm BLP}}(r) &=& \exp(-\mt_{\rm BLP}^2\, r^2)\,, \nonumber
\ee
corresponding respectively to the Mat\'ern 1/2, Mat\'ern 5/2, inverse multiquadric and Gaussian kernel.

\begin{center}
\begin{figure}
\centering
\includegraphics[width=0.39\textwidth]{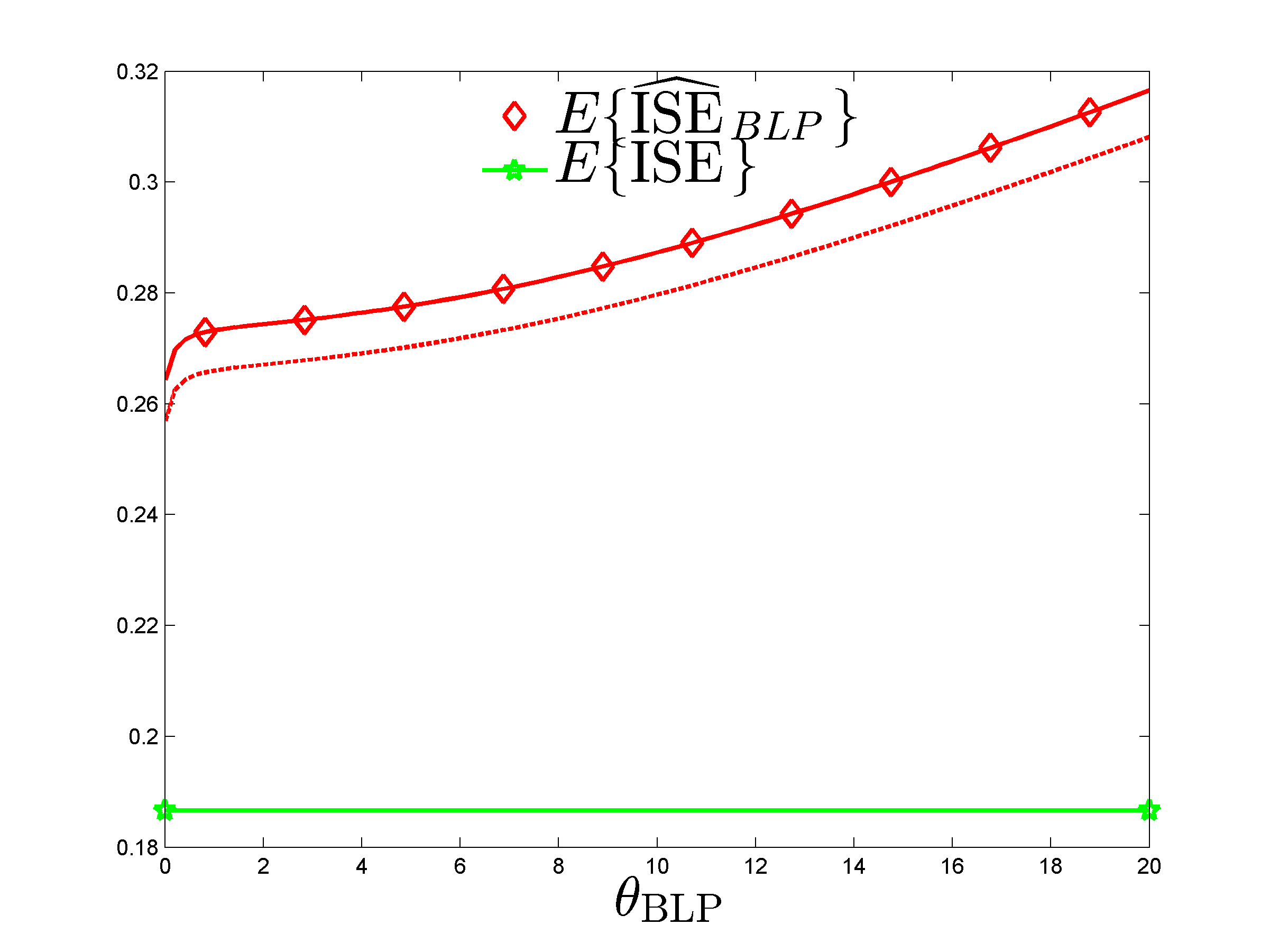}
\includegraphics[width=0.39\textwidth]{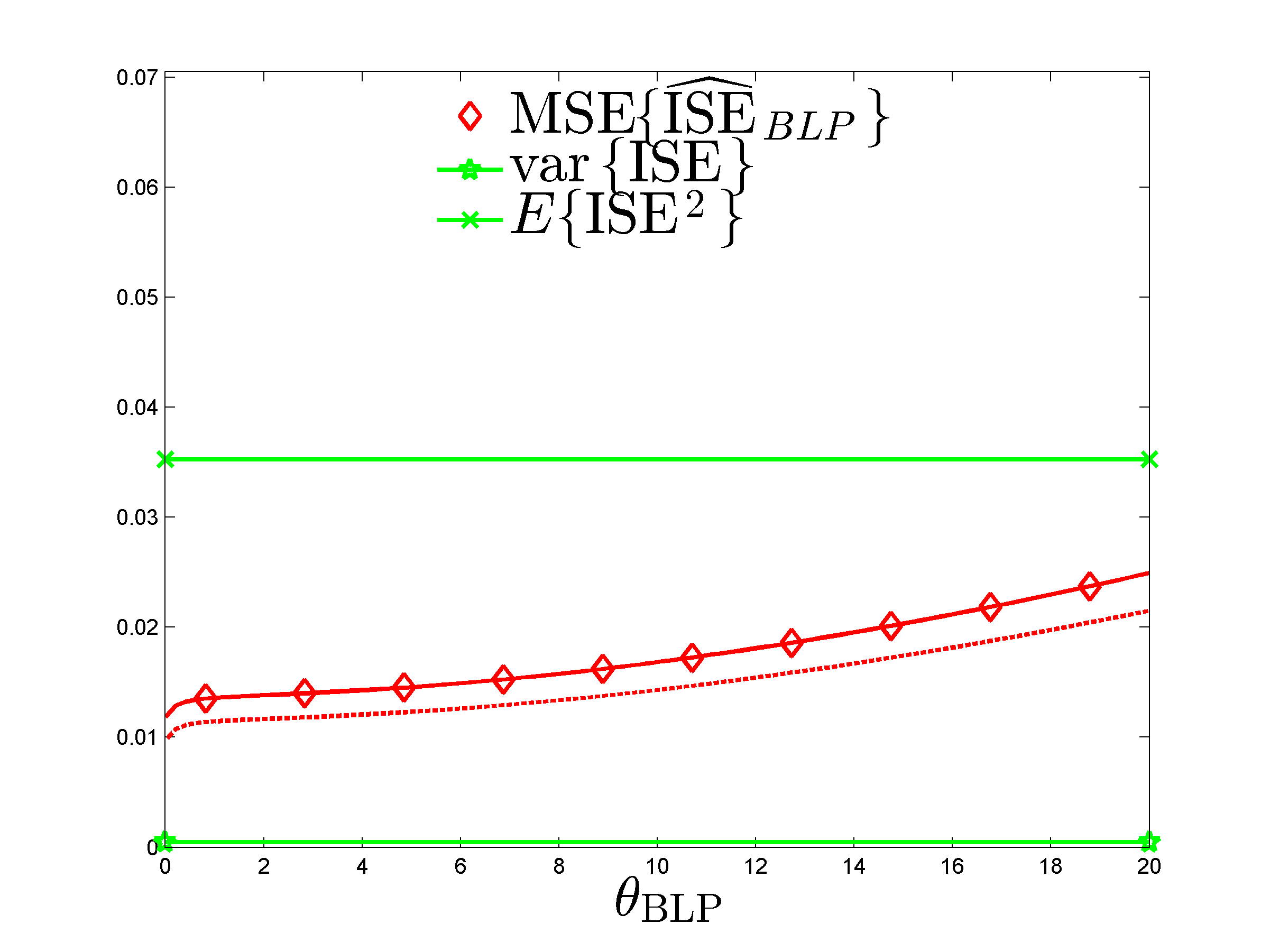} \\
\includegraphics[width=0.39\textwidth]{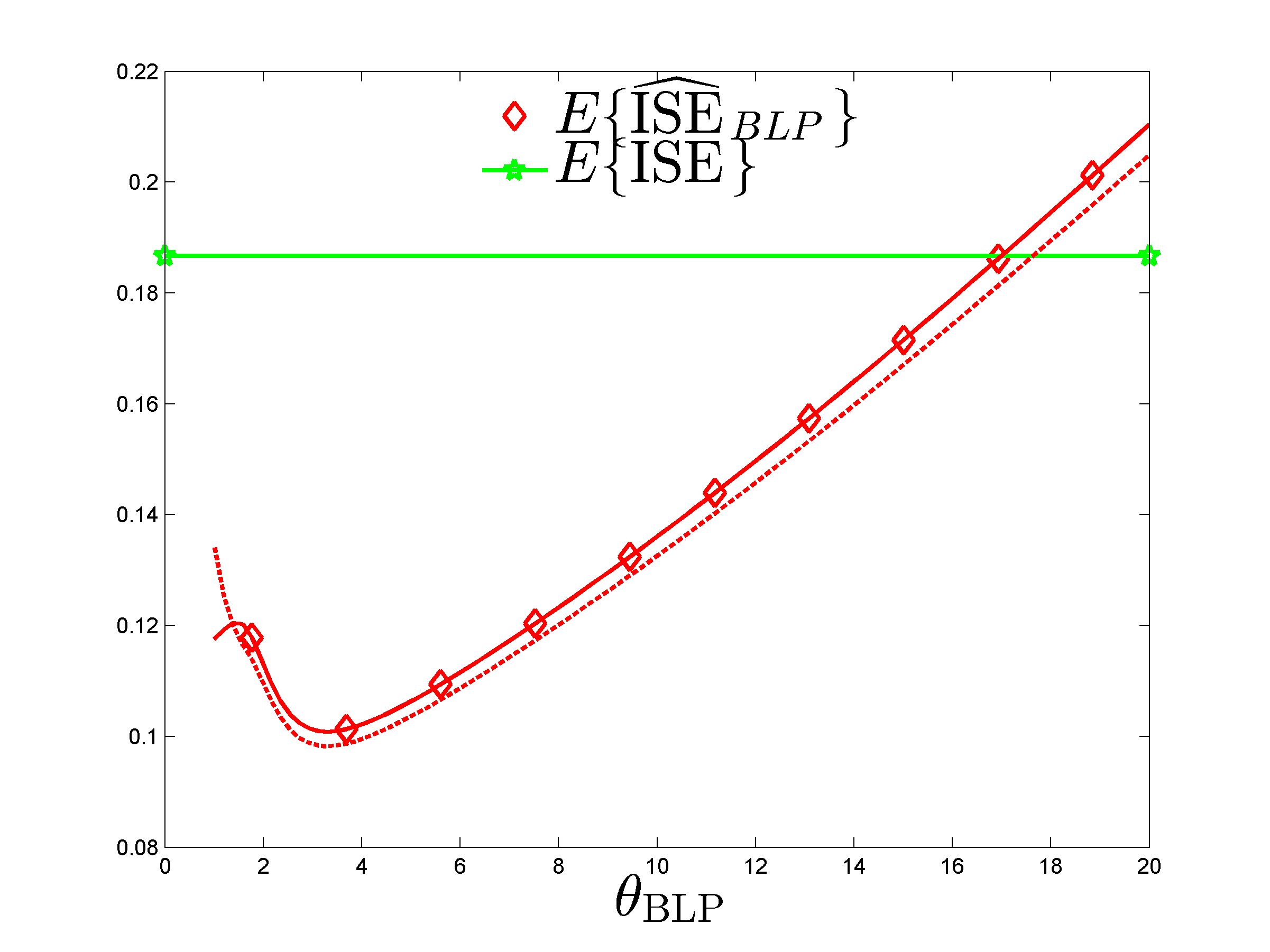}
\includegraphics[width=0.39\textwidth]{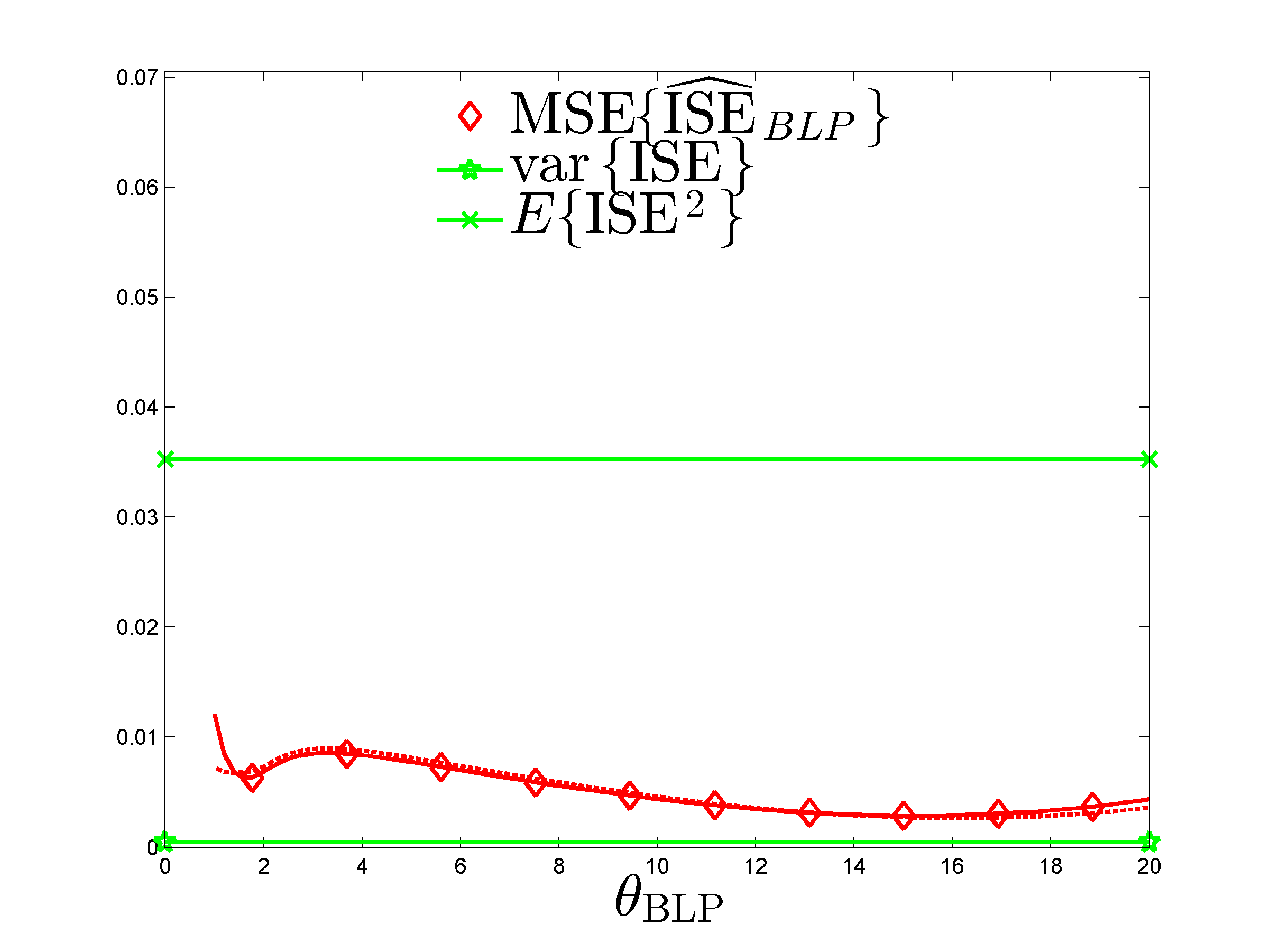}\\
\includegraphics[width=0.39\textwidth]{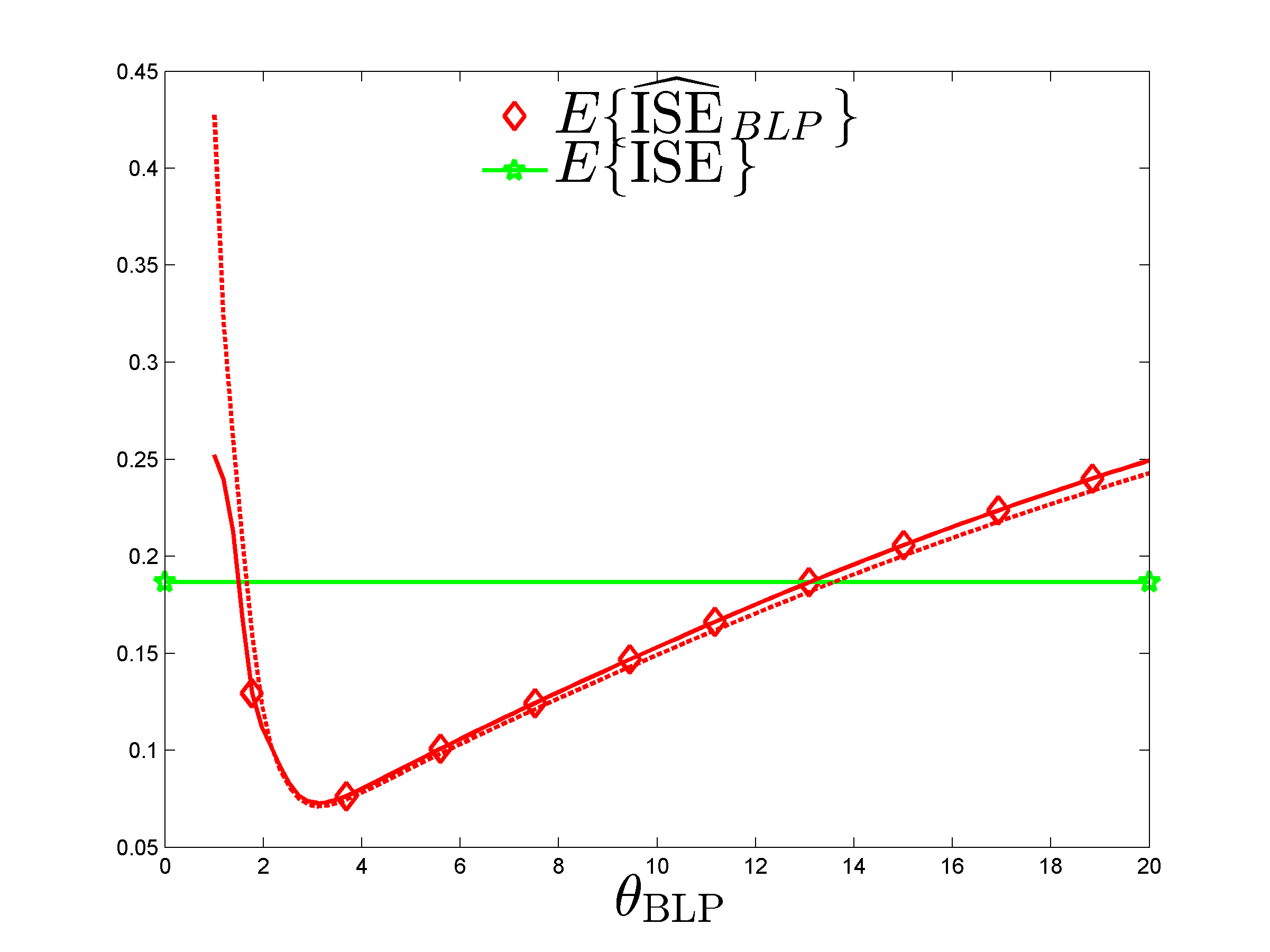}
\includegraphics[width=0.39\textwidth]{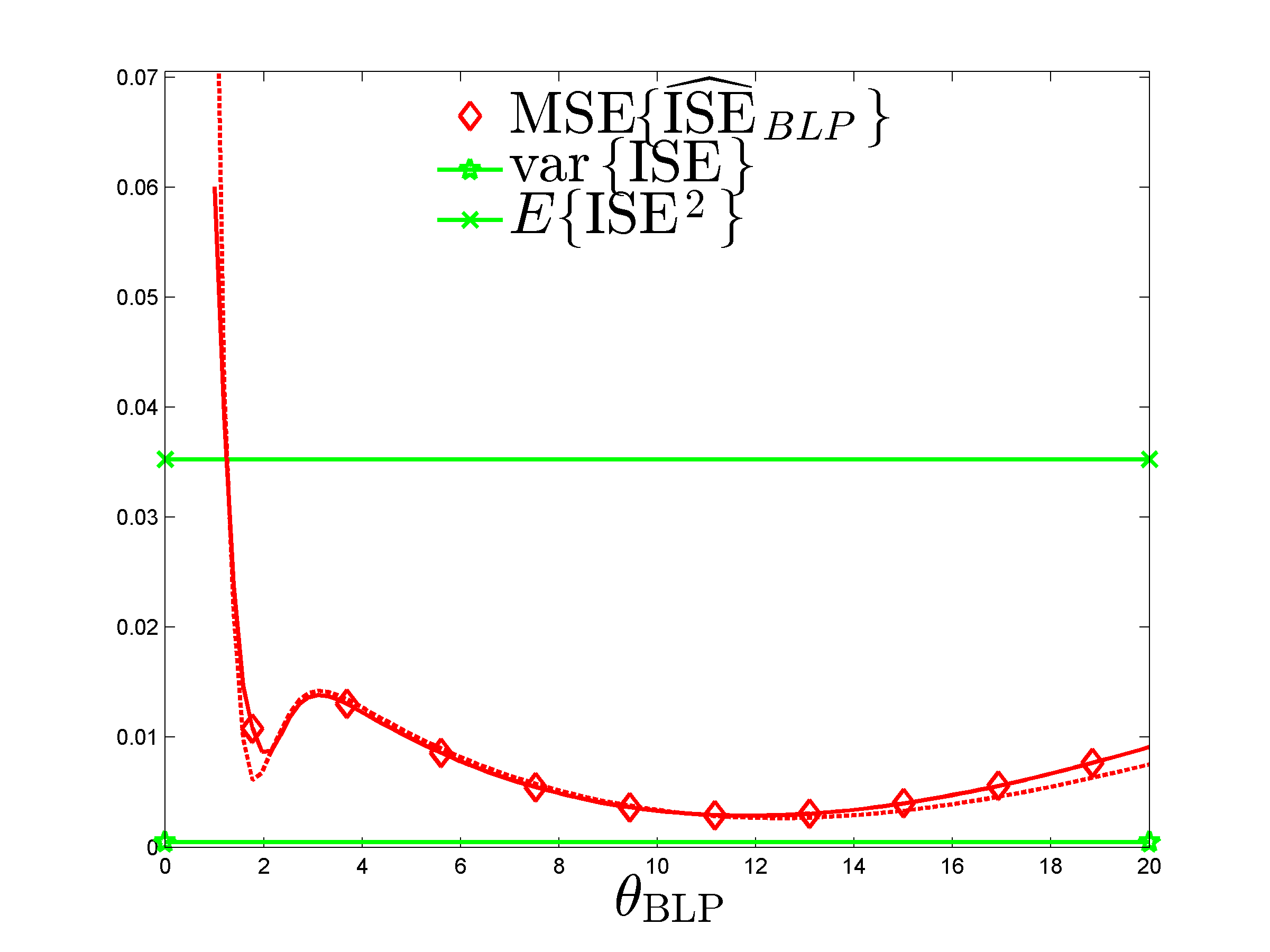}\\
\includegraphics[width=0.39\textwidth]{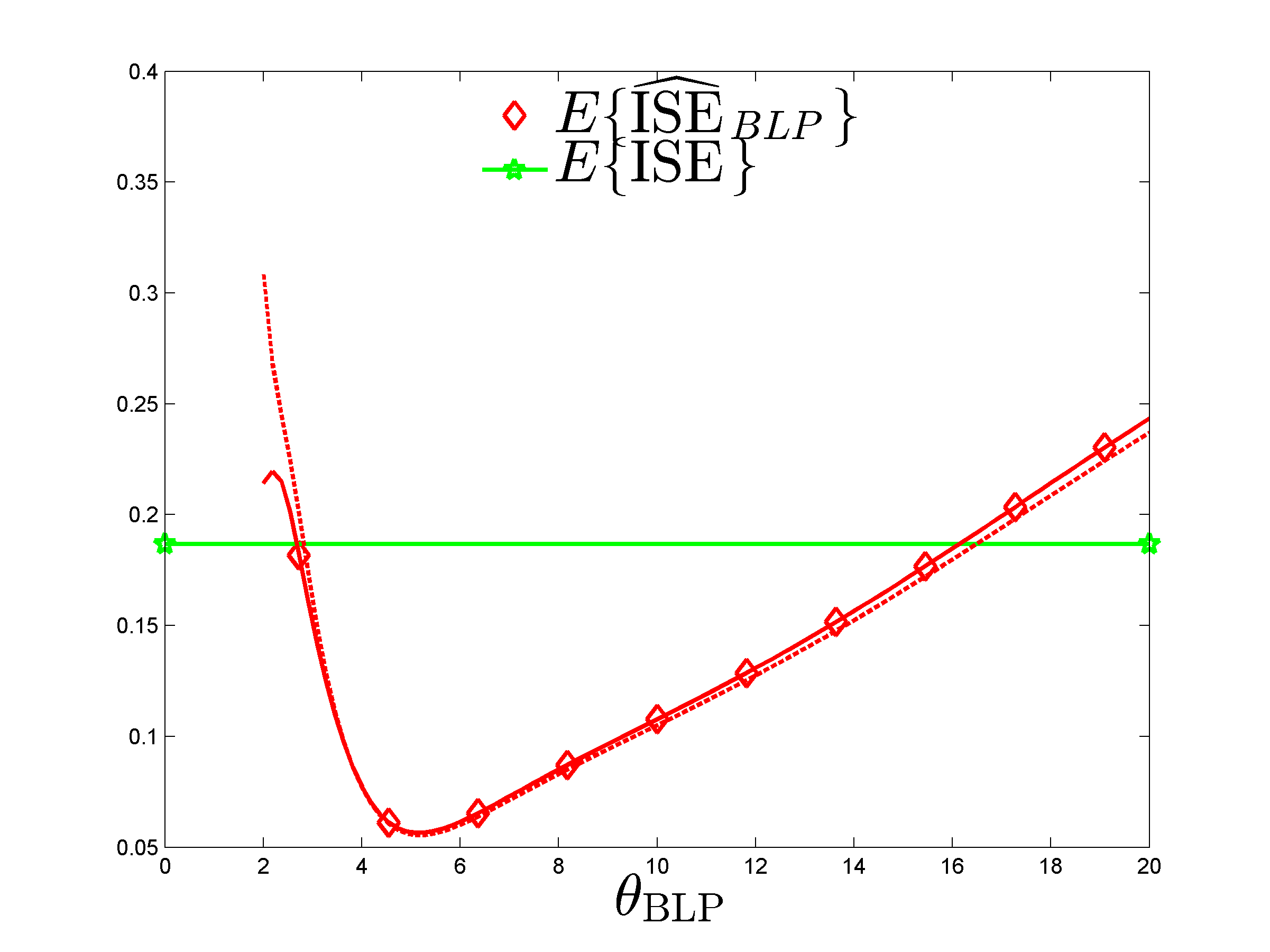}
\includegraphics[width=0.39\textwidth]{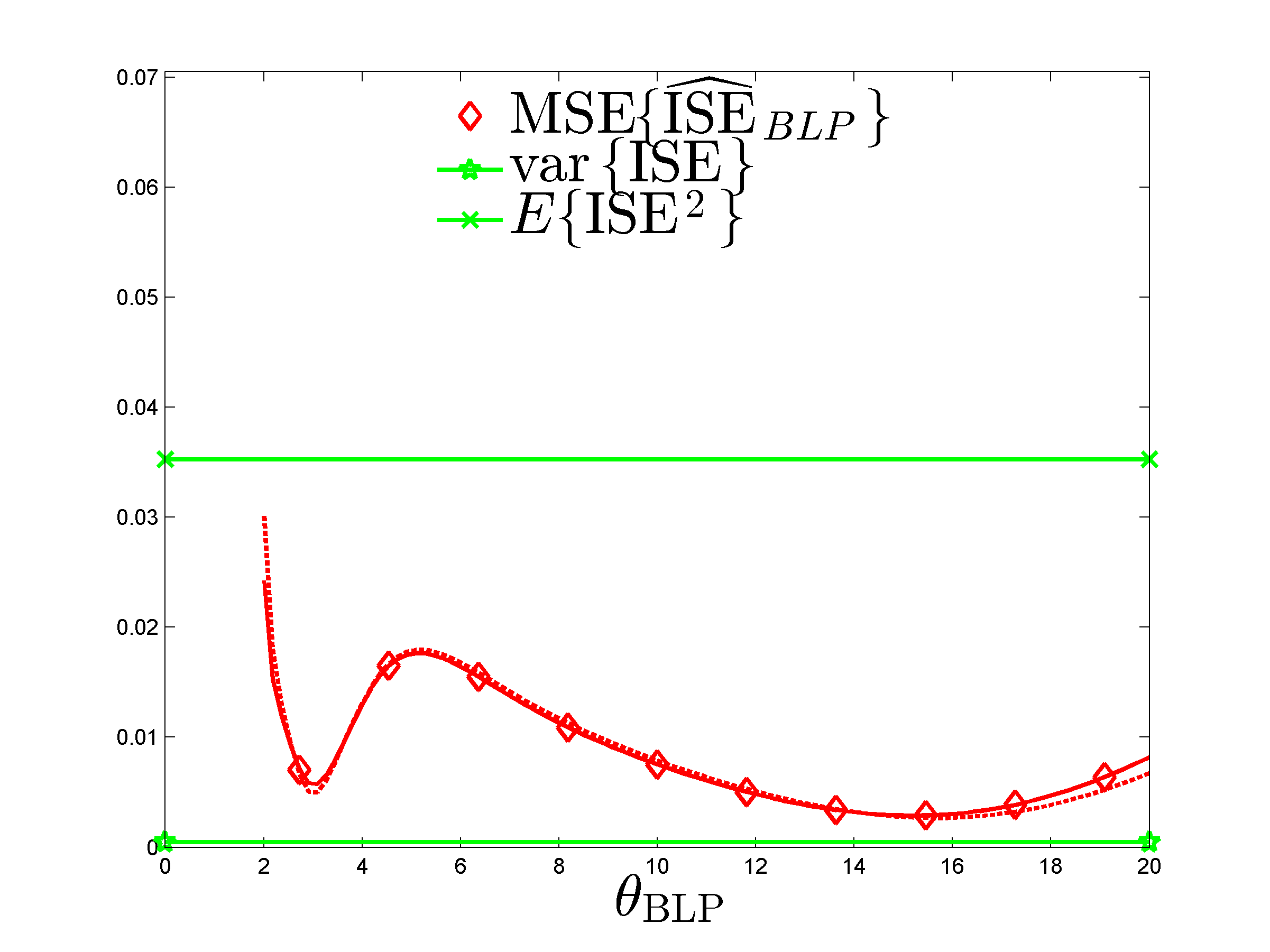}
\caption{\small Performance of $\widehat\ISE_{BLP}(\eta_n)$ when $\eta_n$ is the simple-kriging predictor for the model $\GP(0,K_{5/2,5})$ on $[0,1]^2$; $Y_\xb\sim\GP(0,K_{3/2,10})$, ($\Xb_n$ is a regular grid of 100 design points). From top to bottom: $\psi^{(e)}=\psi_{1/2,\mt_{\rm BLP}}$, $\psi_{5/2,\mt_{\rm BLP}}$, $\psi_{\IM,\mt_{\rm BLP}}$ and $\psi_{\infty,\mt_{\rm BLP}}$.
Empirical values for $100$ repetitions are in dotted lines.} \label{F:ISE_SK=M1-M52-theta5_d2_n100_M0-M32-theta10_M2-M12-M52}
\end{figure}
\end{center}

Figure~\ref{F:ISE_SK=M1-M52-theta5_d2_n100_M0-M32-theta10_M2-M12-M52} shows how
$\widehat\ISE_{BLP}(\eta_n)$ behaves when $\mt_{\rm BLP}$ varies in the four kernels $K^{(e)}$ considered (the behavior for the Mat\'ern 3/2 kernel $K_{3/2,\mt_{\rm BLP}}$ has already been illustrated in Figure~\ref{F:ISE=M1-M52-theta5_d2_n100_M0-M32-theta10_M2-M32}).
Unsurprisingly, the more regular $K^{(e)}$ is, the stronger is the numerical instability for small $\mt_{\rm BLP}$. The independent limits (for $\mt_{\rm BLP}\to +\infty$) are nevertheless practically identical for the four choices of $K^{(e)}$ (see Table~\ref{Tb:indep-lims-ex2}).
The choice of $K^{(e)}$ does not appear to be essential, provided it is regular enough (possibly more regular than $K$) and $\mt_{\rm BLP}$ is not excessively small. For each kernel considered, there is a reasonably large range of values of $\mt_{\rm BLP}$ such that $\widehat\ISE_{BLP}(\eta_n)$ provides an accurate estimate of $\ISE(\eta_n)$ (compare with the values of $\Ex\{\widehat\ISE_{LOO}(\eta_n)\}$ and $\MSE\{\widehat\ISE_{LOO}(\eta_n)\}$ given in Table~\ref{Tb:indep-lims-ex2}), the most stable performance being achieved for $K^{(e)}=K_{5/2,\mt_{\rm BLP}}$.



\section{Average performance of $\hISE_{LOO}(\eta_n)$, $\widehat{\ISE}_{BLP}(\eta_n)$ and $\widehat{\ISE}_{BLUP}(\eta_n)$ for GP realizations with $d\in\{4,6,8\}$}\label{S:behavior-d-n}

In this section we present the values of $\Ex\{\ISE(\eta_n)\}$, $\Ex\{\hISE(\eta_n)\}$ and $\MSE\{\hISE(\eta_n)\}$ for three different ISE estimators, $\hISE_{LOO}(\eta_n)$ \eqref{ISE_{LOO}}, $\hISE_{BLP}(\eta_n)$ \eqref{estimate-ISE-general} and its unbiased version $\hISE_{BLUP}(\eta_n)$ of Section~\ref{S:ISE_BLUP}, when $f$ is the realization of a GP. The design space $\SX$ is the hypercube $[0,1]^d$, with $d\in\{4,6,8\}$, and we consider designs $\Xb_n$ given by the first $n$ points of a scrambled Sobol' sequence in $\SX$, with  $n\in\{10\,d, 20\,d, 50\,d, 100\,d, 200\,d\}$. The measure $\mu$ is uniform on set $\SX_N$ given by the first $N=2^{13+\lfloor d/2 \rfloor}$ Sobol' points in $\SX$.

We suppose that the data are generated with the model $\GP(0,\ms^2 K)$ where $\ms^2=1$ and $K(\xb,\xb')=\psi_{3/2,2}(\|\xb-\xb'\|)$, see \eqref{Matern32}. The predictor $\eta_n$ is the BLUP $\eta_n^*$ for the kernel $K^{(p)}(\xb,\xb')=\psi_{5/2,\mt_p}(\|\xb-\xb'\|)$, see \eqref{psi52} and
$\hISE_{BLP}(\eta_n)$ and $\hISE_{BLUP}(\eta_n)$ assume the model $\GP(0,\ms_e^2,K^{(e)})$ with $K^{(e)}(\xb,\xb')=\psi_{\IM,\mt_{\rm BLP}}(\|\xb-\xb'\|)$, see \eqref{psiIM}. As we do not simulate data (we calculate exact average performance), we cannot estimate $\mt_p$ and $\mt_{\rm BLP}$ from $\yb_n$. We thus adapt their choice to the design, following the suggestion in \cite{PR-Technometrics2016}: $\mt_p$ (respectively, $\mt_{\rm BLP}$) is such that $\psi_{5/2,\mt_p}(D_n)=0.25$ (respectively, $\psi_{\IM,\mt_{\rm BLP}}(D_n)=0.25$), with $D_n=D_n[k]$ the largest of the distances from the $N$ point in $\SX_N$ to their $k$-th nearest neighbor in $\Xb_n$. It ensures that for every point $\xb$ in $\SX_N$ there exist at least $k$ points $\xb_i$ in $\Xb_n$ such that $\psi_{5/2,\mt_p}(\|\xb-\xb_i\|)\geq 0.25$ (respectively, $\psi_{\IM,\mt_{\rm BLP}}(\|\xb-\xb_i\|)\geq 0.25$). This implies that we assume more regularity for smaller designs: it is indeed illusory to pretend to model a highly variable function if $\Xb_n$ is very sparse (see Section~\ref{S:poor-designs} for an illustration). The choice of $k$ is not critical and we use $k=5$.

\begin{center}
\begin{figure}[!h]
\centering
\includegraphics[width=0.45\textwidth]{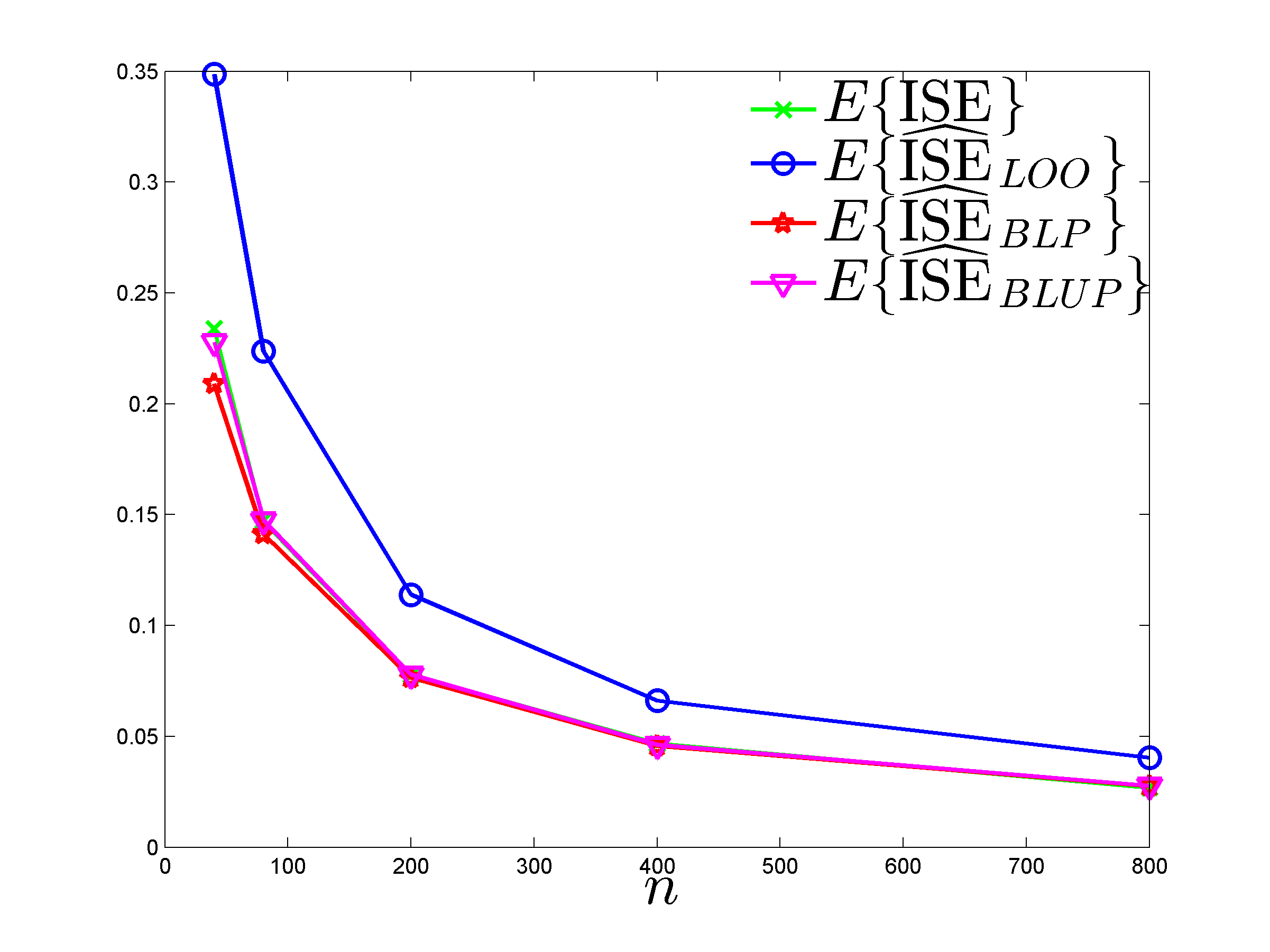}
\includegraphics[width=0.45\textwidth]{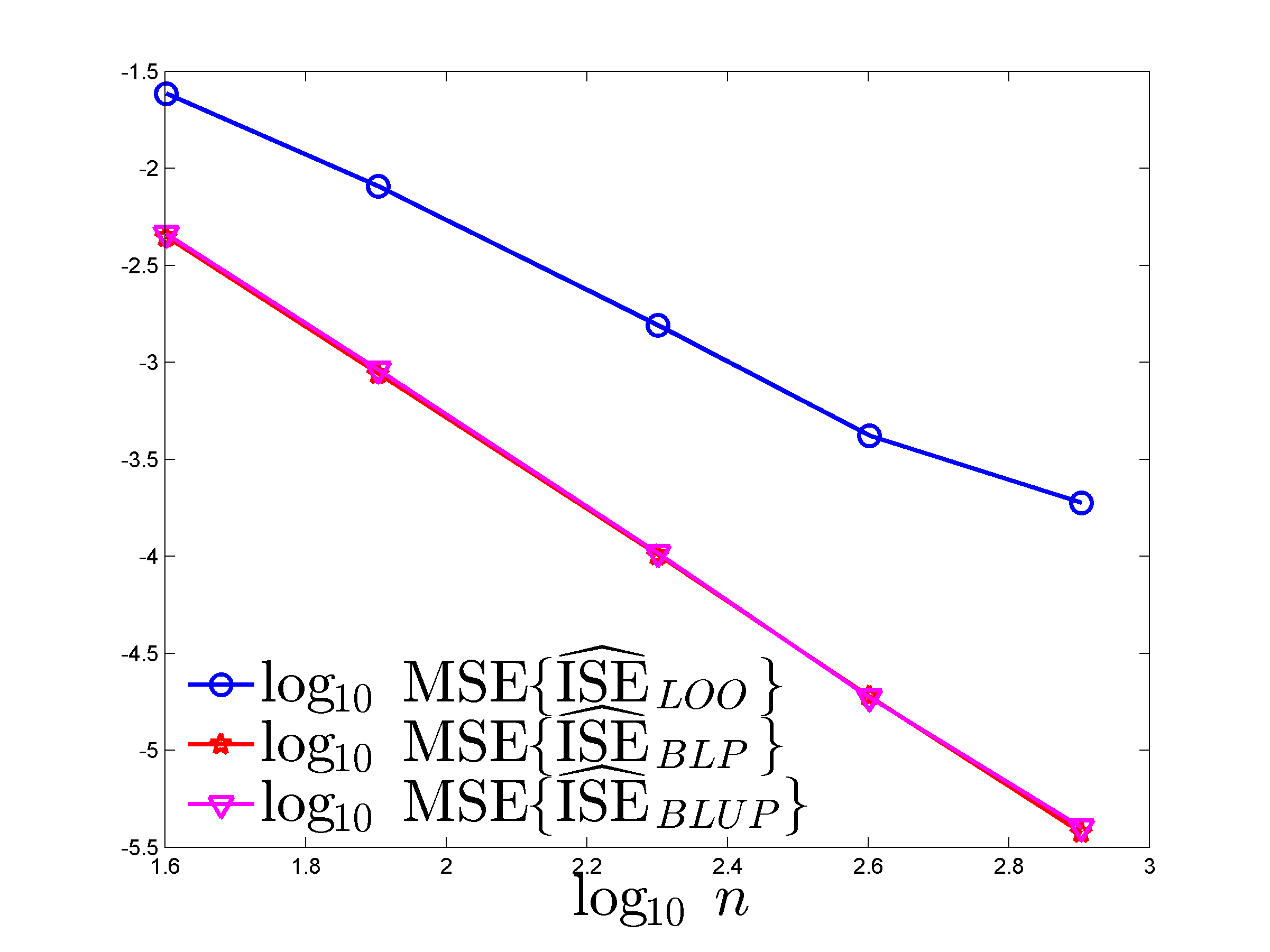} \\
\includegraphics[width=0.45\textwidth]{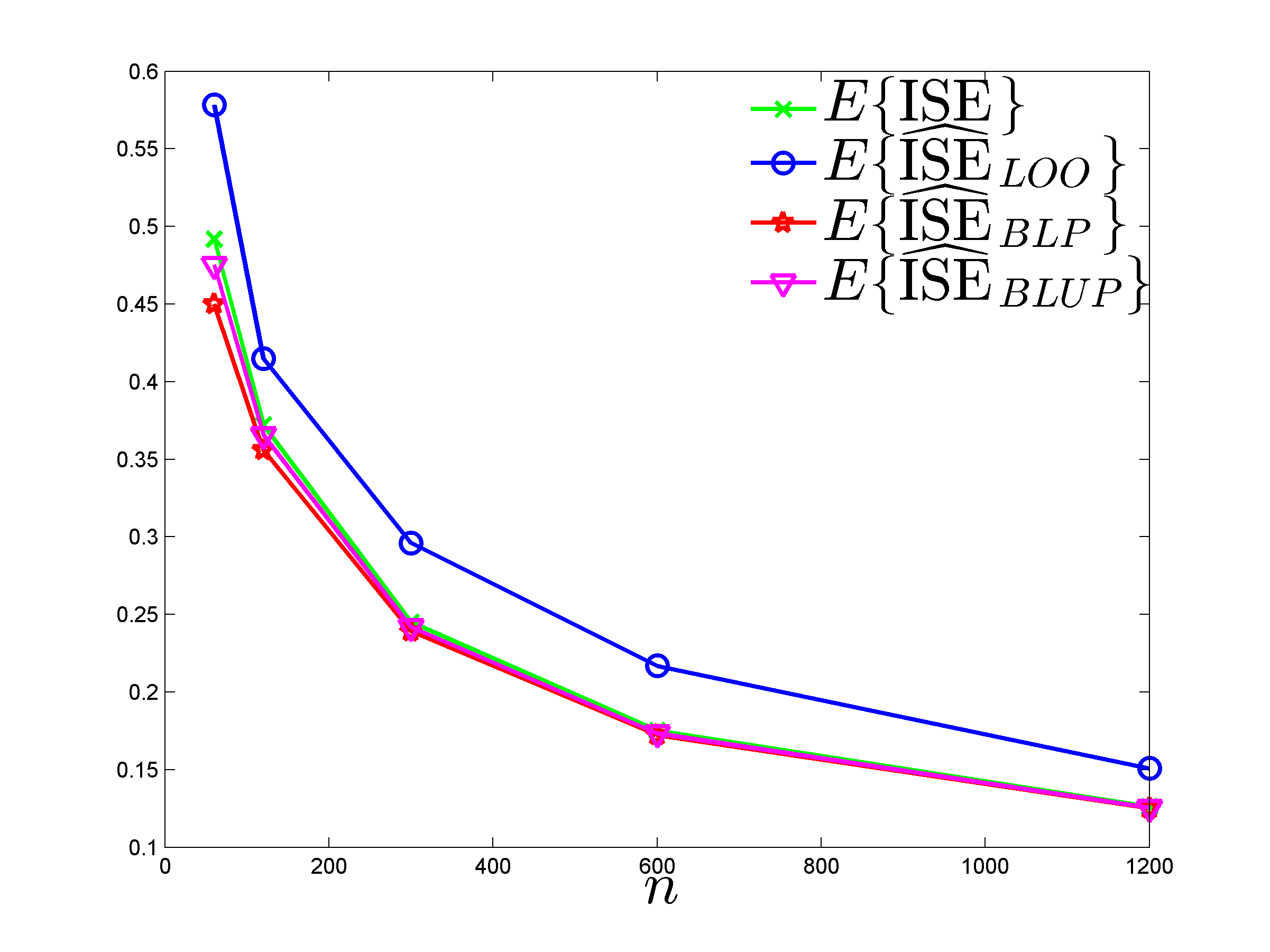}
\includegraphics[width=0.45\textwidth]{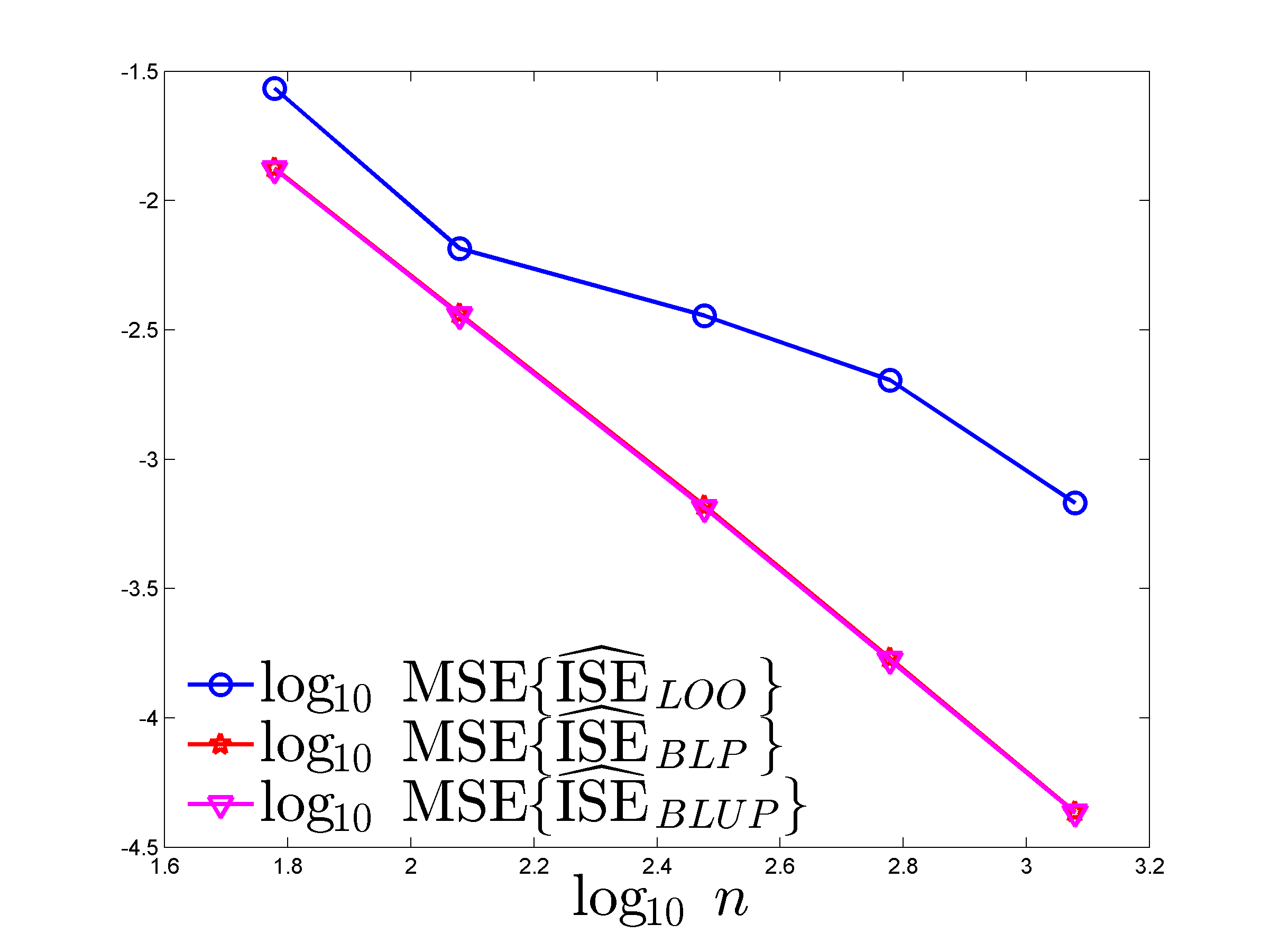} \\
\includegraphics[width=0.45\textwidth]{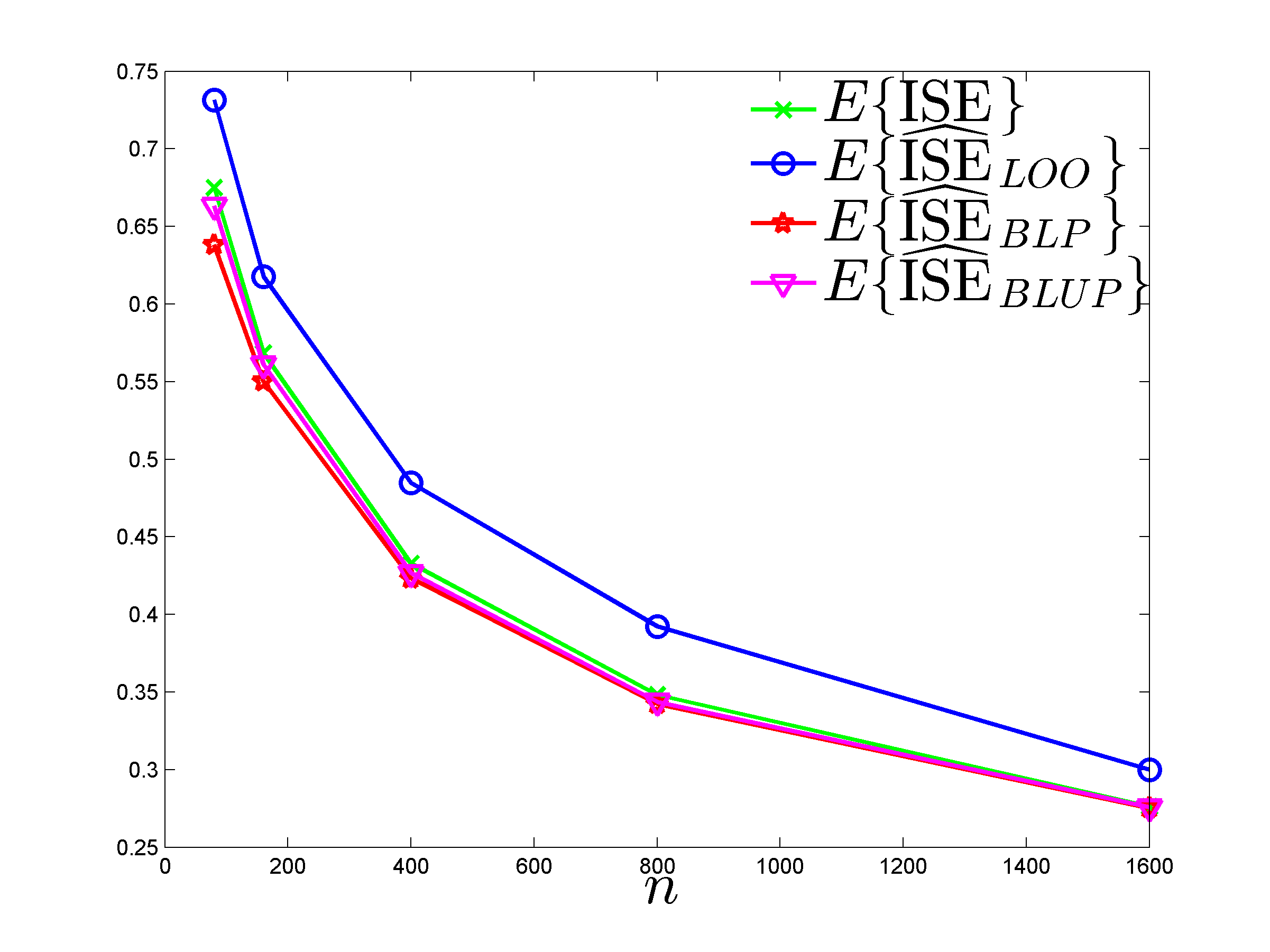}
\includegraphics[width=0.45\textwidth]{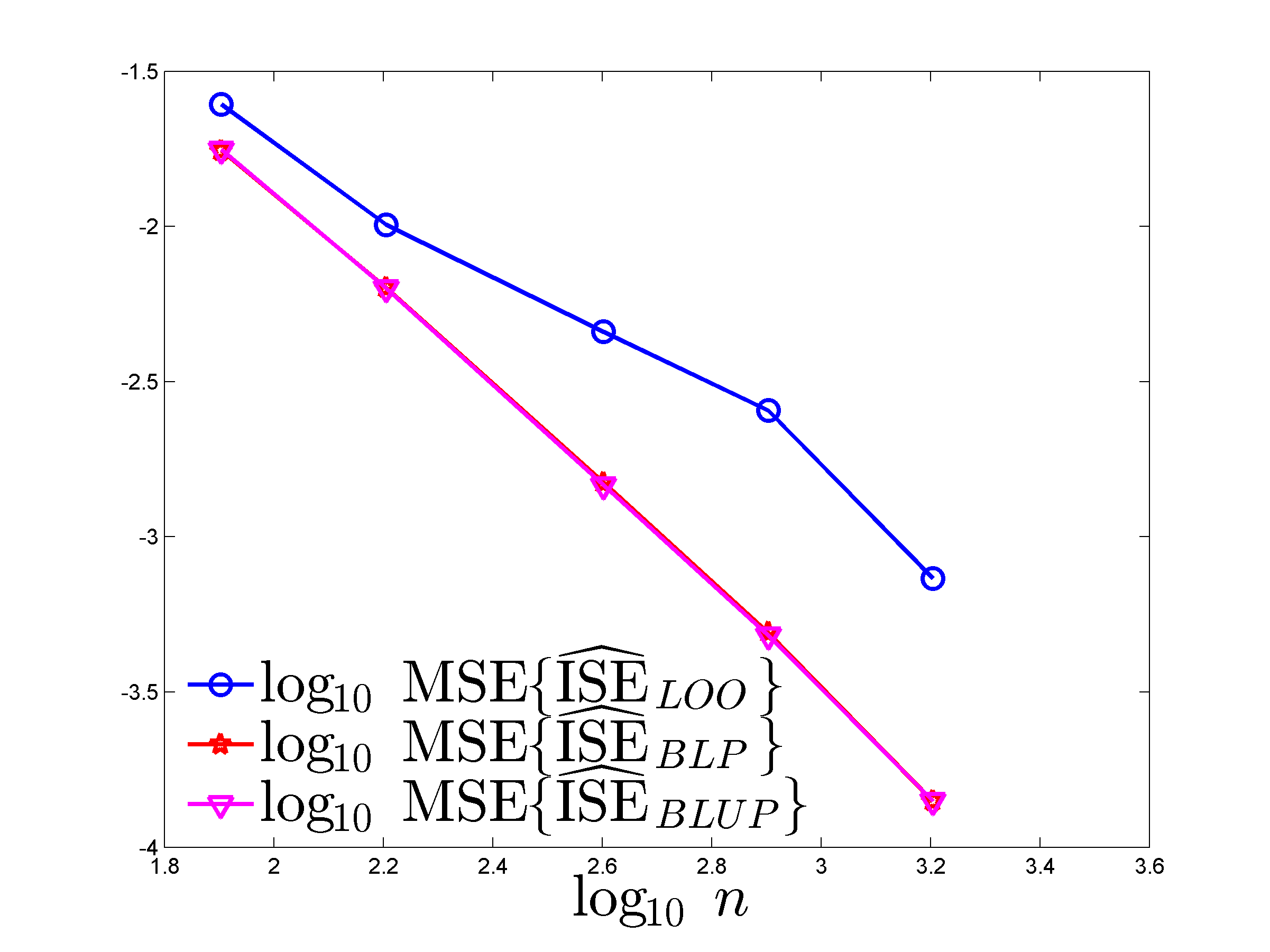}
\caption{\small Performance of $\hISE_{LOO}(\eta_n)$, $\hISE_{BLP}(\eta_n)$ and $\hISE_{BLUP}(\eta_n)$ when $Y_\xb\sim\GP(0,K_{3/2,2})$;
$\eta_n$ is the BLUP for $K^{(p)}(\xb,\xb')=\psi_{5/2,\mt_p}(\|\xb-\xb'\|)$; $\hISE_{BLP}(\eta_n)$ and $\hISE_{BLUP}(\eta_n)$ use the kernel $K^{(e)}(\xb,\xb')=\psi_{\IM,\mt_{\rm BLP}}(\|\xb-\xb'\|)$. From top to bottom: $d=4,6,8$.} \label{F:EISE-d_theta0-2_theta1-adapt_theta2-adapt}
\end{figure}
\end{center}

The left-hand side of Figure~\ref{F:EISE-d_theta0-2_theta1-adapt_theta2-adapt} shows that $\hISE_{BLUP}(\eta_n)$ is slightly closer than $\hISE_{BLP}(\eta_n)$ to $\Ex\{\ISE(\eta_n)\}$ for $n=10\,d$ and $n=20\,d$, but the difference is not visible for larger $n$. On the right-hand side the plots of $\MSE\{\hISE_{BLP}(\eta_n)\}$ and $\MSE\{\hISE_{BLUP}(\eta_n)\}$ are practically confounded.
(As the calculation of $\var\{\ISE(\eta_n)\}$ requires the computation of double integral (a double sum in this example), see \eqref{varISE} and \eqref{Vn}, we omit the term $\var\{\ISE(\eta_n)\}$ is the calculation of $\MSE\{\hISE(\eta_n)\}$, see \eqref{MSE-gamma}: our plots of $\MSE\{\hISE(\eta_n)\}$ thus present in fact $\MSE\{\hISE(\eta_n)\}-\var\{\ISE(\eta_n)\}$.) This suggests that there is little point in using $\hISE_{BLUP}(\eta_n)$ rather than the simpler estimator $\hISE_{BLP}(\eta_n)$. $\hISE_{LOO}(\eta_n)$ overestimates $\ISE(\eta_n)$ is all the cases considered.

\begin{center}
\begin{figure}
\centering
\includegraphics[width=0.49\textwidth]{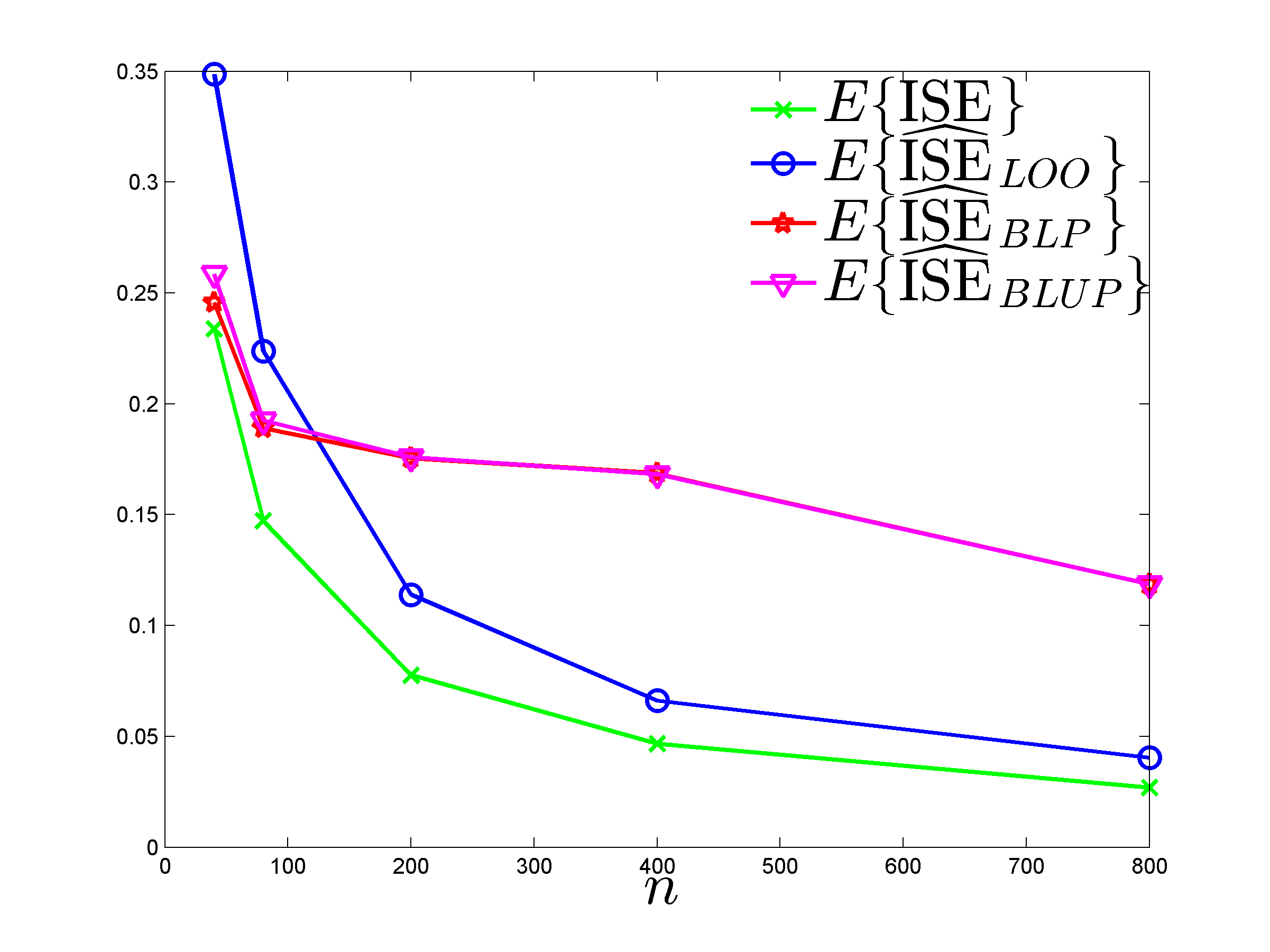}
\includegraphics[width=0.49\textwidth]{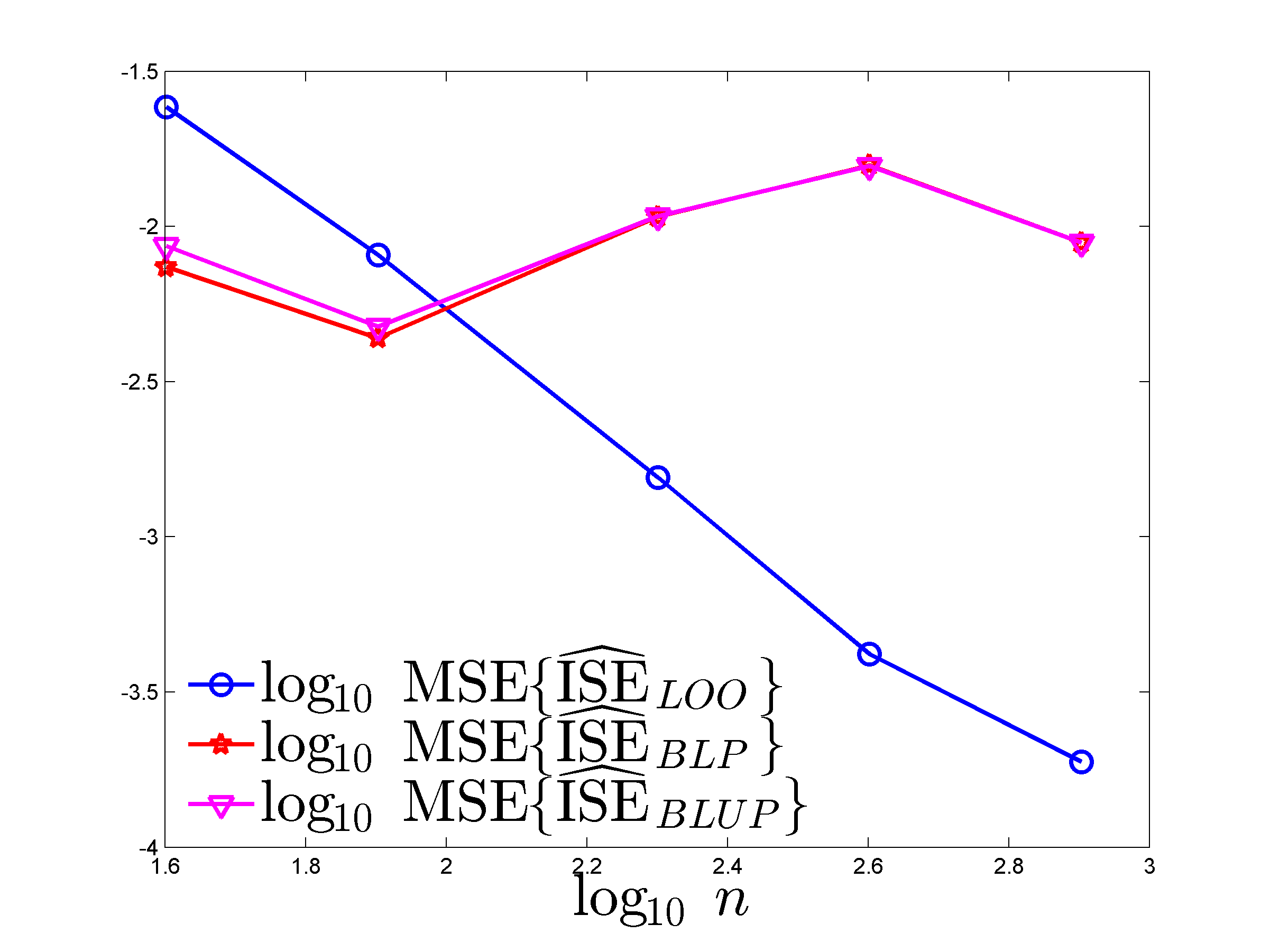} \\
\includegraphics[width=0.49\textwidth]{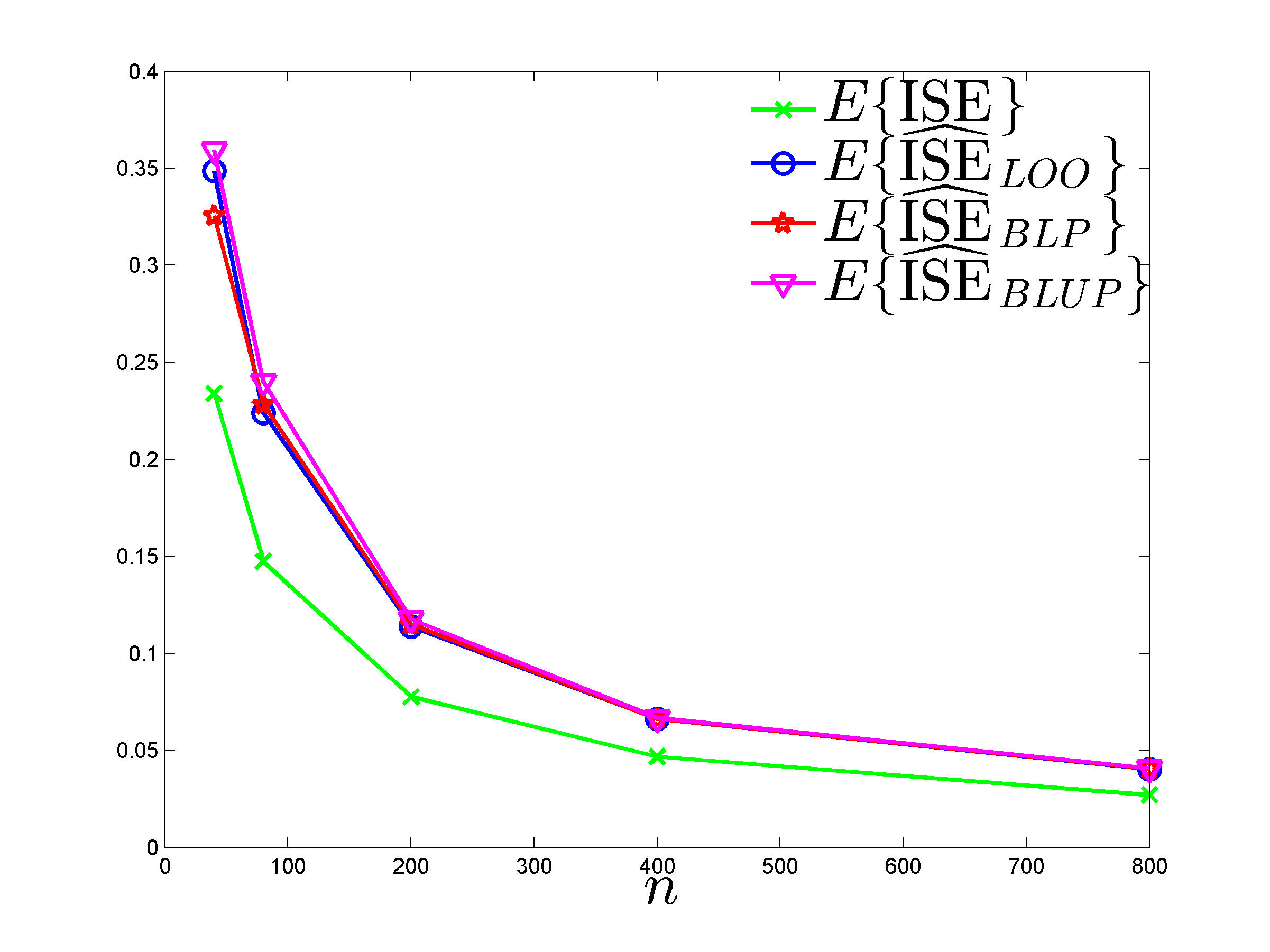}
\includegraphics[width=0.49\textwidth]{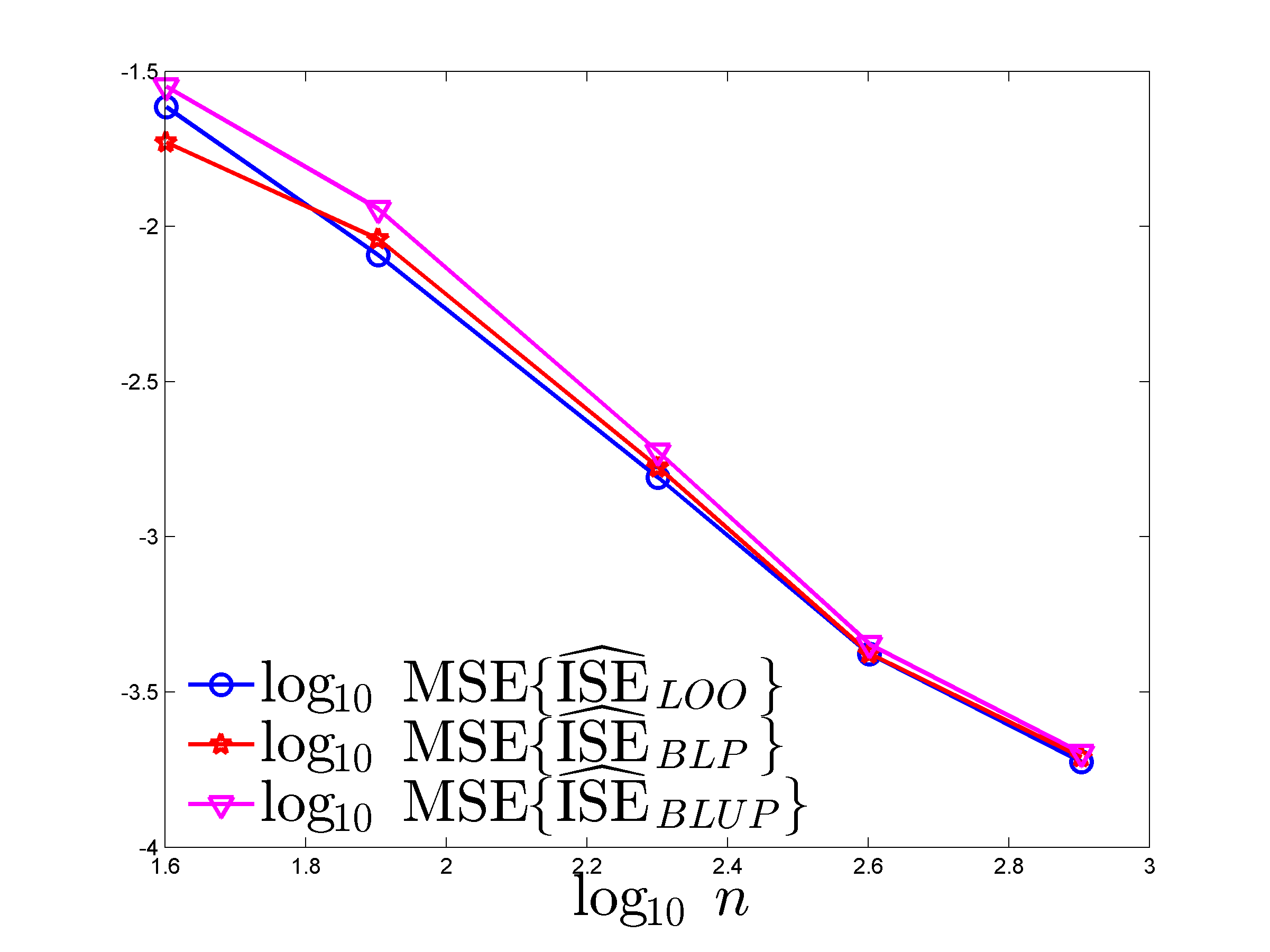}
\caption{\small Same as Figure~\ref{F:EISE-d_theta0-2_theta1-adapt_theta2-adapt} but with $\mt_{\rm BLP}=1$ (top row) and $\mt_{\rm BLP}=20$ (second row): $\hISE_{BLP}(\eta_n)>\hISE_{LOO}(\eta_n)$ is a sign of a poor choice of $\mt_{\rm BLP}$.} \label{F:EISE-d_theta0-2_theta1-adapt_theta2-1}
\end{figure}
\end{center}

To show the importance of an appropriate choice for the kernel $K^{(e)}$, we keep the same $K$ and $K^{(p)}$ as before (with $\mt_p$ thus adapted to the design via the rule $\psi_{5/2,\mt_p}(D_n[5])=0.25$) but use a fixed $\mt_{\rm BLP}$ independently of $\Xb_n$. We first set $\mt_{\rm BLP}=1$ (top row of Figure~\ref{F:EISE-d_theta0-2_theta1-adapt_theta2-1}) with $d=4$: the model used is not flexible enough and $\hISE_{BLP}(\eta_n)$ and $\hISE_{BLUP}(\eta_n)$ severely overestimate $\ISE(\eta_n)$ ($\hISE_{LOO}(\eta_n)$ is the same as in the top row of Figure~\ref{F:EISE-d_theta0-2_theta1-adapt_theta2-adapt}). When $\mt_{\rm BLP}=20$ (second row of Figure~\ref{F:EISE-d_theta0-2_theta1-adapt_theta2-1}), performance deteriorates compared to Figure~\ref{F:EISE-d_theta0-2_theta1-adapt_theta2-adapt}-top but is similar to that of $\hISE_{LOO}(\eta_n)$. A further increase in $\mt_{\rm BLP}$ leads to the independent limit behavior studied in Section~\ref{S:limits}, here with slightly poorer performance than $\hISE_{LOO}(\eta_n)$. A rather general observation is that a value $\hISE_{BLP}(\eta_n)$ larger than $\hISE_{LOO}(\eta_n)$ indicates a bad choice of $K^{(e)}$.

\section{Behavior for random functions that are not GP realizations}\label{S:f-not-GP}

\begin{center}
\begin{figure}[ht]
\centering
\includegraphics[width=0.49\textwidth]{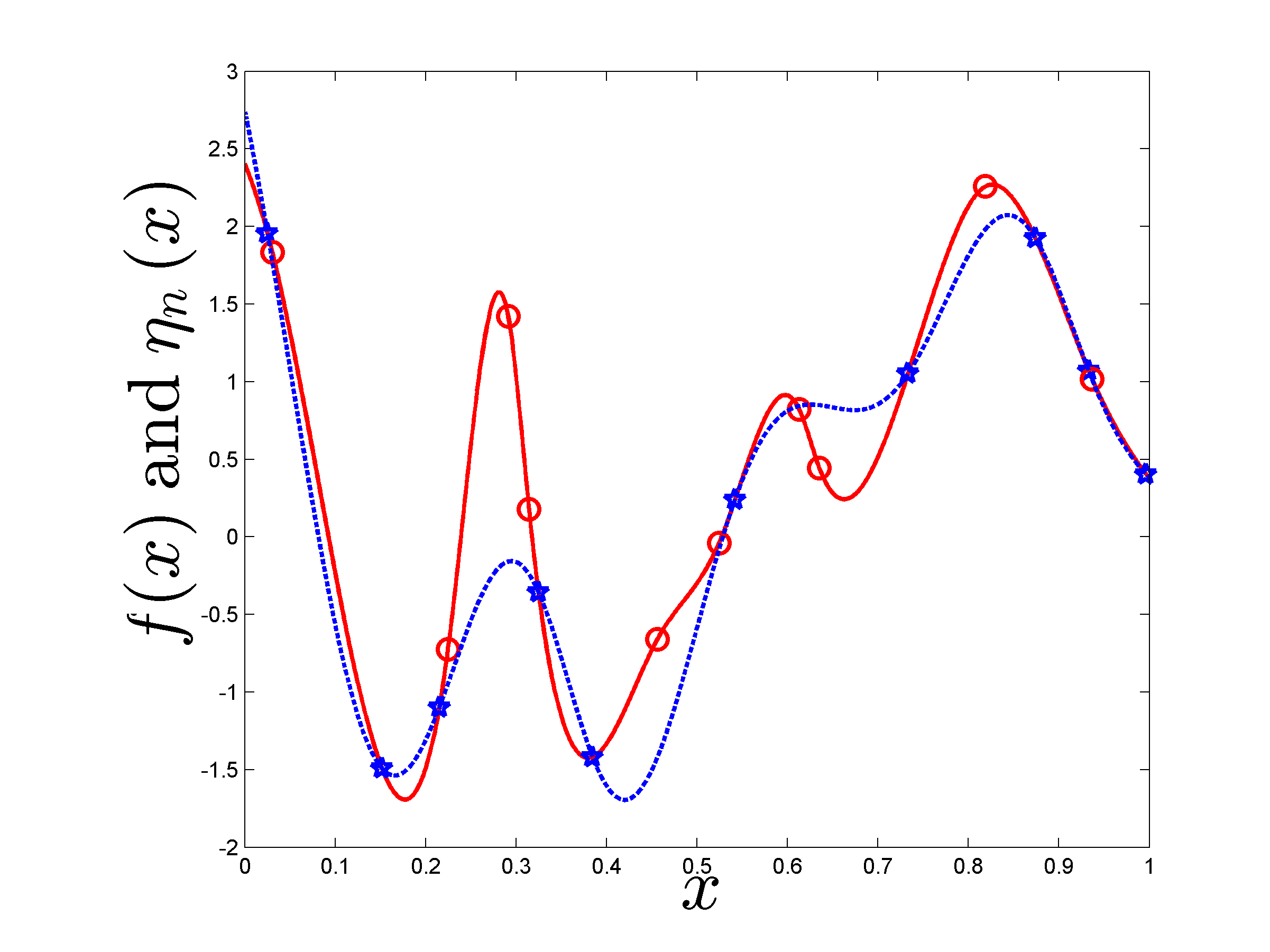}
\includegraphics[width=0.49\textwidth]{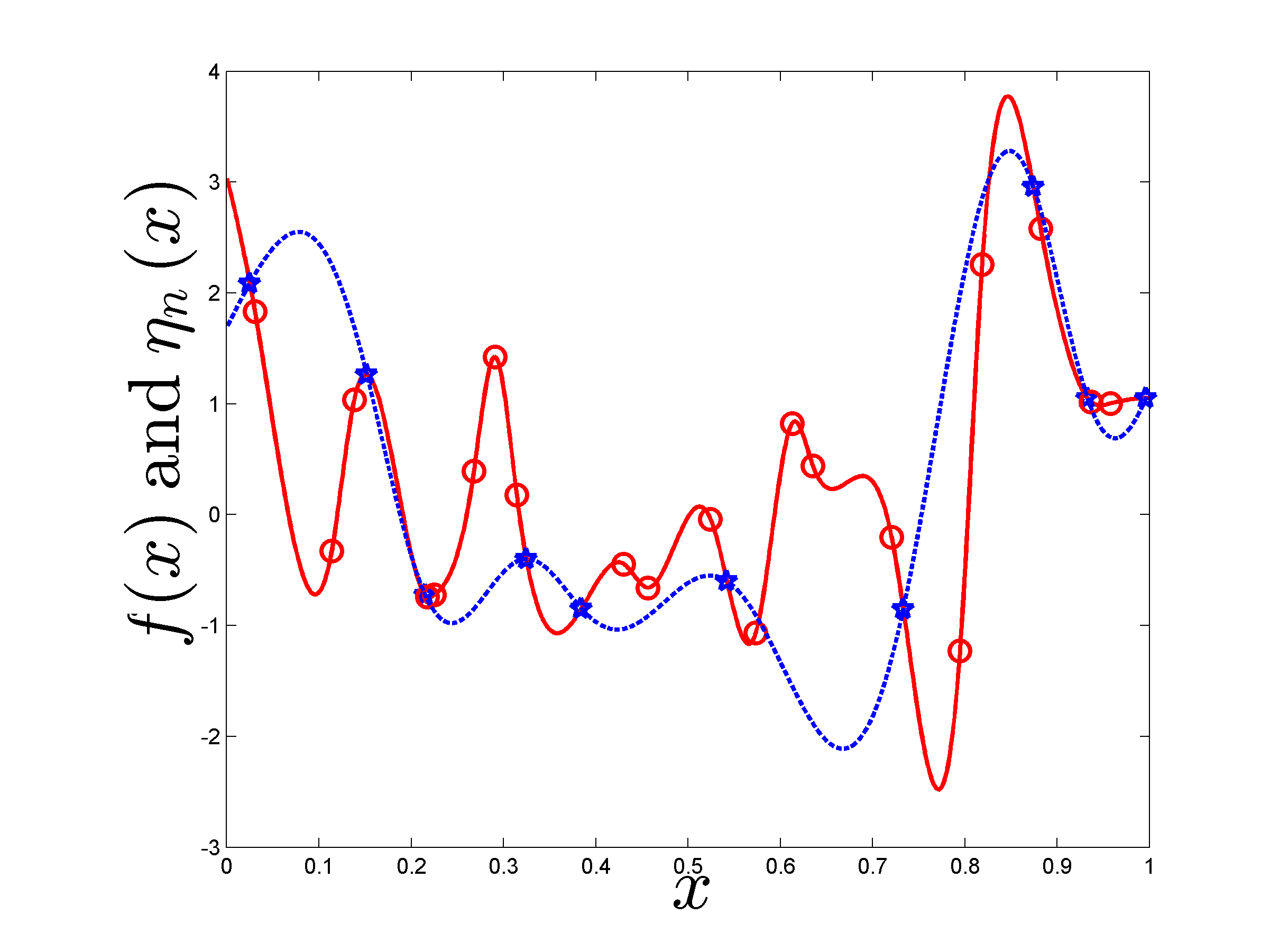}
\caption{\small One realization of a random function $f_m$ ({\color{red} ---}) given by the BLUP for a GP simulation at the design points $\Zb_m$ ({\color{red} $\circ$}); prediction $\eta_n$ of $f_m$ ({\color{blue} $\cdots$}) based on evaluations of $f_m$ at the design points $\Xb_n$ ({\color{blue} $\star$}). Left: $m=n=10$; right: $n=10$, $m=20$ ($\Xb_n$ is identical on both sides).}
\label{F:f-eta-d1_theta0-adapt-theta1-LOO_theta2-LOO}
\end{figure}
\end{center}

The computation of $\ISE(\eta_n)$ requires the evaluation of $f$ on a large set of points $\SX_N$ (we have used $N=2^{10}$ Sobol' points in Sections~\ref{S:numerical-robustness} and \ref{S:different-regularity}), which is restrictive if we want to generate $f$ as the realization of a GP (we need to manipulate $N\times N$ matrices). In this section we follow a different route and (\textit{i}) simulate a GP on a set $\Zb_m$ of small size $m$, then (\textit{ii}) construct $f_m$ as the BLUP, for another GP model, on the design $\Zb_m$. The evaluation of $f_m$ on $\SX_N$ then only involves matrices of size $m\times N$. As Figure~\ref{F:f-eta-d1_theta0-adapt-theta1-LOO_theta2-LOO} illustrates in the case $d=1$, the complexity of the functions $f_m$ can be controlled by the value of $m$.

\subsection{Simulations with various $d$ and $n$}\label{S:simulations-m-n-d}
The design space is always the hypercube $\SX=[0,1]^d$ and the design $\Xb_n$ is given by the first $n$ points of a scrambled Sobol' sequence in $\SX$. As in Section~\ref{S:behavior-d-n}, the measure $\mu$ is uniform on set $\SX_N$ given by the first $N=2^{13+\lfloor d/2 \rfloor}$ Sobol' points in $\SX$. The design $\Zb_m$ corresponds to the first $m$ points of another scrambled Sobol' sequence in $\SX$ ($\Zb_m$ is changed for each simulation of a random function). Data simulation on $\Zb_m$ is with the model $\GP(0,\ms^2 K)$ where $\ms^2=1$ and $K(\xb,\xb')=\psi_{3/2,50}(\|\xb-\xb'\|)$, see \eqref{Matern32}; $f_m$ is the BLUP for $K(\xb,\xb')=\psi_{3/2,\mt^0}(\|\xb-\xb'\|)$ based on the data generated on $\Zb_m$. As in Section~\ref{S:behavior-d-n}, the predictor $\eta_n$ whose ISE we want to estimate is the BLUP for $K^{(p)}(\xb,\xb')=\psi_{5/2,\mt_p}(\|\xb-\xb'\|)$, see \eqref{psi52}, $\hISE_{BLP}(\eta_n)$ and $\hISE_{BLUP}(\eta_n)$ assume the model $\GP(0,\ms_e^2,K^{(e)})$ with $K^{(e)}(\xb,\xb')=\psi_{\IM,\mt_{\rm BLP}}(\|\xb-\xb'\|)$, see \eqref{psiIM}. The value of $\mt^0$ is chosen as in Section~\ref{S:behavior-d-n} and satisfies $\psi_{3/2,\mt^0}(D_n[k])=0.25$ (with $k=5$); $\mt_p$ and $\mt_{\rm BLP}$ are given by the LOO estimates $\widehat\mt_{LOO}$ for the corresponding models: $\mt_p$ minimizes $\hISE_{LOO}(\eta_n^*)$ for the BLUP $\eta_n^*$ associated with the model $\GP(0,\ms^2 K_{5/2,\mt})$ and $\mt_{\rm BLP}$ does the same for the model $\GP(0,\ms^2 K_{\IM,\mt})$. Figure~\ref{F:f-eta-d1_theta0-adapt-theta1-LOO_theta2-LOO} gives an illustration for $d=1$ and shows a realization of $f_m$ with the predictor $\eta_n$ for $n=10$ with $m=10$ (left) and $m=20$ (right).

\begin{center}
\begin{figure}[ht]
\centering
\includegraphics[width=0.49\textwidth]{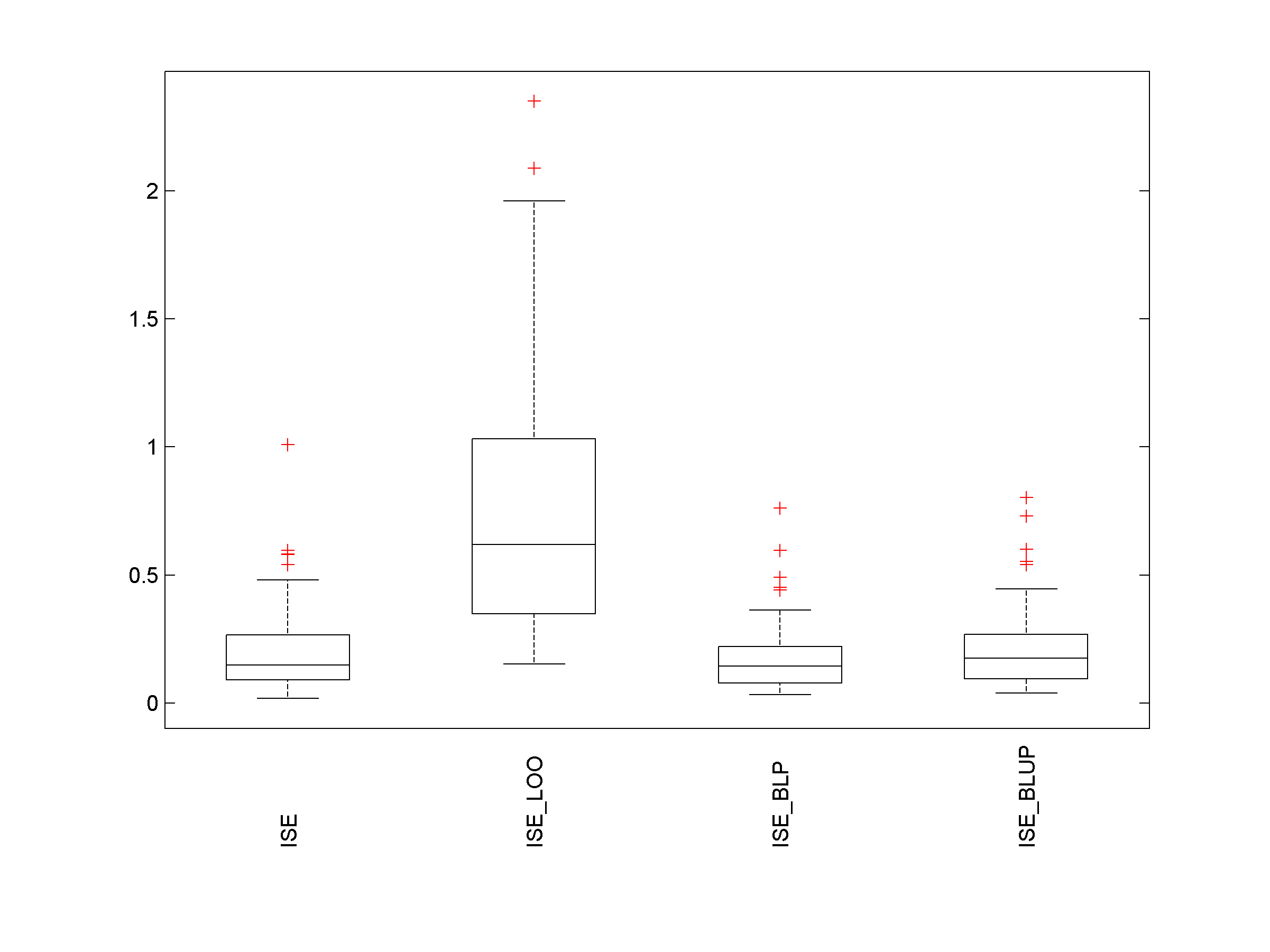}
\includegraphics[width=0.49\textwidth]{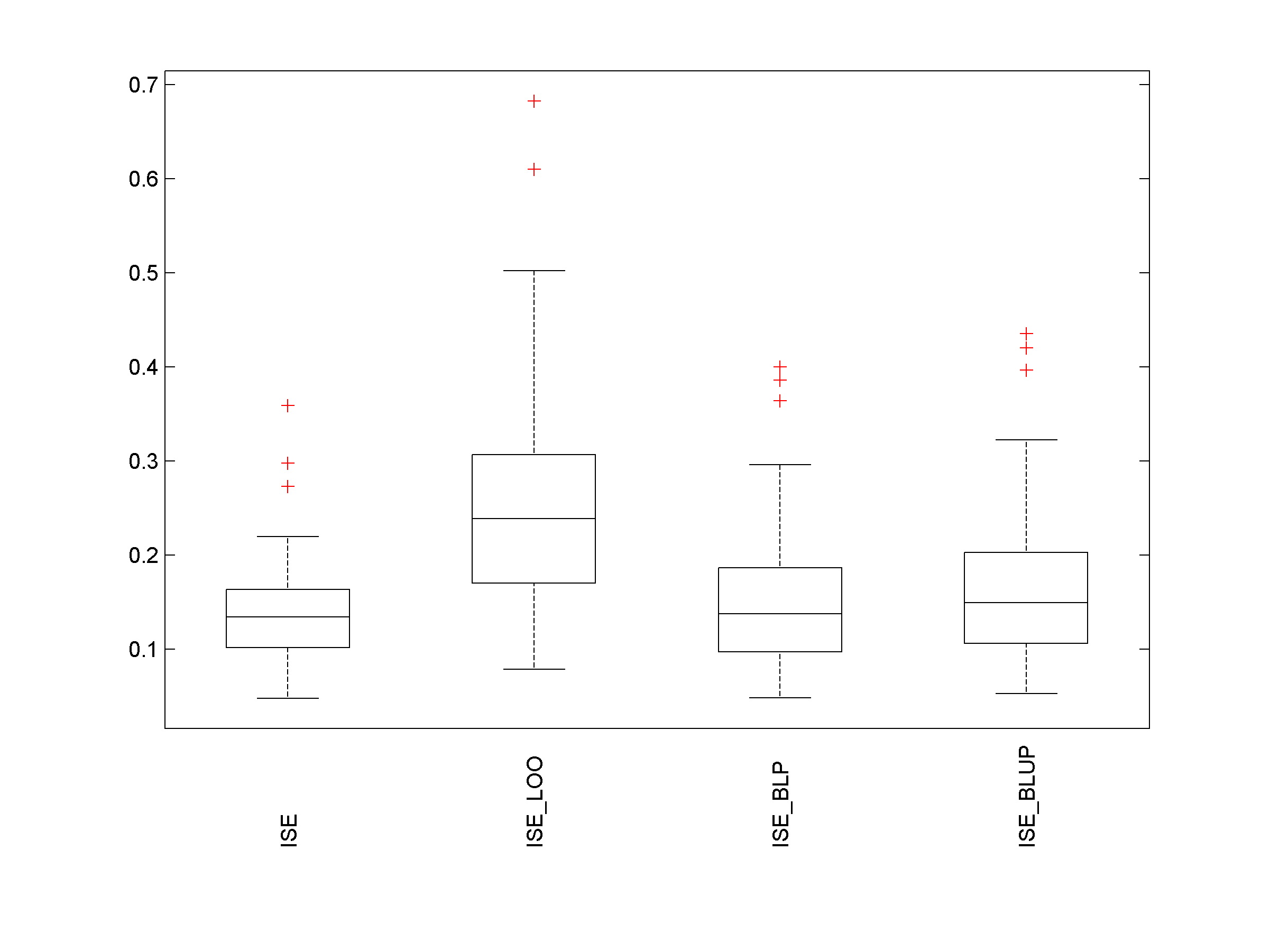} \\
\includegraphics[width=0.49\textwidth]{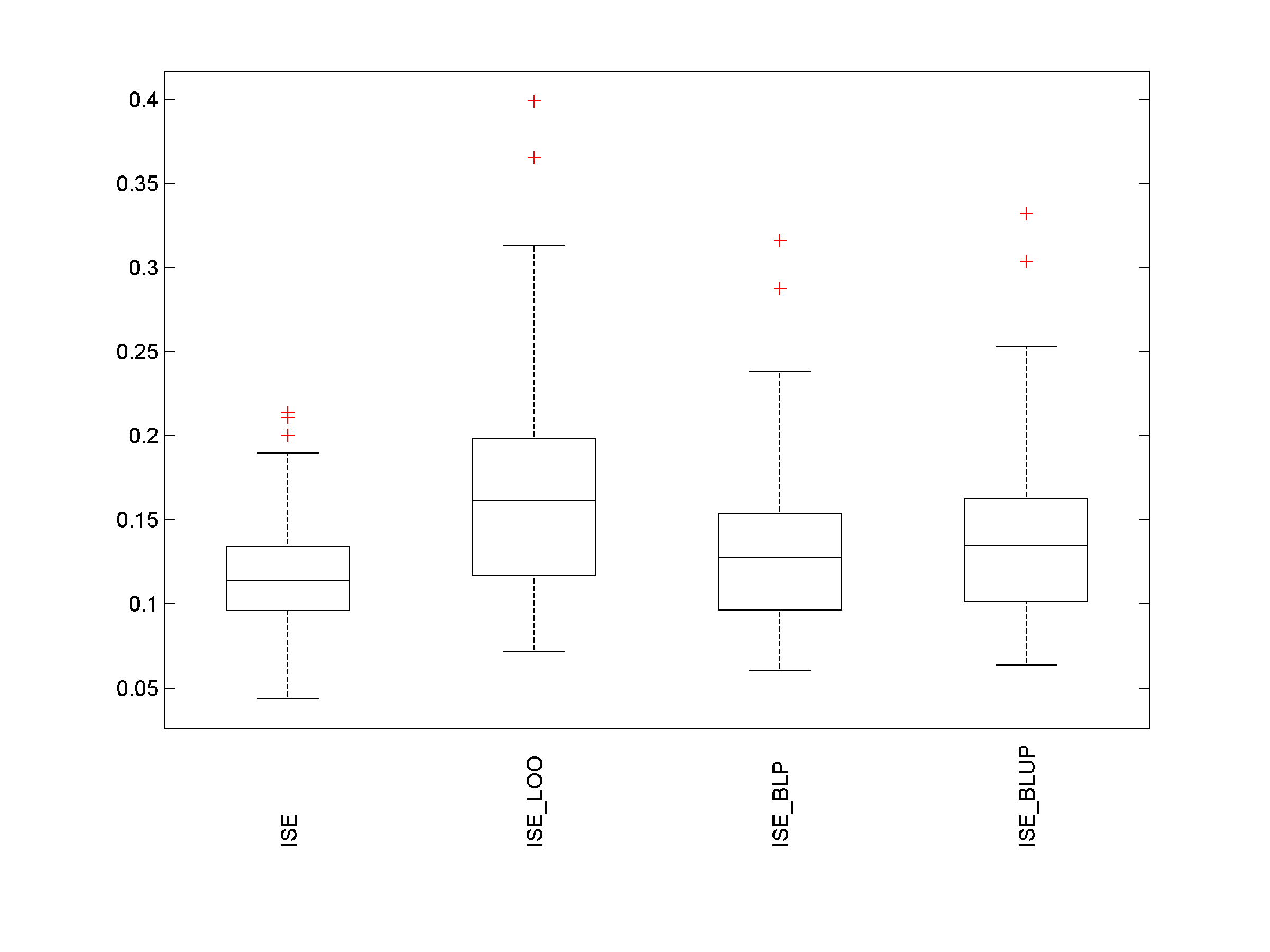}
\includegraphics[width=0.49\textwidth]{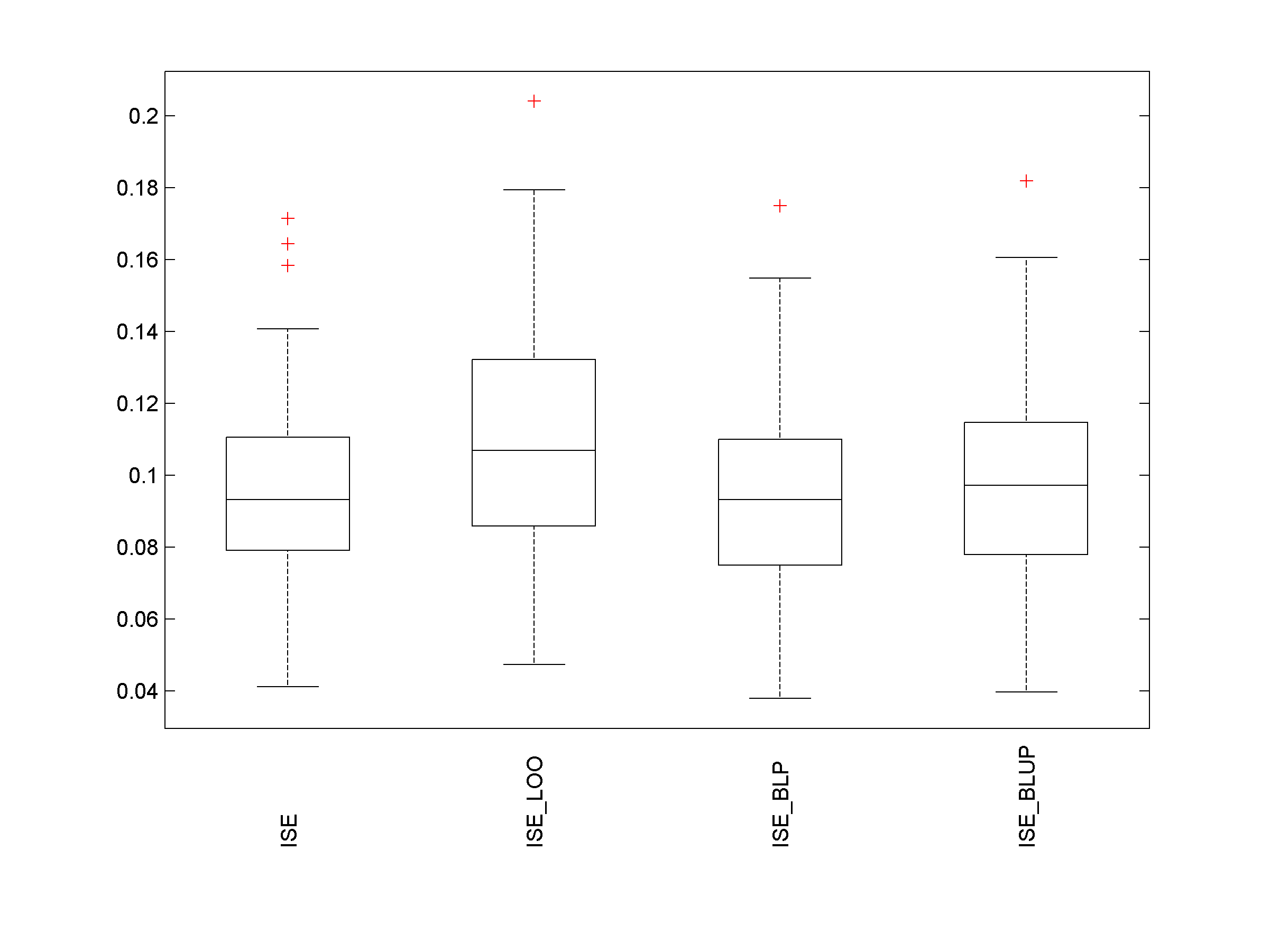}
\caption{\small Boxplots of $\ISE(\eta_n)$, $\hISE_{LOO}(\eta_n)$, $\hISE_{BLP}(\eta_n)$ and $\hISE_{BLUP}(\eta_n)$ for random functions $f_m$ (100 realizations) and Sobol' designs $\Xb_n$ with $m=n=10\,d$. From top to bottom and left to right: $d=2,4,6,8$.} \label{F:hist-theta0-adapt_theta1-LOO_theta2-LOO}
\end{figure}
\end{center}

Figure~\ref{F:hist-theta0-adapt_theta1-LOO_theta2-LOO} presents boxplots of $\ISE(\eta_n)$, $\hISE_{LOO}(\eta_n)$, $\hISE_{BLP}(\eta_n)$ and $\hISE_{BLUP}(\eta_n)$ for different $d$ and (small) designs of size $n=10\,d$ (see \cite{LoeppkySW2009}), obtained from 100 realizations of random $f_m$ generated as indicated above, with $m=n$. Therefore, $m=10\,d$, and the construction used makes $f_m$ easier to approximate as $d$ increases, hence the observation of decreasing values of $\ISE(\eta_n)$ with $d$. For all values of $d$ considered, estimation of $\ISE(\eta_n)$ is more precise with $\hISE_{BLP}(\eta_n)$ and $\hISE_{BLUP}(\eta_n)$ (both behave similarly) than with $\hISE_{LOO}(\eta_n)$, but this superiority tends to vanish as $d$ increases. The complexity of the evaluation of $\hISE_{BLP}(\eta_n)$ is of the order $\SO(N n^3)$, see Section~\ref{S:ISE-any-predictor}, and it grows similarly for $\hISE_{BLUP}(\eta_n)$. For $d=4$, $n=200$ and $N=2^{15}$, the average computational time\footnote{Computations are in Matlab, on a PC with a clock speed of 2.5 GHz and 32 GB RAM.} for the joint evaluations of $\hISE_{BLP}(\eta_n)$ and $\hISE_{BLUP}(\eta_n)$ is about 0.7 s (for 100 repetitions, with standard deviation $\simeq 0.017 $).

\begin{center}
\begin{figure}[ht]
\centering
\includegraphics[width=0.49\textwidth]{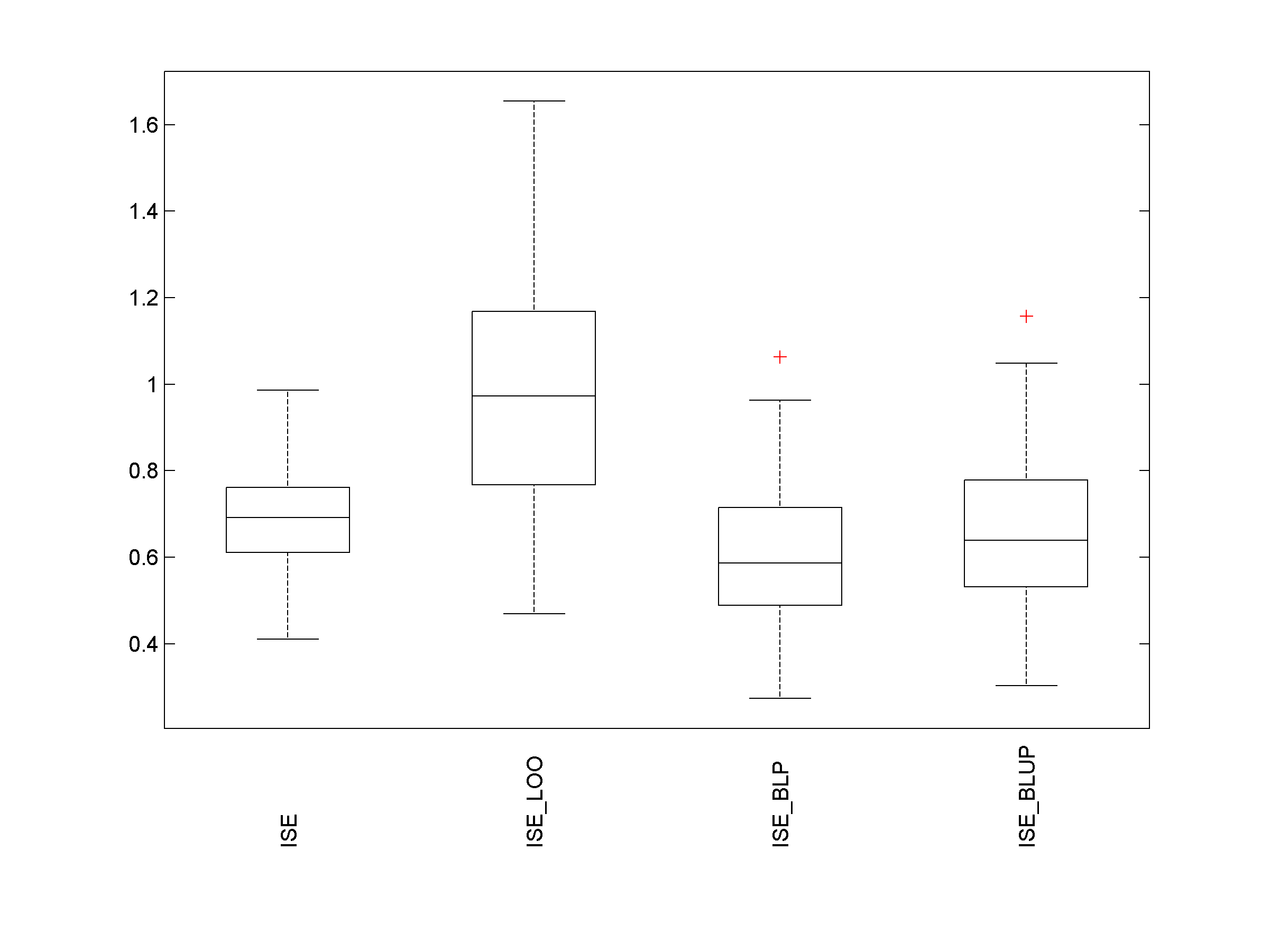}
\includegraphics[width=0.49\textwidth]{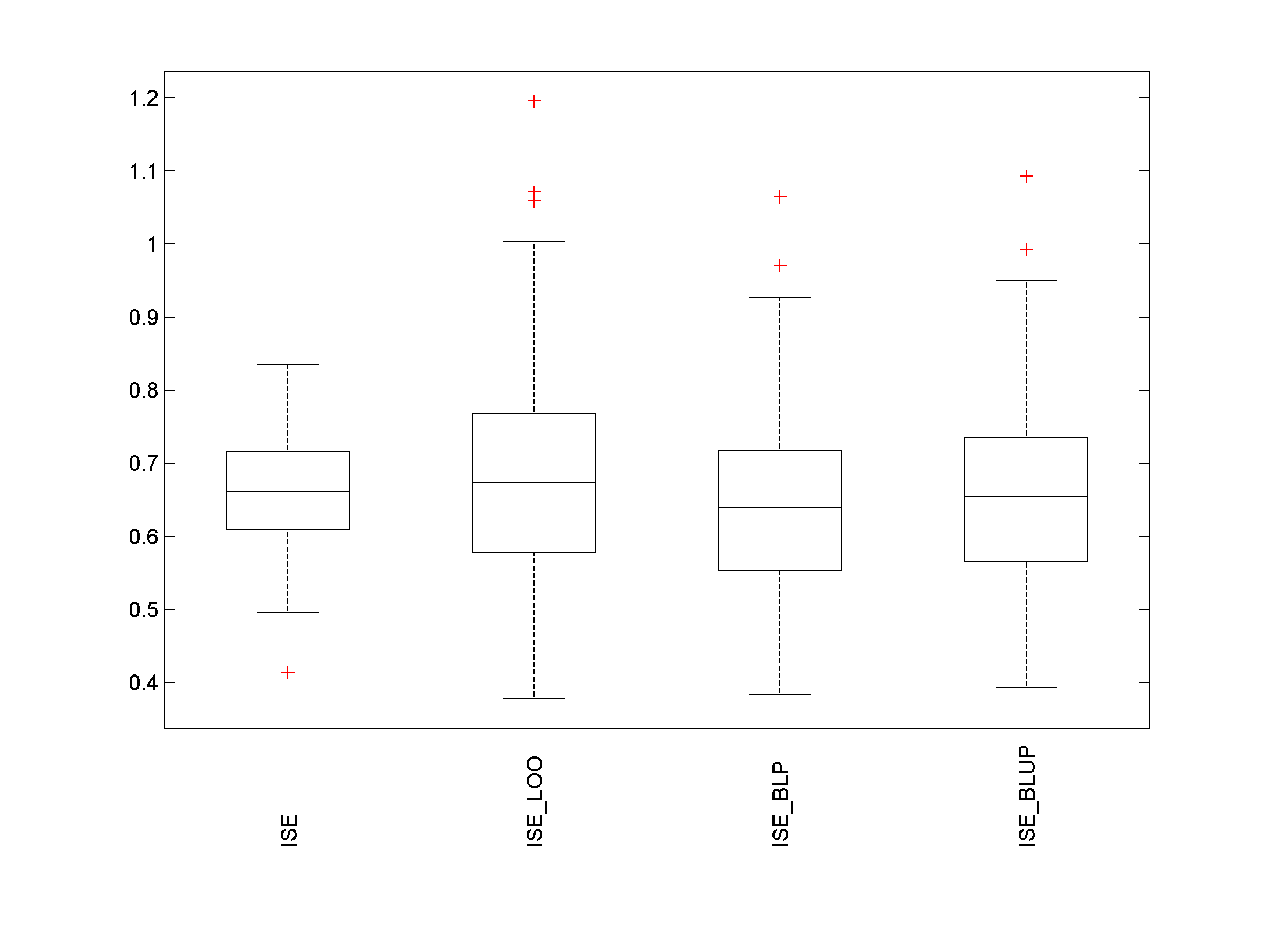}
\caption{\small Same as Figure~\ref{F:hist-theta0-adapt_theta1-LOO_theta2-LOO}-top-right but for random functions $f_m$ with $m=200$ and a Sobol' design $\Xb_n$ with $n=40$ in $[0,1]^4$. Left: $\eta_n$ is the BLUP for $K_{5/2,\mt_p}$ with $\mt_p$ estimated by LOOCV; right: $\eta_n$ is the empirical mean $\overline{\eta_n}=\1b_n\TT\yb_n/n$.} \label{F:hist-d4_theta0-adapt-m10d_theta1-LOO_theta2-LOO}
\end{figure}
\end{center}

We take now $m=5\,n$, making the functions $f_m$ much more complex than above where we had $m=n$. Let us consider the case $d=4$ (with still $n=10\,d$). The same predictor $\eta_n$ (i.e., the BLUP for $K_{5/2,\mt_p}$ with $\mt_p=\widehat\mt_{LOO}$) now performs very poorly: compare the boxplots of $\ISE(\eta_n)$ on the left panel of Figure~\ref{F:hist-d4_theta0-adapt-m10d_theta1-LOO_theta2-LOO} and on the top-right panel of Figure~\ref{F:hist-theta0-adapt_theta1-LOO_theta2-LOO}. In fact, $\eta_n$ performs even worse than the simple empirical mean (i.e., $\overline{\eta_n}=\1b_n\TT\yb_n/n$), whose performance is shown on the right panel of Figure~\ref{F:hist-d4_theta0-adapt-m10d_theta1-LOO_theta2-LOO}. The three ISE estimators $\hISE_{LOO}(\eta_n)$, $\hISE_{BLP}(\eta_n)$ and $\hISE_{BLUP}(\eta_n)$ are capable to uncover this poor performance of $\eta_n$.

\begin{center}
\begin{figure}[ht]
\centering
\includegraphics[width=0.49\textwidth]{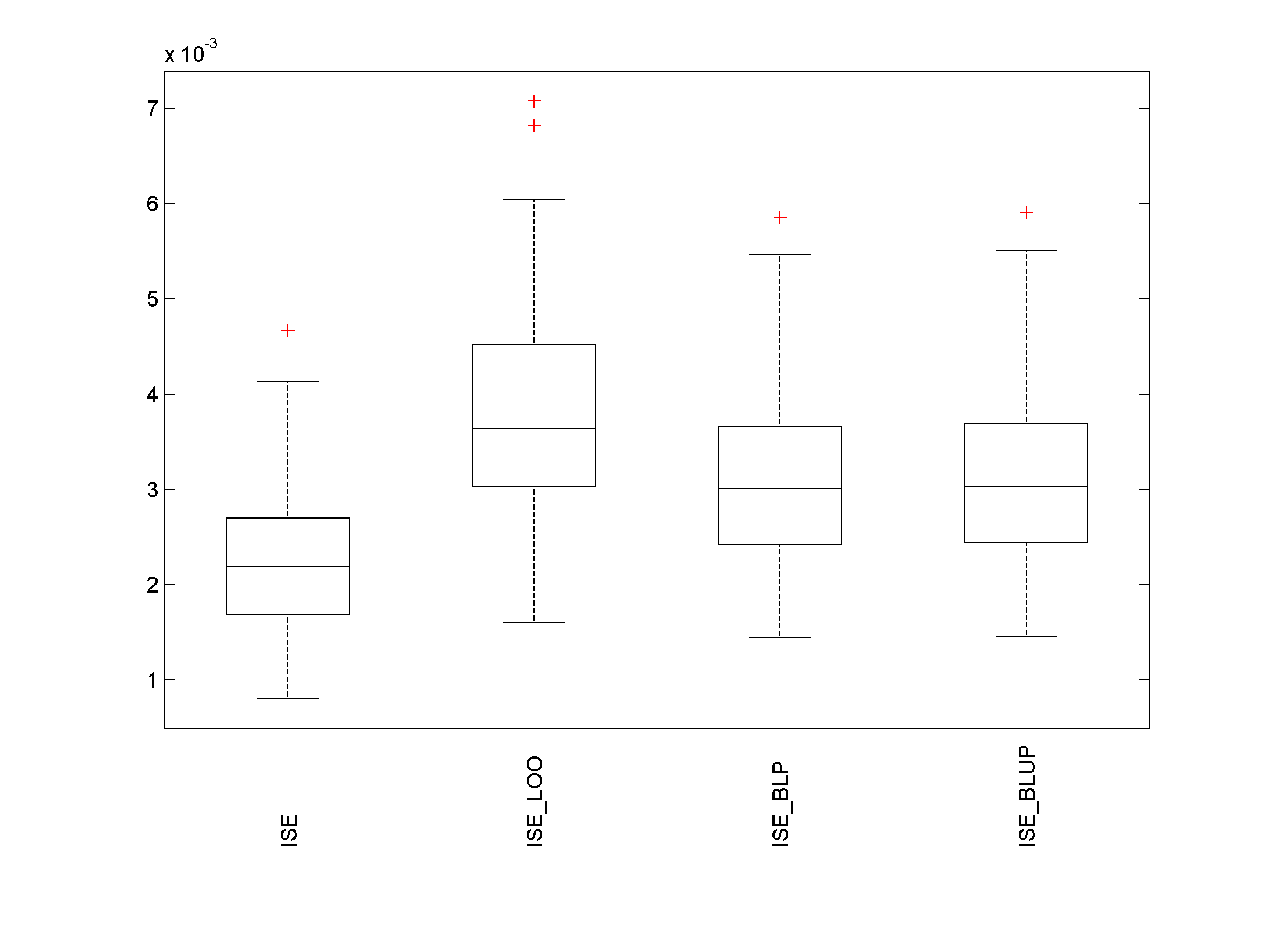}
\includegraphics[width=0.49\textwidth]{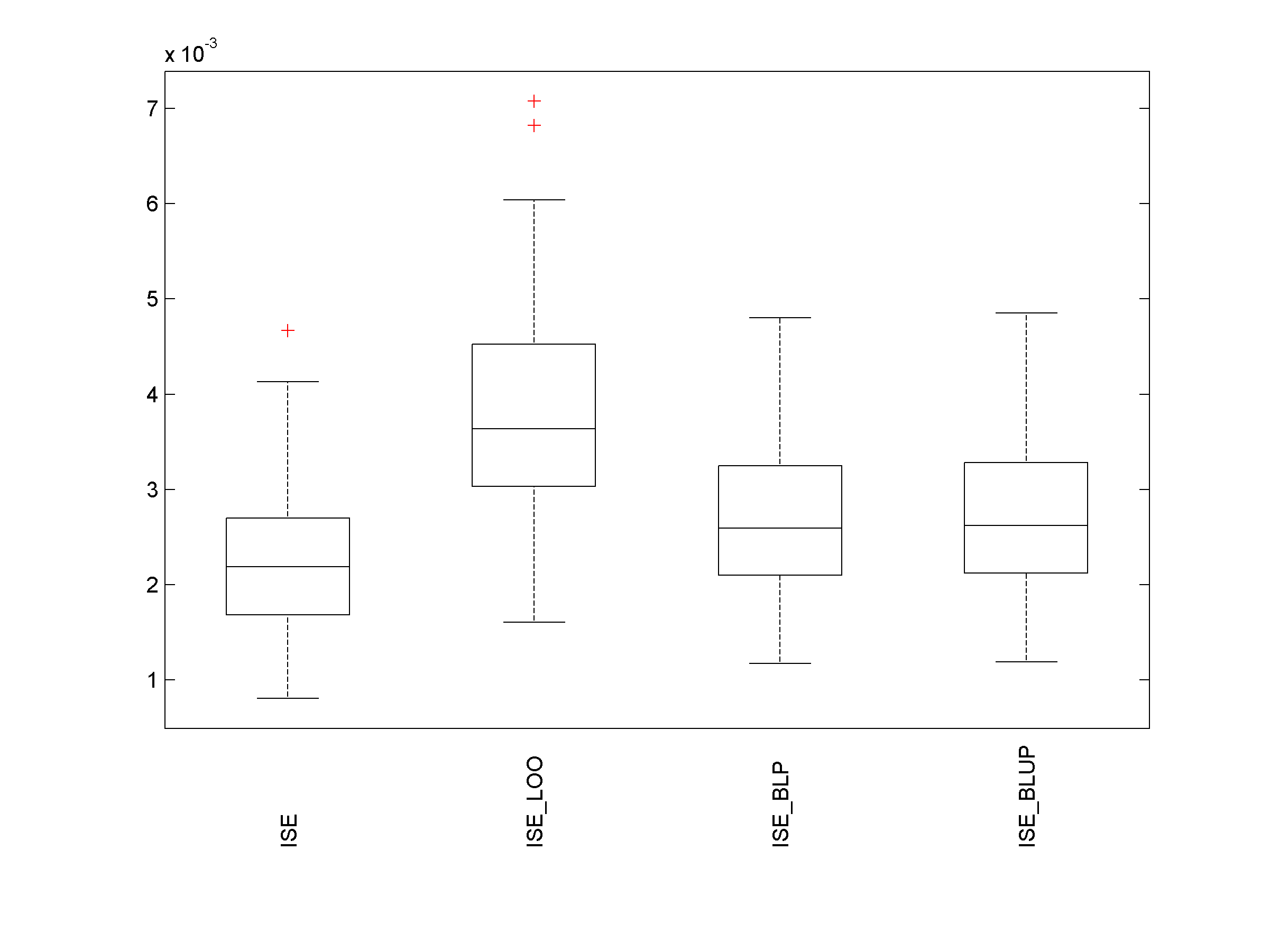}
\caption{\small Same as Figure~\ref{F:hist-theta0-adapt_theta1-LOO_theta2-LOO}-top-right (random functions $f_m$ with $m=40)$ but with $\Xb_n$ a Sobol' design with $n=400$ points in $[0,1]^4$. Left: $\mt_{\rm BLP}=\widehat\mt_{LOO}$ (and $\mt_{\rm BLP}\in(1.55,2.25)$); right: $\mt_{\rm BLP}\simeq 3.8$ satisfies $\psi_{\IM,\mt_{\rm BLP}}(D_n[5])=0.25$. $\ISE(\eta_n)$ and $\hISE_{LOO}(\eta_n)$ are the same on both panels.} \label{F:hist-d4_theta0-adapt_theta1-LOO_theta2-adapt_m40_n400}
\end{figure}
\end{center}

On the contrary, if we keep $m=10\,d$ and increase $n$, $\ISE(\eta_n)$ decreases and is difficult to estimate accurately (in terms of relative precision). Figure~\ref{F:hist-d4_theta0-adapt_theta1-LOO_theta2-adapt_m40_n400} is for $d=4$, $m=40$ and $n=400$. On the left panel, $\mt_{\rm BLP}=\widehat\mt_{LOO}$ (which gives $\mt_{\rm BLP}\in(1.55,2.25)$ with an average value $\simeq 1.94$ for the 100 realizations); on the right panel, $\mt_{\rm BLP}$ is chosen with the rule of Section~\ref{S:behavior-d-n}, i.e., $\psi_{\IM,\mt_{\rm BLP}}(D_n[5])=0.25$ (which gives $\mt_{\rm BLP}\simeq 3.8$). We can notice slightly better performance for $\hISE_{BLP}(\eta_n)$ and $\hISE_{BLUP}(\eta_n)$ on the right-hand panel, but the main observation concerns the low sensitivity to the choice of $K^{(e)}$ and the relevance of the rule of Section~\ref{S:behavior-d-n} (for which, moreover, no numerical optimization with respect to $\mt_{\rm BLP}$ is required).

\subsection{Noisy observations}\label{S:noisy-observations-supp}

We consider the same experimental framework as in Section~\ref{S:simulations-m-n-d} when $d=4$, with $m=n=10\,d=40$, but the observations $\yb_n$ are now given by $y(\xb_i)=f_m(\xb_i)+\zeta_i$, $i=1,\ldots,n$, where the measurement errors $\zeta_i$ are i.i.d.\ $\SN(0,\mg^2)$. Following Section~\ref{S:noisy-observations}, for the construction of $\hISE_{BLP}(\eta_n)$ (and $\hISE_{BLUP}(\eta_n)$) we assume that the data obey the model $Y_\xb \sim \GP(0,\ms^2 K^{'(e)}_{r^{(e)}})$, where
$K^{'(e)}_r(\xb,\xb')=K^{(e)}(\xb,\xb')+r\,\delta_{\xb,\xb'}=\psi_{\IM,\mt_{\rm BLP}}(\|\xb-\xb'\|)+r\, \delta_{\xb,\xb'}$ with $\delta_{\xb,\xb'}=1$ when $\xb=\xb'$ and is zero otherwise. We do not attempt to estimate $r^{(e)}$ from the data, but rather investigate the dependence of performance on the choice of $r^{(e)}$. The construction of $f_m$ is based on GP's with variance $\ms^2=1$ (see Section~\ref{S:simulations-m-n-d}), and we have set a fairly high noise level $\mg=0.25$. The predictor $\eta_n$ is now the BLUP for the kernel $K^{(p)}(\xb,\xb')=\psi_{5/2,\mt_p}(\|\xb-\xb'\|)+\mg^2\,\delta_{\xb,\xb'}$; $\mt_p$ (respectively, $\mt_{\rm BLP}$) is obtained by minimization of $\hISE_{LOO}(\eta_n^*)$, with $\eta_n^*$ the BLUP for $\GP(0,K^{(p)})$ (respectively, for $\GP(0,K^{'(e)}_r)$).

\begin{center}
\begin{figure}[ht]
\centering
\includegraphics[width=0.49\textwidth]{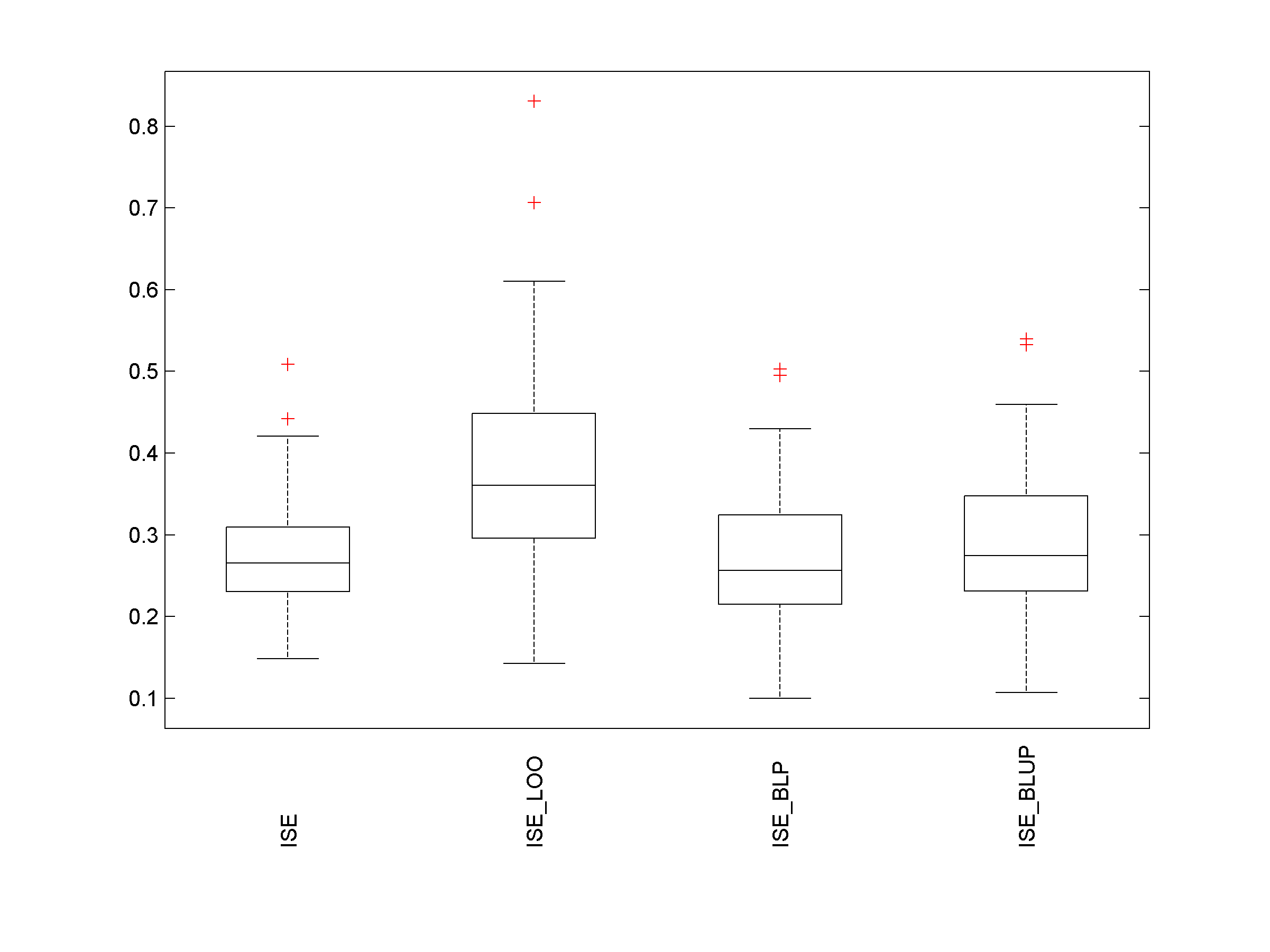}
\includegraphics[width=0.49\textwidth]{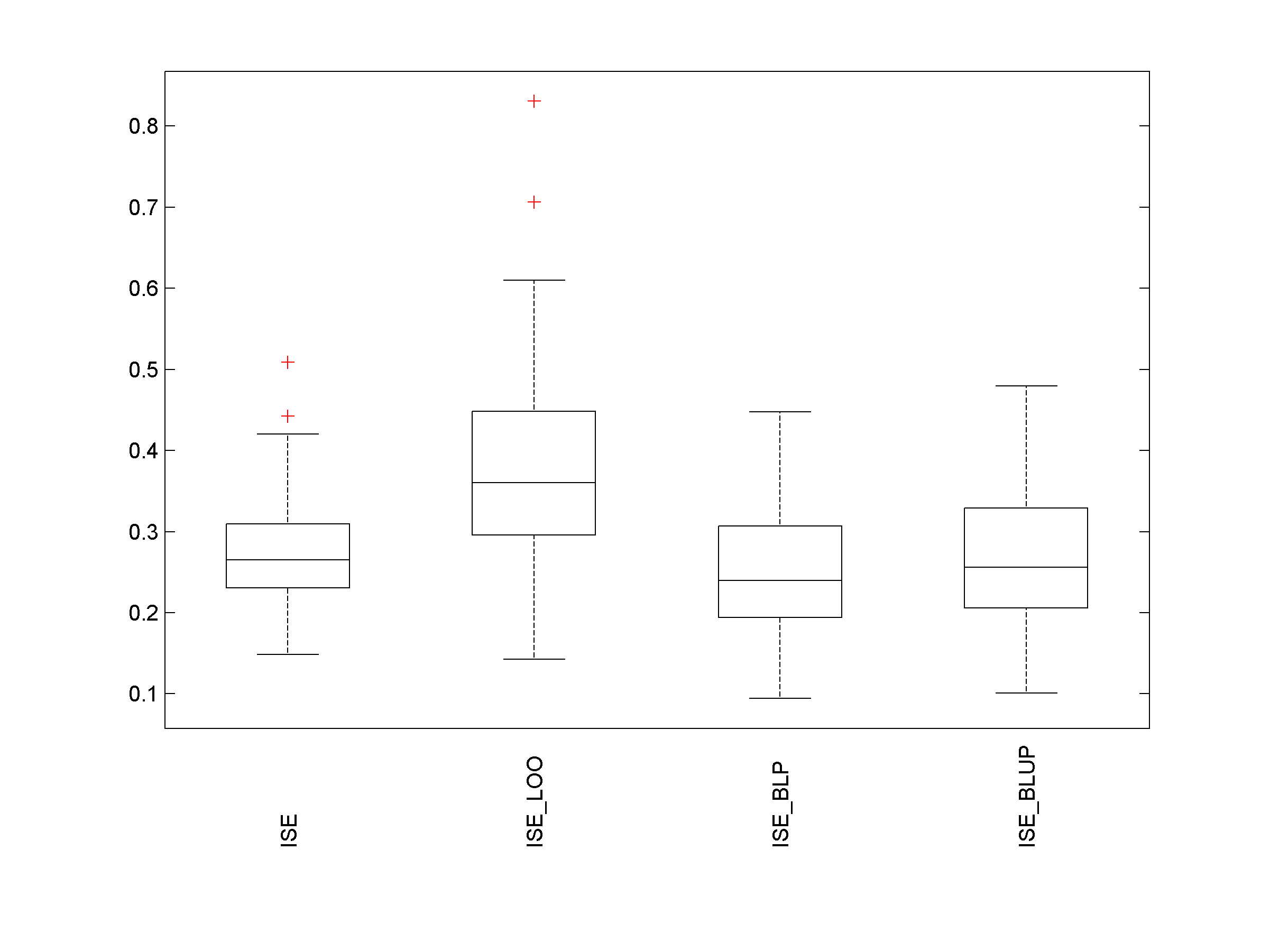} \\
\includegraphics[width=0.49\textwidth]{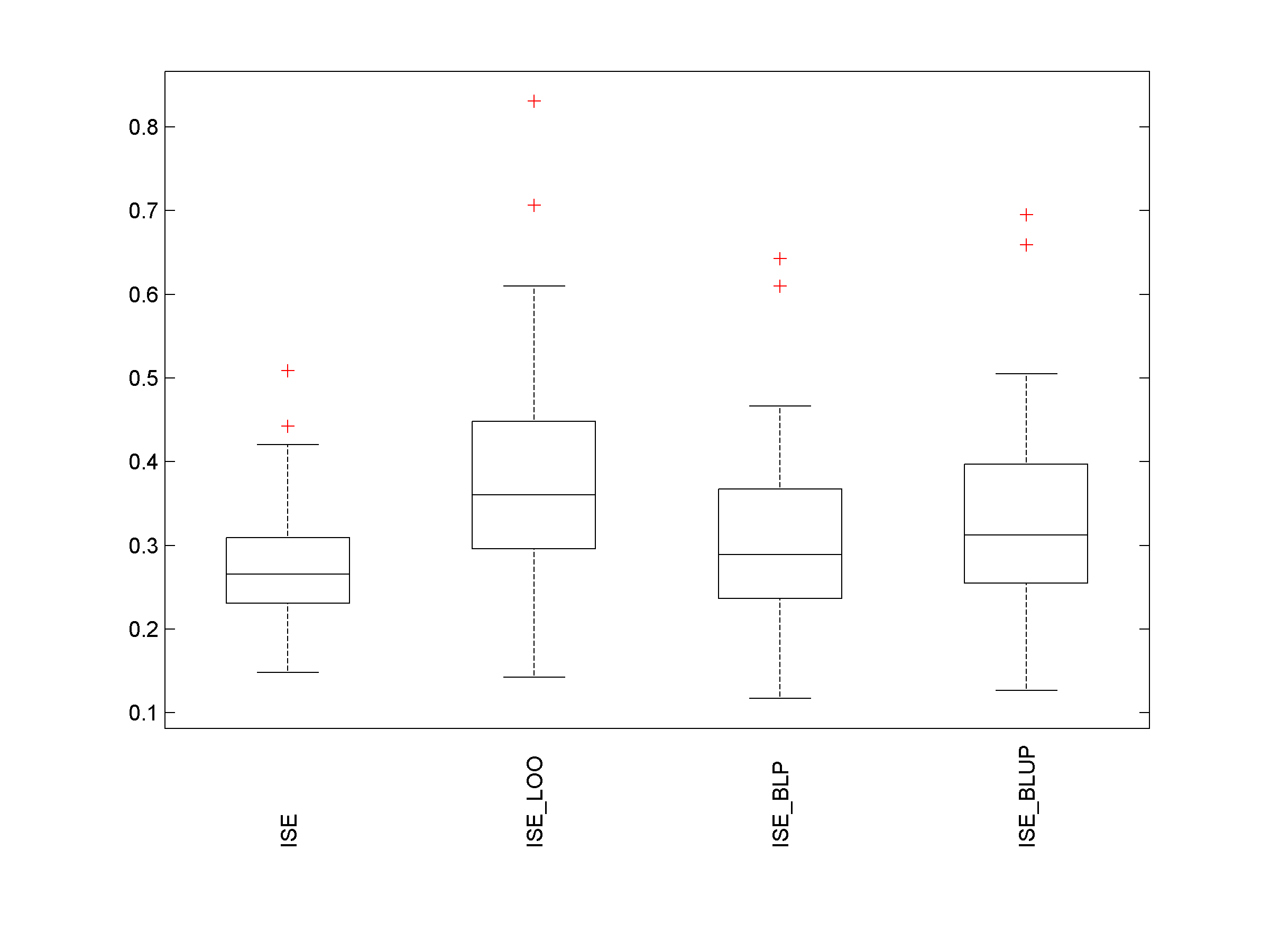}
\includegraphics[width=0.49\textwidth]{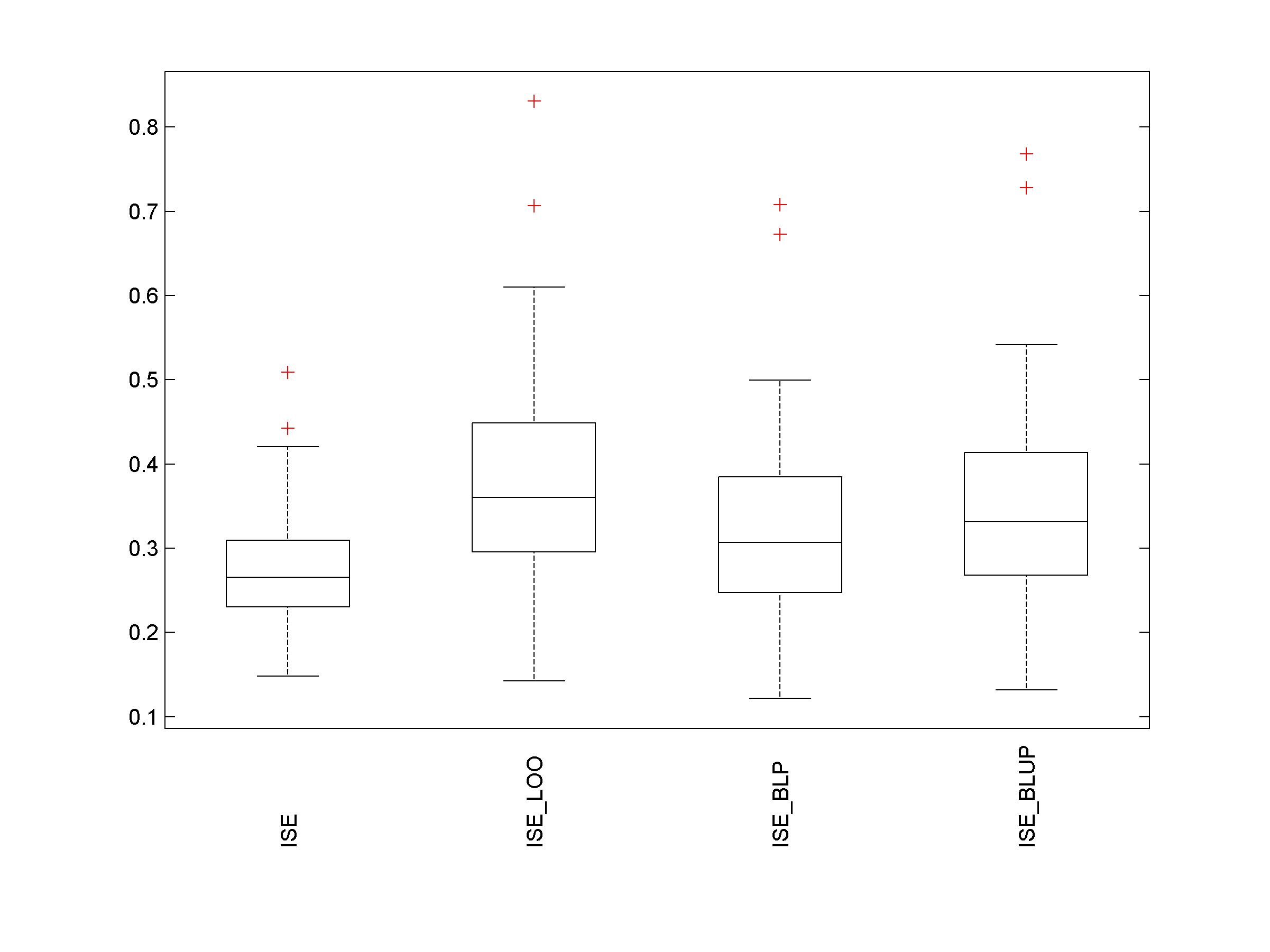}
\caption{\small Same as Figure~\ref{F:hist-theta0-adapt_theta1-LOO_theta2-LOO}-top-right (random functions $f_m$ with $d=4$ and $m=n=40)$ but with noisy observations with standard deviation $\mg=0.25$. First row: $r^{(e)}=\mg^2$ (left) and $r^{(e)}=0.1\,\mg^2$ (right); second row: $r^{(e)}=5\,\mg^2$ (left) and $r^{(e)}=10\,\mg^2$ (right). $\ISE(\eta_n)$ and $\hISE_{LOO}(\eta_n)$ are the same on all panels.} \label{F:hist-d4_theta0-adapt_theta1-LOO_theta2-LOO_nugget0625_factor}
\end{figure}
\end{center}

Figure~\ref{F:hist-d4_theta0-adapt_theta1-LOO_theta2-LOO_nugget0625_factor} shows boxplots of $\ISE(\eta_n)$, $\hISE_{LOO}(\eta_n)$, $\hISE_{BLP}(\eta_n)$ and $\hISE_{BLUP}(\eta_n)$ for 100 random functions $f_m$ and noise realizations, for four different $r^{(e)}$: $r^{(e)}=\mg^2=0.0625$ (top left), which can be considered as a natural choice in the present context; a severely underestimated value $r^{(e)}=\mg^2/10$ (top right); and two overestimated values, $r^{(e)}=5\, \mg^2$ and $r^{(e)}=10\, \mg^2$ (second row). Unsurprisingly, the best performance is obtained for $r^{(e)}$ close to $\mg^2$, but using a (much) smaller value has little effect; performance deteriorates when $r^{(e)}$ becomes much larger than $\mg^2$, but remains acceptable (and superior to that of $\hISE_{LOO}(\eta_n)$).

\subsection{Unreliable estimation of quantiles and conditional values-at-risk}\label{S:quantiles}

The method proposed in the paper provides an estimate $\widehat{\mve_n^2}(\xb)$ of $\mve_n^2(\xb)$ at any $\xb$, and in all the examples presented, $\ISE(\eta_n)$ has been estimated by the empirical mean $(1/N) \sum_{i=1}^N \widehat{\mve_n^2}(\xb^{(i)})$ calculated for $N$ points $\xb^{(i)}$ distributed with $\mu$. One may thus think of using the $N$ estimates $\widehat{\mve_n^2}(\xb^{(i)})$ to compute an empirical quantile (or value-at-risk) $Q_\ma$ and conditional value at risk $\CVaR_\ma$ (see, e.g. \cite{SarykalinSU2008} and the references therein) at a given level $\ma$. However, the errors $\mve_n^2(\xb)$ as well as the estimates $\widehat{\mve_n^2}(\xb)$ are correlated\footnote{With the notation of Section~\ref{S:LOOCV-GPmodel}, under the assumption $Y_\xb \sim \GP(0,\ms^2\,K)$, $\cov\{\mve_n^2(\xb),\mve_n^2(\xb')\}=2\,\ms^4\,\rho_n^4(\xb,\xb')$ and $\cov\{\widehat{\mve_n^2}(\xb),\widehat{\mve_n^2}(\xb')\}=2\,\ms^4\,\betab\TT(\xb)(\Rb_n\TT\Kb_n \Rb_n)^{\odot 2}\betab(\xb')$ with $\betab(\xb)$ given by \eqref{beta_BLP} for $\widehat{\mve_n^2}_{BLP}(\xb)$ (Section~\ref{S:ISE-any-predictor}) and by \eqref{beta_BLUP}  for the unbiased version $\widehat{\mve_n^2}_{BLUP}(\xb)$ (Section~\ref{S:ISE_BLUP}).}, and the distributions of $\mve_n^2(\xb)$ and $\widehat{\mve_n^2}(\xb)$ may significantly differ. The following example provides an illustration.

\begin{center}
\begin{figure}[ht]
\centering
\includegraphics[width=0.49\textwidth]{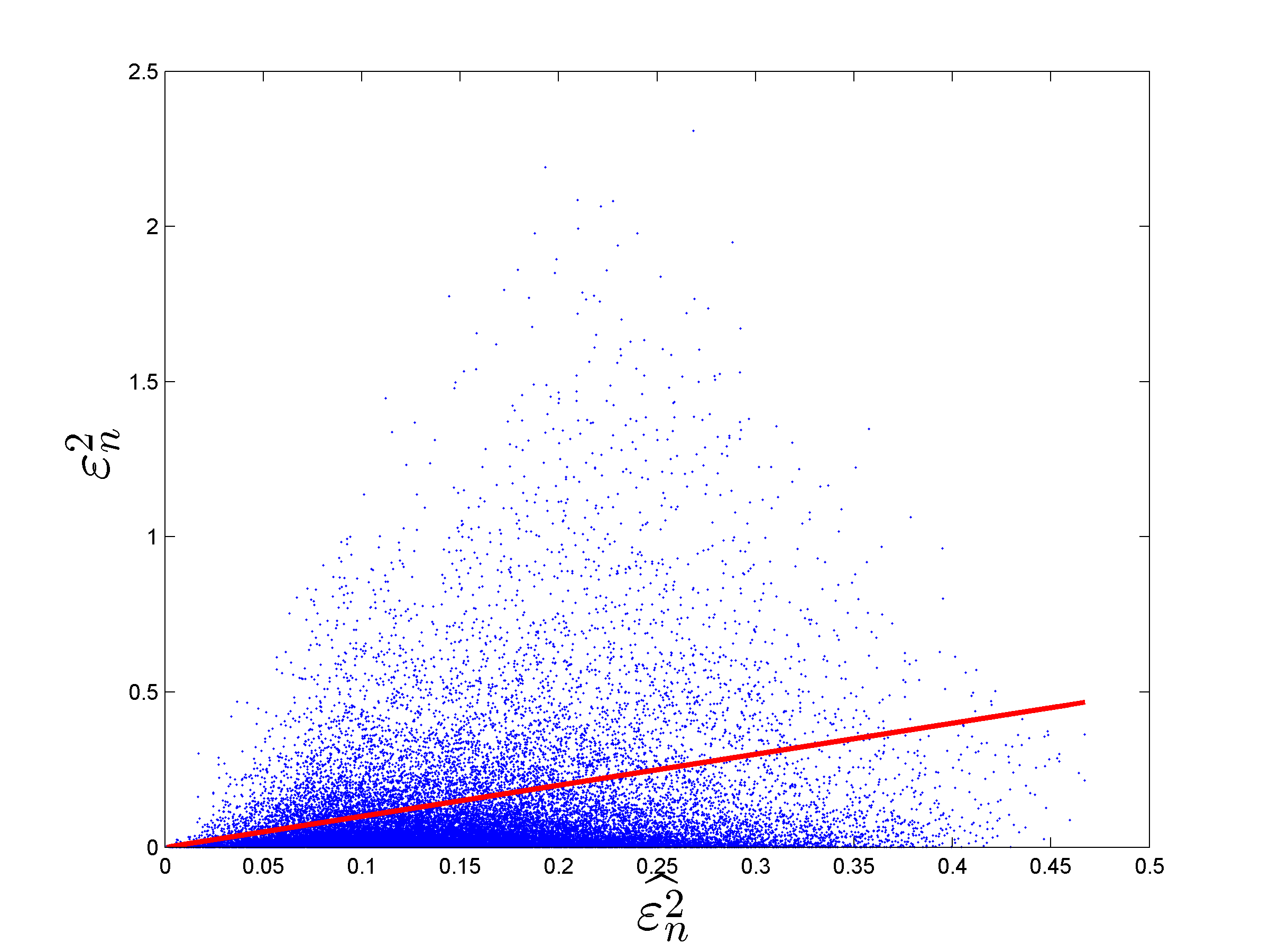}
\includegraphics[width=0.49\textwidth]{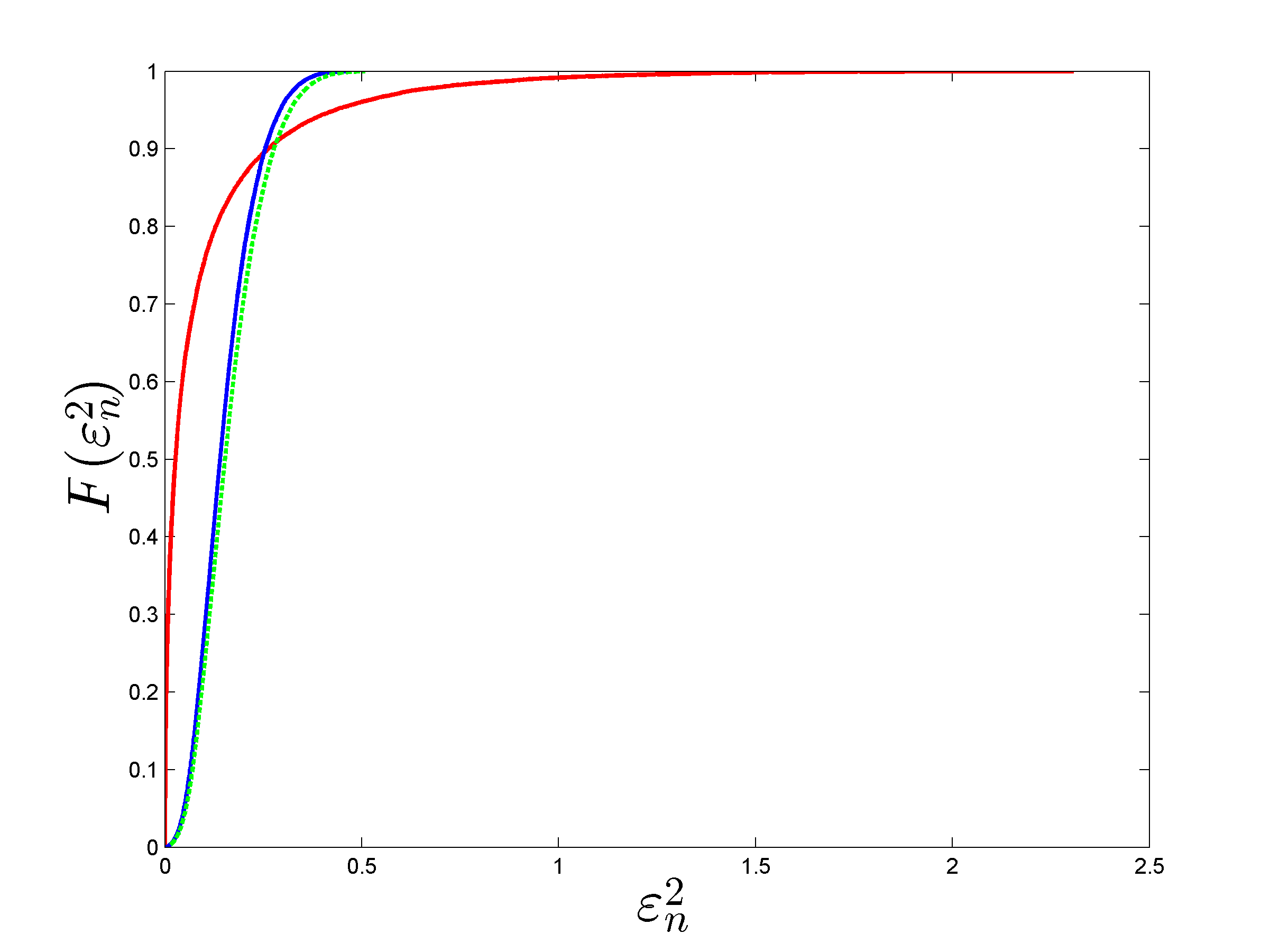}
\caption{\small Left: scatter plot of $(\widehat{\mve_n^2}_{BLP}(\xb^{(i)}),\mve_n^2(\xb^{(i)}))$ for one random $f_m$. Right: empirical c.d.f.\ of the true squared errors $\mve_n^2(\xb^{(i)})$ ({\color{red} ---}) and of their BLP and BLUP estimates $\widehat{\mve_n^2}_{BLP}(\xb^{(i)})$ ({\color{blue} ---}) and $\widehat{\mve_n^2}_{BLUP}(\xb^{(i)})$ ({\color{green} $\cdots$}), for the same random $f_m$.} \label{scatterplot_errors2}
\end{figure}
\end{center}

The framework is as in Section~\ref{S:simulations-m-n-d} for $d=4$, with $m=n=10\,d=40$ and $N=2^{15}$.
The left panel of Figure~\ref{scatterplot_errors2} presents a scatter plot of $(\widehat{\mve_n^2}_{BLP}(\xb^{(i)}),\mve_n^2(\xb^{(i)}))$ for one random $f_m$, with a red solid line showing the first diagonal. There are more small squared errors $\mve_n^2(\xb^{(i)})$ than small squared errors $\widehat{\mve_n^2}_{BLP}(\xb^{(i)})$, but some $\mve_n^2(\xb^{(i)})$ are much larger than $\widehat{\mve_n^2}_{BLP}(\xb^{(i)})$.
The right panel shows, for the same simulation, the empirical c.d.f.\ $F_T$ of the $2^{15}$ true squared prediction errors $\mve_n^2(\xb^{(i)})$ (red solid line) and the empirical c.d.f.\ $F_{BLP}$ and $F_{BLUP}$ of the BLP and BLUP estimates $\widehat{\mve_n^2}_{BLP}(\xb^{(i)})$ and $\widehat{\mve_n^2}_{BLUP}(\xb^{(i)})$ of Sections~\ref{S:ISE-any-predictor} and \ref{S:ISE_BLUP}, respectively in blue solid line and green dashed line. The behavior observed is typical: $F_{BLP}(t)$ and $F_{BLUP}(t)$ are very close; $F_T(t)$ is larger than $F_{BLP}(t)$ and $F_{BLUP}(t)$ for small $t$ but is smaller for large $t$ as $\widehat{\mve_n^2}_{BLP}(\xb)$ and $\widehat{\mve_n^2}_{BLUP}(\xb)$ tend to smooth $\mve_n^2(\xb)$.

\begin{center}
\begin{figure}[ht]
\centering
\includegraphics[width=0.49\textwidth]{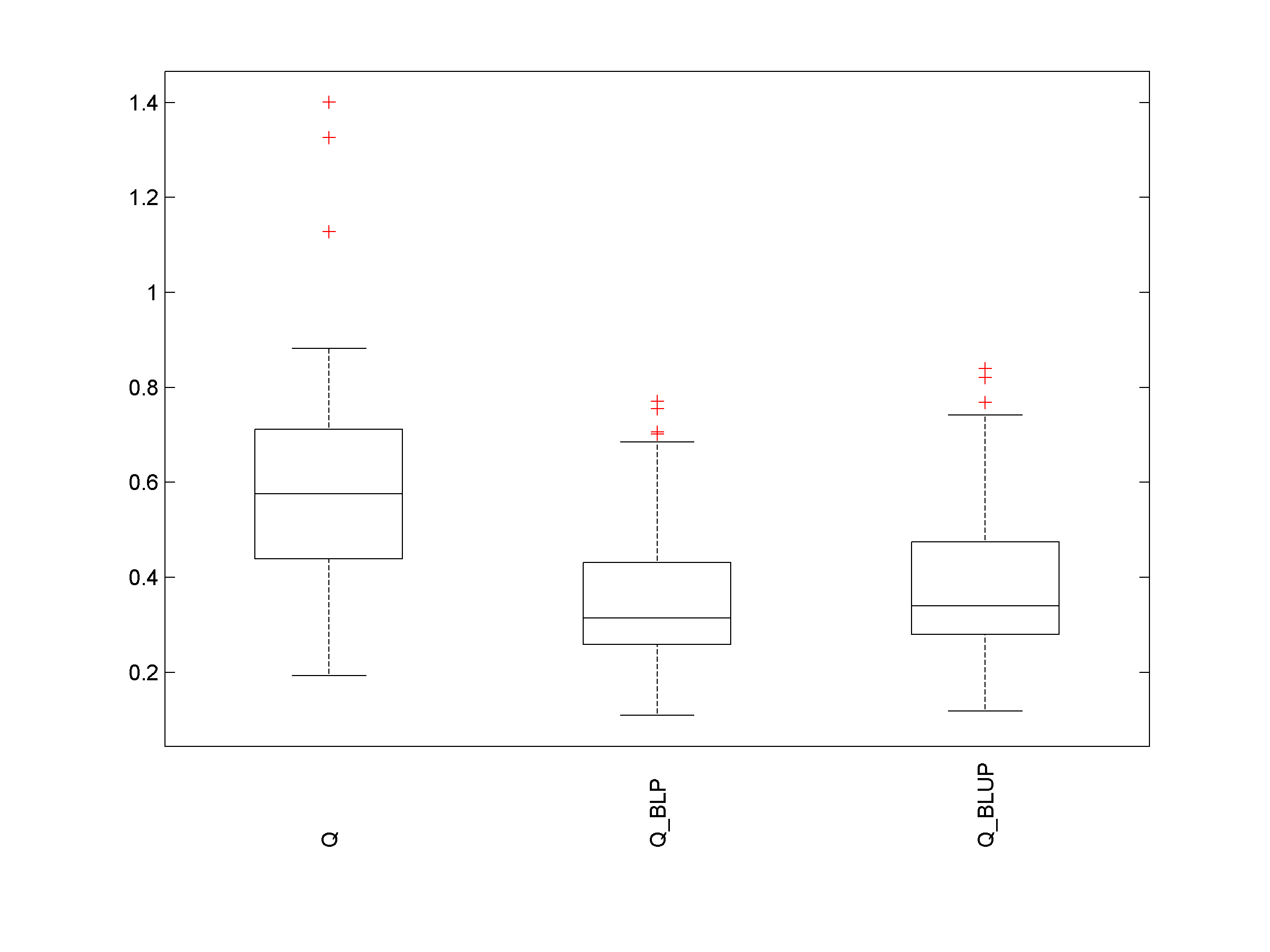}
\includegraphics[width=0.49\textwidth]{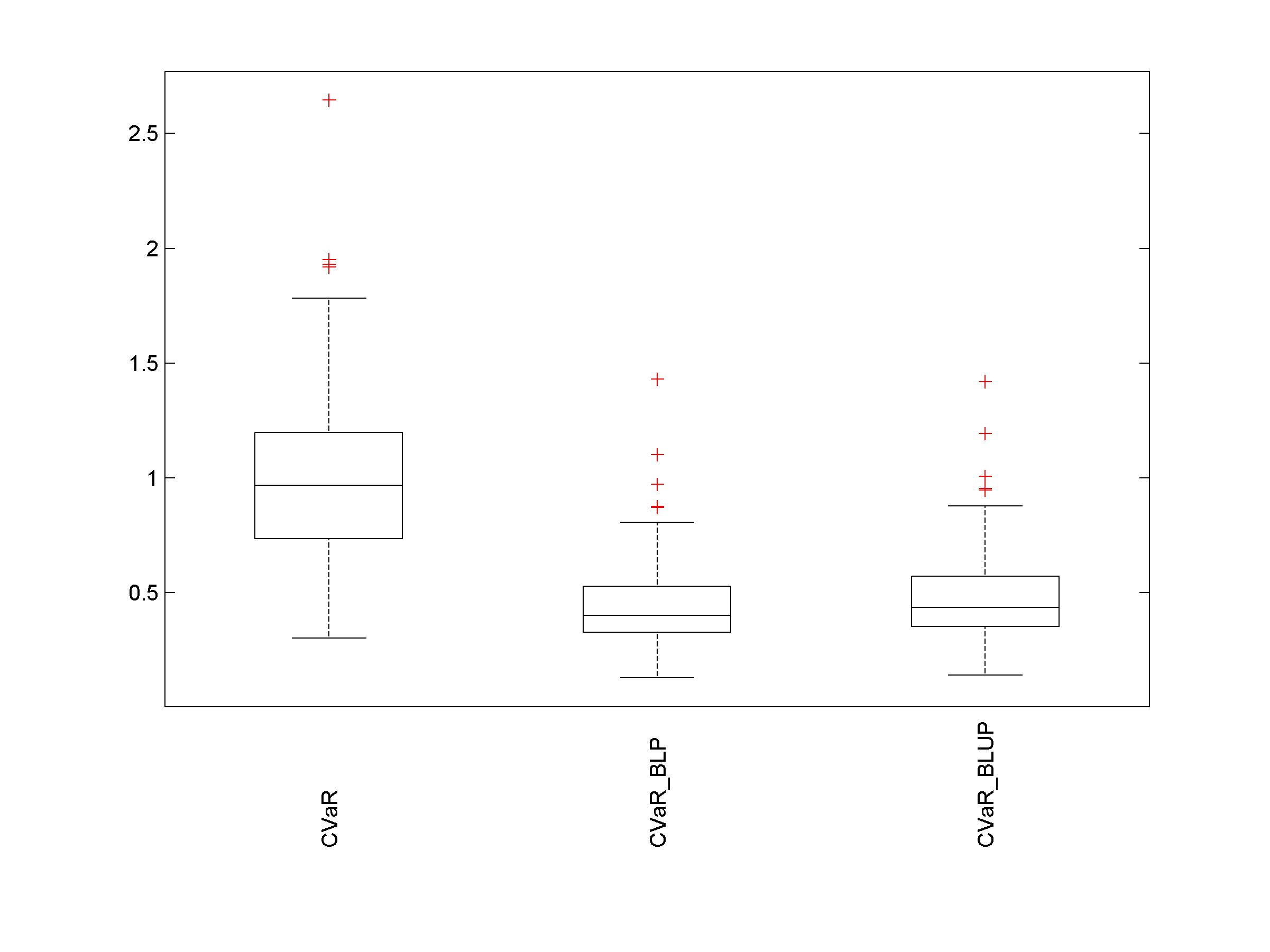}\\
\includegraphics[width=0.49\textwidth]{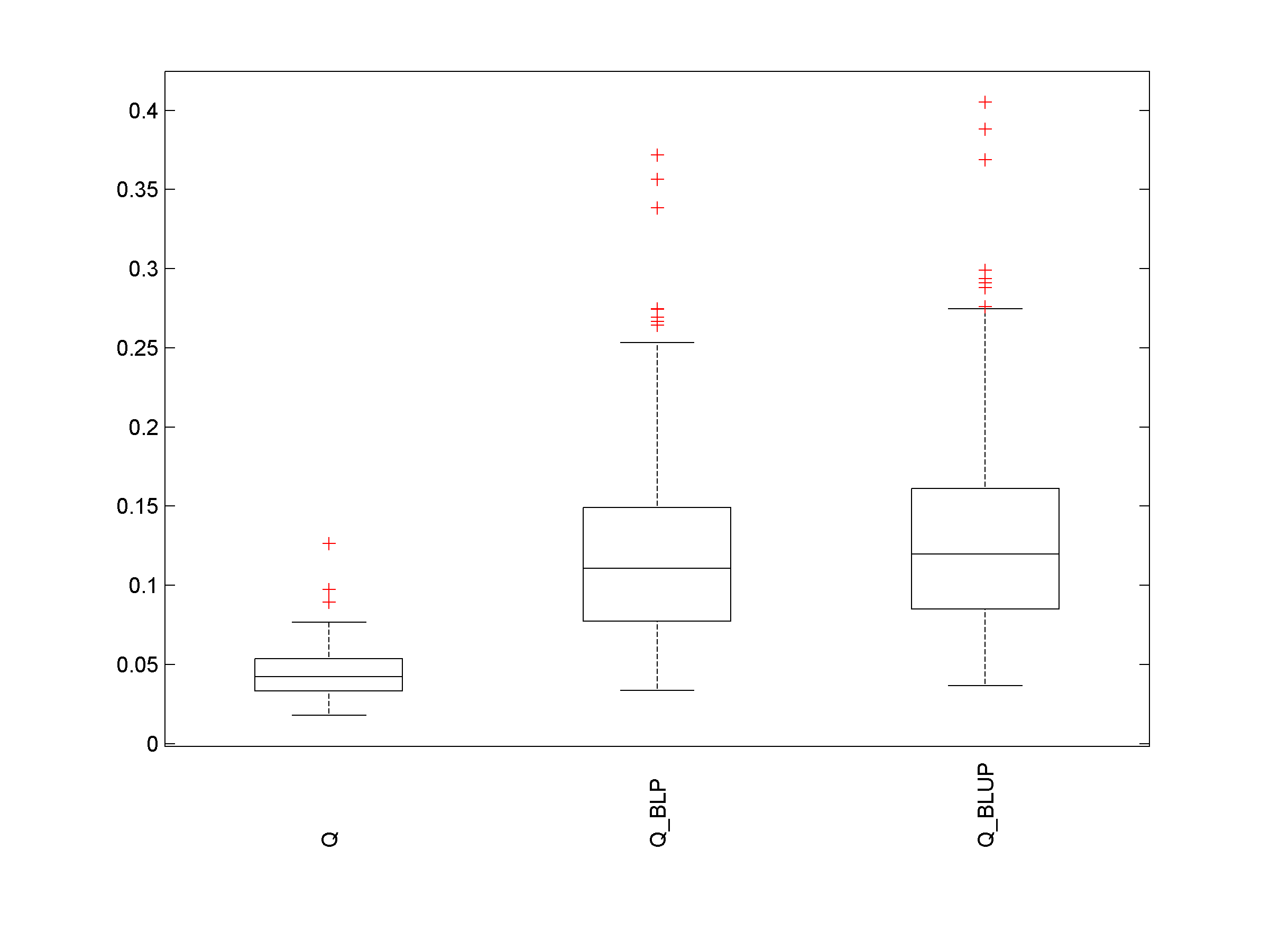}
\includegraphics[width=0.49\textwidth]{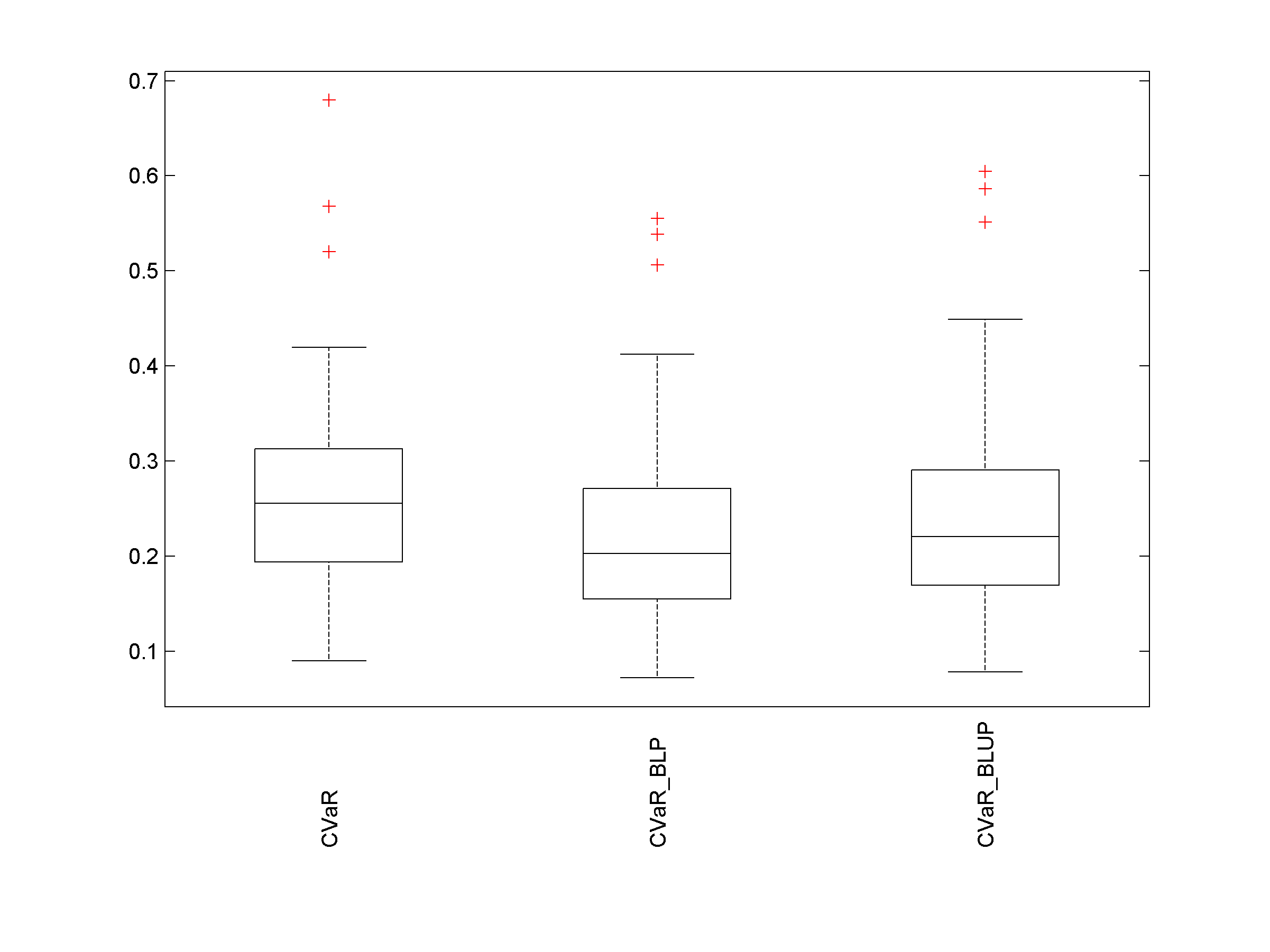}
\caption{\small Boxplots of the quantiles $Q_\ma$ (left) conditional value-at-risk $\CVaR_\ma$ (right) for the $\mve_n^2(\xb^{(i)})$ and their estimates $\widehat{\mve_n^2}(\xb^{(i)})$ (the $\mve_n^2(\xb^{(i)})$ and $\widehat{\mve_n^2}(\xb^{(i)})$ are the same as in the top-right panel of Figure~\ref{F:hist-theta0-adapt_theta1-LOO_theta2-LOO}); top row:  $\ma=0.95$; bottom row: $\ma=0.5$.} \label{F:histQ-d4_theta0-adapt-m10d_theta1-LOO_theta2-LOO}
\end{figure}
\end{center}

Figure~\ref{F:histQ-d4_theta0-adapt-m10d_theta1-LOO_theta2-LOO} shows boxplots of $Q_\ma$ (left) $\CVaR_\ma$ (right) for the true squared prediction errors $\mve_n^2(\xb^{(i)})$ and the BLP and BLUP estimates $\widehat{\mve_n^2}_{BLP}(\xb^{(i)})$ and $\widehat{\mve_n^2}_{BLUP}(\xb^{(i)})$ for $\ma=0.95$ (top row) and $\ma=0.5$ (bottom row), for 100 random functions $f_m$.
In agreement with Figure~\ref{scatterplot_errors2}-right, we observe that $Q_\ma$ is strongly underestimated (respectively, overestimated) for $\ma=0.95$ (respectively, $\ma=0.5$). The presence of squared errors $\mve_n^2(\xb^{(i)})$ much larger than $\widehat{\mve_n^2}_{BLP}(\xb^{(i)})$ explains that $\CVaR_\ma$ is underestimated for both values of $\ma$ (however, performance improves when $\ma$ decreases, as $\CVaR_\ma \to \ISE(\eta_n)$ when $\ma \to 0$). The information that the $\widehat{\mve_n^2}_{BLP}(\xb^{(i)})$ and $\widehat{\mve_n^2}_{BLUP}(\xb^{(i)})$ provide on the tail distribution of the squared prediction errors $\mve_n^2(\xb)$ is therefore very unreliable. We have observed similar disappointing behavior with other examples.

\section{GP with parameterized mean}\label{S:parameterised-mean}

The developments of Section~\ref{S:ISE_BLP} assumed that $f$ is the realization of GP with zero mean. Here we show how to estimate the ISE of a given linear predictor $\eta_n(\cdot)=\wb_n\TT(\cdot)\yb_n$ when this assumption is relaxed.

\subsection{Universal kriging model}\label{S:A-UK}
Consider the framework of universal kriging, and assume that $f$ is the realization of a GP $Y_\xb \sim \GP(\taub\TT\hb(\xb),\ms^2 K)$, with $\hb(\xb)=[h_1(\xb),\ldots,h_p(\xb)]\TT$ a vector of $p$ known functions on $\SX$ and $\taub$ a vector of unknown parameters. We then have
\bea
\ISE(\eta_n) &=& \int_\SX [Y_\xb - \wb_n\TT(\xb)\yb_n]^2\,\mu(\dd\xb) \\
&=& \int_\SX \left[Z_\xb+\taub\TT\hb(\xb)-\wb_n\TT(\xb)(\zb_n+\Hb_n\taub)\right]^2 \, \mu(\dd\xb)\,,
\eea
where $Z_\xb=Y_\xb-\taub\TT\hb(\xb) \sim\GP(0,\ms^2 K)$, $\Hb_n$ is the $n\times p$ matrix with $i$-th row equal to $\hb\TT(\xb_i)$, and $\zb_n=\yb_n-\Hb_n\taub$. This gives
\be
\ISE(\eta_n) &=&  \ISE_0(\eta_n) + I(\taub) + 2\, I_n(\taub) \,, \label{ISE-UK}
\ee
where
\be
\ISE_0(\eta_n) &=& \int_\SX [Z_\xb - \wb_n\TT(\xb)\zb_n]^2 \, \mu(\dd\xb)\,, \label{ISE_0} \\
I(\taub) &=& \int_\SX \left\{\taub\TT\left[\hb(\xb)-\Hb_n\TT\wb_n(\xb)\right]\right\}^2 \, \mu(\dd\xb) \,, \nonumber \\
I_n(\taub) &=& \int_\SX [Z_\xb - \wb_n\TT(\xb)\zb_n] \left\{\taub\TT\left[\hb(\xb)-\Hb_n\TT\wb_n(\xb)\right]\right\}\, \mu(\dd\xb) \,. \nonumber
\ee
In \eqref{ISE-UK}, $\ISE_0(\eta_n)$ is the ISE for the centered GP model $Z_\xb\sim\GP(0,\ms^2 K)$, which can be estimated with the method presented in Section~\ref{S:ISE_BLP}, $I(\taub)$ is a constant and $I_n(\taub)$ has zero mean.

A simple approach to estimate $\ISE(\eta_n)$, assuming a kernel $K^{(e)}$, is therefore as follows.
\begin{itemize}
  \item[(i)] Estimate the parametric trend, using for example the BLUE for $\taub$ given by
  \be\label{tau_ML}
  \widehat\taub^n=(\Hb_n\TT{\Kb_n^{(e)}}^{-1}\Hb_n)^{-1} \Hb_n\TT{\Kb_n^{(e)}}^{-1} \yb_n \,.
  \ee
  \item[(ii)] Remove $\Hb_n\widehat\taub^n$ from the observations $\yb_n$ and estimate $\ISE_0(\eta_n)$ \eqref{ISE_0} for these centered observations $\zb_n$ under the assumption $Z_\xb\sim\GPmodel{e}$;
  \item[(iii)] Add $I(\widehat\taub^n)$ to the estimated ISE.
\end{itemize}

This approach neglects the error due to the estimation of $\taub$, which is acceptable when $p \ll n$; other approaches, more accurate, could certainly be developed, at the expense of increased complexity. A direct application of the approach of Section~\ref{S:ISE-any-predictor} through the calculation of $\Ex\{\mve^2(\xb)\mve_{-i}^2\}$ and $\Ex\{\mve_{-i}^2(\xb)\mve_{-j}^2\}$ for the model $Y_\xb \sim \GP(\taub\TT\hb(\xb),\ms^2 K)$ would also be possible. However, these expressions depend explicitly on $\taub$ and $\ms^2$, whereas the estimation of $\ISE_0(\eta_n)$ by \eqref{estimate-ISE-general} does not require the construction of an estimator of $\ms^2$.

\subsection{Ordinary kriging model}\label{S:A-OK}
The model $Y_\xb\sim\GP(\tau,\ms^2 K)$ with $\tau\in\RR$ and $\hb(\xb)\equiv 1$ (ordinary kriging) is frequently used. In this case we get $\widehat\tau^n=(\1b_n\TT\Kb_n^{-1}\yb_n)/(\1b_n\TT\Kb_n^{-1}\1b_n)$ and $I(\tau)=\tau^2\, \int_\SX [1-\wb_n\TT(\xb)\1b_n]^2\,\mu(\dd\xb)$.

If $\eta_n$ is such that $\wb_n\TT(\xb)\1b_n=1$ for all $\xb$, then any translation of the observations leaves the ISE invariant (since the prediction $\eta_n$ itself is invariant) and the LOO residuals $\mve_{-i}$ are invariant too. We have $\ISE(\eta_n)=\int_\SX [Z_\xb - \wb_n\TT(\xb)\zb_n]^2 \, \mu(\dd\xb)$ and $\widehat{\ISE}_{BLP}(\eta_n)$ can be calculated for a centered $\GPmodel{e}$ without centering the observations. The case of the ordinary kriging predictor is a typical example.
More generally, denote by $\Ex_\tau\{\cdot\}$ and $\MSE_{\tau}\{\cdot\}$ the expectation and MSE under the model $\GP(\tau,\ms^2 K)$. Direct calculation shows that, when $\widehat{\ISE}_{BLP}(\eta_n)$ is calculated for $\GPmodel{e}$ (i.e., assuming that $\tau=0$), we have
\bea
\Ex_\tau\{\widehat{\ISE}_{BLP}(\eta_n)\}=\Ex_0\{\widehat{\ISE}_{BLP}(\eta_n)\}+\tau^2\, (\1b_n\TT\Rb_n)^{\odot 2}\Sb_n^{-1}\bb_n \,,
\eea
where $\Ex_0\{\widehat{\ISE}_{BLP}(\eta_n)\}$ is given by \eqref{E_0},
$\Rb_n$ is the matrix in \eqref{Rn} and $\Sb_n$ and $\bb_n$ are respectively given by \eqref{Sn} and \eqref{bn}. We obtain similarly
\bea
\MSE_\tau\{\widehat{\ISE}_{BLP}(\eta_n)\}=\MSE_0\{\widehat{\ISE}_{BLP}(\eta_n)\} + \tau^2\, C_n \,,
\eea
where $\MSE_0\{\widehat{\ISE}_{BLP}(\eta_n)\}$ is given by \eqref{MSE_0}
and $C_n$ tends to zero when $\int_\SX [1-\wb_n\TT(\xb)\1b_n]^2\,\mu(\dd\xb)$ and $\|\Rb_n\TT\1b_n\|$ tend to zero.
This indicates that the performance of $\widehat{\ISE}_{BLP}(\eta_n)$ constructed under the assumption $\tau=0$ is preserved when $\tau$ is small or when the predictor $\eta_n$ is such that, for all $n$ and $\Xb_n$, $\wb_n\TT(\xb)\1b_n\approx 1$ for all $\xb$; see Section~\ref{S:environmental-model} for an example. Similar developments show that $\widehat{\ISE}_{BLP}(\eta_n)$ calculated for $\GPmodel{e}$ behaves
similarly when the true data generating model is $\GP(\taub\TT\hb(\xb),\ms^2 K)$ or $\GP(0,\ms^2 K)$ provided that the predictor $\eta_n$ satisfies $\Hb_n\TT\wb_n(\xb)\approx \hb(\xb)$ for all $\xb$. The correction of Section~\ref{S:A-UK} can be applied otherwise.

\section{Mixtures of GP models}\label{S:mixtures}

In the construction of $\hISE_{BLP}(\eta_n)$, instead of assuming that $f$ is the realization of a unique GP, $Y_\xb\sim\GPmodel{e}$, we may consider a mixture of GP; that is, consider a family $\{K_t\}_{t=1,\ldots,T}$ of $T$ different kernels (stationary or not, with different regularities\ldots) and assume that $Y_\xb|s \sim\GP(0,\ms_e^2K_s)$, with $\Pr\{s=t\}=\nu_t$. (The infinite mixture model could be considered as well but is computationally more difficult to handle.) All expectations under this finite mixture model can be decomposed as
\bea
\Ex\{X\} = \sum_{t=1}^T \nu_t \Ex\{X|Y_\xb\sim\GP(0,\ms_e^2K_t)\} \,.
\eea
This gives for instance $\Ex\{\mve^2(\xb)\}=\ms^2 \sum_{t=1}^T \nu_t \rho_{n,t}^2(\xb)$ where $\rho_{n,t}^2(\xb)$ is given by \eqref{rho_n^2-b} with $K=K_t$, and with obvious notation
$\Ex\{\mveb_{LOO}\mveb_{LOO}\TT\}=\ms^2 \Rb_n^T \left(\sum_{t=1}^T \nu_t \Kb_{n,t}\right) \Rb_n$, $\Ex\{\mveb_{LOO}\mve_n(\xb)\}=\ms^2 \Rb_n^T \sum_{t=1}^T \nu_t\tb_{n,t}(\xb)$, etc. Developments similar to those of Section~\ref{S:ISE_BLP} then yield the expressions of $\hISE_{BLP}(\eta_n)$ and $\hISE_{BLUP}(\eta_n)$.
The weights $\nub=(\nu_1,\ldots,\nu_T)\TT$ can be adjusted to the data $\yb_n$, as in Bayesian Model Averaging (BMA), see \cite{PR-Technometrics2016} (with a proposition concerning the choice of prior weights in Section~5 of the same paper).

Since a mixture $\eta_n(\cdot)=\sum_{t=1}^T \nu_t\, \eta_{n,t}(\cdot)$ of kriging predictors obtained by BMA with {\em fixed weights} (i.e., not depending on $\yb_n$) remains linear in $\yb_n$, ISE estimation by $\hISE_{BLP}(\cdot)$ can also be applied to such mixture models. When the weights satisfy $\nub\TT\1b_T=1$, then, with the notation of Section~\ref{S:ISE_LOO}, the $i$-th LOO error becomes $\mve_{-i}=\sum_{t=1}^T \nu_t\, \left[y_i-\eta_{n\setminus i,t}(\xb_i)\right]=\sum_{t=1}^T \nu_t\,\mve_{-i,t}$ and the squared LOO errors $\mveb_{LOO}^{\odot 2}$ are quadratic in $\nub$. Denoting $\Eb_{LOO}$ the $T\times n$ matrix with $\{\Eb_{LOO}\}_{t,i} = \mve_{-i,t}$, any ISE estimator of the form $\hISE(\eta_n[\nub])=\gammab\TT\mveb_{LOO}^{\odot 2}$ (thus in particular $\hISE_{BLP}(\eta_n[\nub])$ and $\hISE_{BLUP}(\eta_n[\nub])$) can be written as
\bea
\hISE(\eta_n)=\nub\TT \Eb_{LOO}\Gammab\Eb_{LOO}\TT \nub\,,
\eea
where $\Gammab=\diag\{\mg_i,\, i=1,\ldots,n\}$. Minimization of $\hISE(\eta_n[\nub])$ with respect to $\nub$ under the constraint $\nub\TT\1b_T=1$ yields the optimal predictor $\eta_n[\nub^*]$ (in the sense of $\hISE(\cdot)$) with
\bea
\nub^*=\frac{(\Eb_{LOO}\Gammab\Eb_{LOO}\TT)^{-1} \1b_T}{\1b_T\TT(\Eb_{LOO}\Gammab\Eb_{LOO}\TT)^{-1} \1b_T}\,.
\eea
It is tempting to iterate the process, as suggested in Section~\ref{S:conclusion} of the paper for model selection: indeed, the optimal weights $\nub^*$ could be used to define a mixture of GP models for the construction of $\hISE(\eta_n[\nub])$, whose optimization would lead to an updated optimal $\nub^*$.

\section{A 1-d example of poor performance due to a bad design}\label{S:poor-designs}

This simple example illustrates a limitation of the method described in the last paragraph of the conclusion section: due to design sparsity, when the predictor is much smoother than $f$, the LOO squared residuals $\mve_{-i}^2$ can be very small and $\hISE_{LOO}(\eta_n)$ and $\hISE_{BLP}(\eta_n)$ may severely underestimate $\ISE(\eta_n)$, so that the inaccuracy of $\eta_n$ can remain undetected. The problem of too sparse a design relative to the variability of $f$ is more serious in high dimensions, but this one-dimensional case already gives a picture of the possible situation.

\begin{center}
\begin{figure}[ht]
\centering
\includegraphics[width=0.49\textwidth]{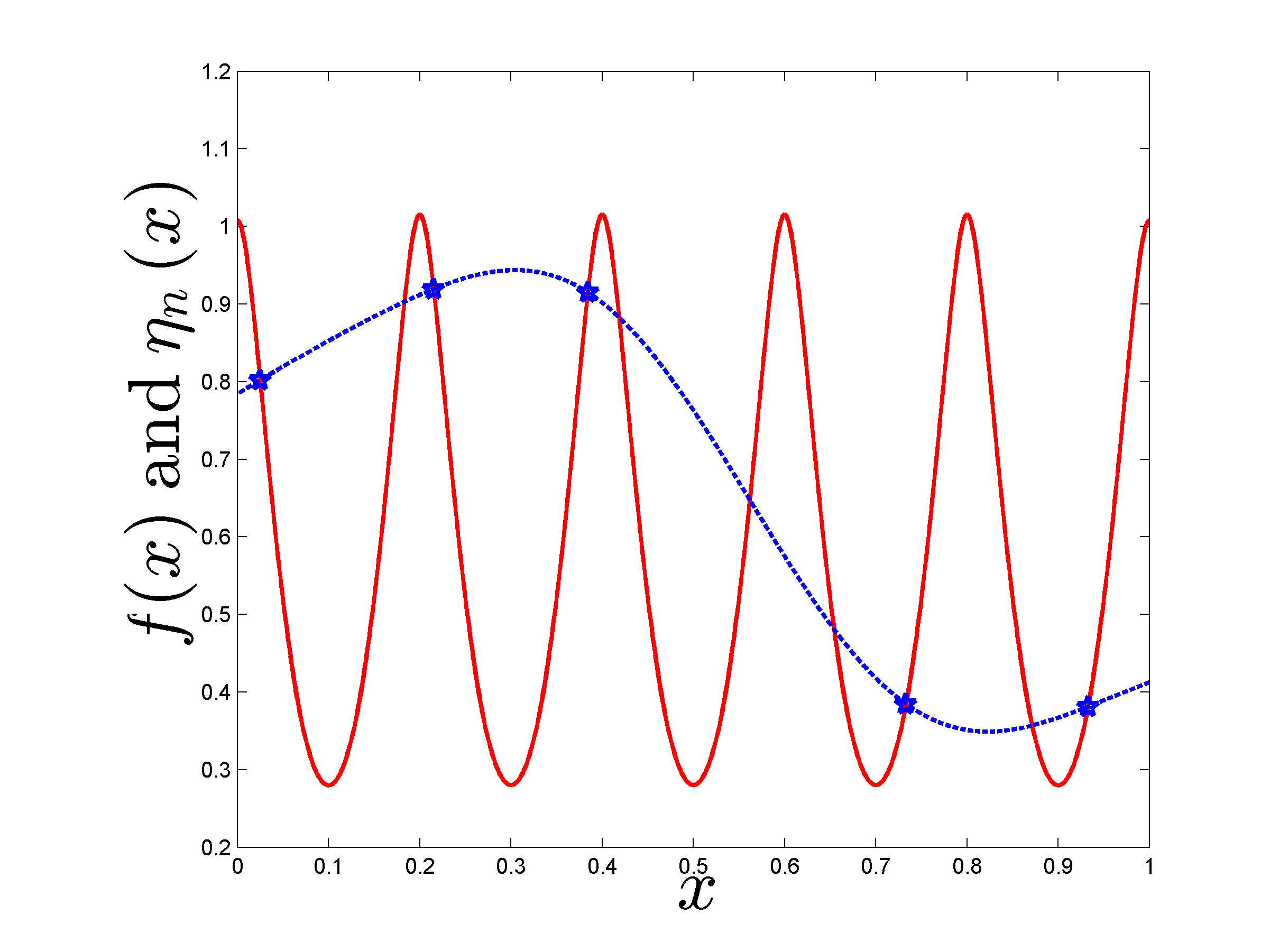}
\includegraphics[width=0.49\textwidth]{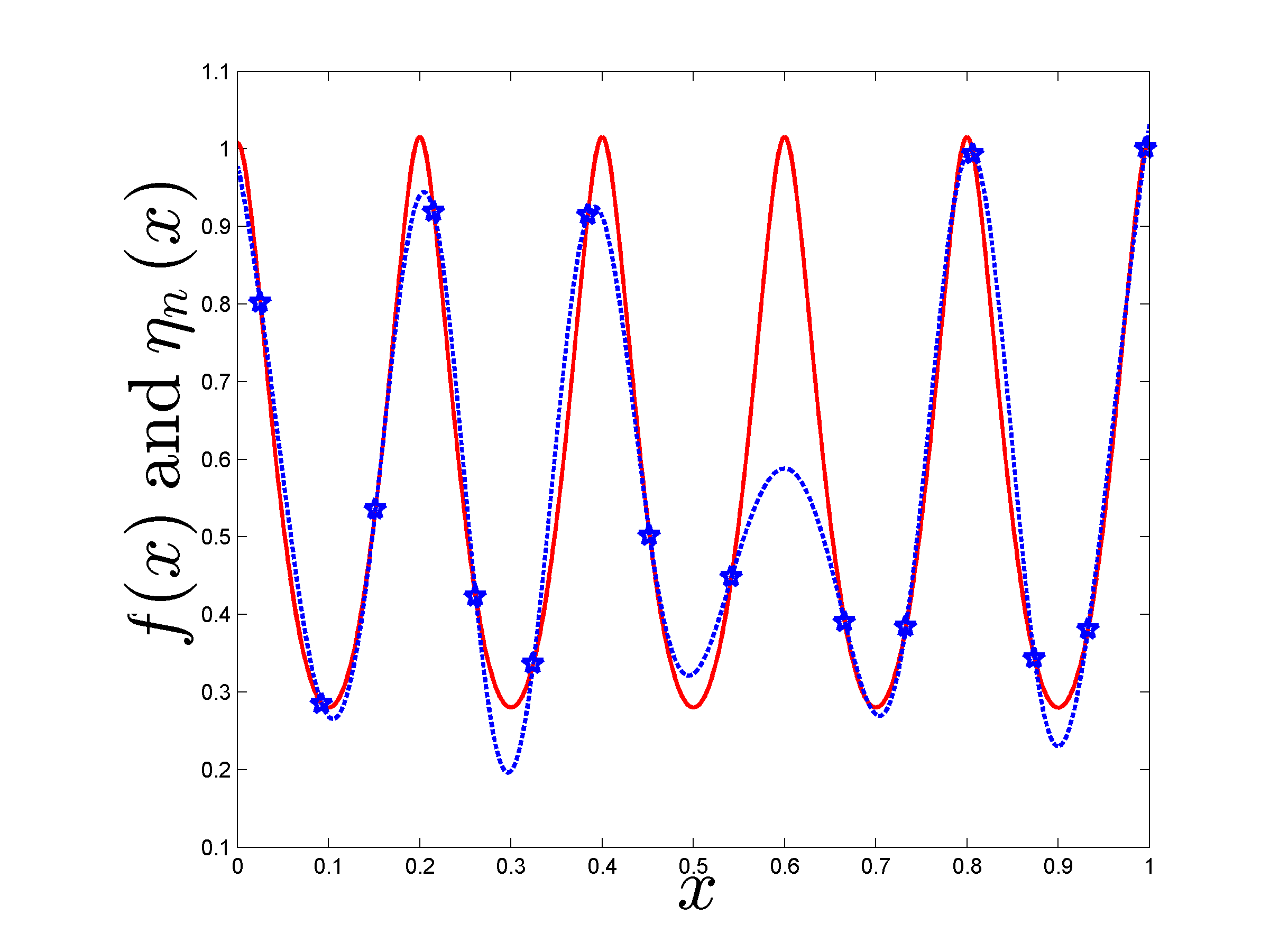}
\caption{\small $f$ and its interpolator $\eta_n$ for $n=5$ (left) and $n=15$ (right).}
\label{F:fdeceptive-eta-d1_n5_nn5_theta1-LOO}
\end{figure}
\end{center}

Here $f(x)=\sum_{i=1}^5 \psi_{3/2,20}(|x-z_i|)$ with $\{z_1,\ldots,z_5\}=\{0,0.2,0.4,0.6,0.8,1\}$ and $\psi_{3/2,\mt}$ given by \eqref{Matern32}; $\Xb_n$ corresponds to the first $n$ points of a scrambled Sobol' sequence in $[0,1]$. The predictor $\eta_n$ and $\hISE_{BLP}(\cdot)$ are constructed as in Section~\ref{S:simulations-m-n-d}, i.e., respectively with the kernels $K^{(p)}(\xb,\xb')=\psi_{5/2,\mt_p}(\|\xb-\xb'\|)$ and $K^{(e)}(\xb,\xb')=\psi_{\IM,\mt_{\rm BLP}}(\|\xb-\xb'\|)$ given by \eqref{psi52} and \eqref{psiIM}; $\mt_p$ and $\mt_{\rm BLP}$ satisfy $\psi_{5/2,\mt_p}(D_n[5]) = 0.25$ and $\psi_{\IM,\mt_{\rm BLP}}(D_n[5]) = 0.25$, see Section~\ref{S:behavior-d-n}. Figure~\ref{F:fdeceptive-eta-d1_n5_nn5_theta1-LOO} shows $f(x)$ and $\eta_n(x)$, $x\in[0,1]$, for $n=5$ (left) and $n=15$ (right).

In the first case, with $n=5$ ($D_n[5]=0.9750$, $\mt_p\simeq 1.63$ and $\mt_{\rm BLP}\simeq 1.78$), $\eta_n$ is very smooth, $\max_i \mve_{-i}^2 \simeq  0.0771$, $\hISE_{LOO}(\eta_n)\simeq 0.029$ and $\hISE_{BLP}(\eta_n)\simeq 3.26\,10^{-4}$, whereas $\ISE(\eta_n)\simeq 0.125$. At the same time, the predictor $\overline{\eta_n}$ given by the empirical mean, $\overline{\eta_n}=\1b_n\TT\yb_n/n$, has larger estimated ISE: $\hISE_{LOO}(\overline{\eta_n})\simeq 0.095$ and $\hISE_{BLP}(\overline{\eta_n})\simeq 0.061$. The inaccuracy of $\eta_n$ thus remains undetected (the true ISE is $\ISE(\overline{\eta_n})\simeq 0.0731$, showing that $\eta_n$ is indeed worse than $\overline{\eta_n}$). It would be easily revealed by additional evaluations of $f$ on a set of test points $z_i$ (unless by bad luck the $z_i$ are such that $\eta_n(z_i)\approx f(z_i)$).

The situation improves with the use of a richer design. When $n=15$, (with $D_n[5]=0.267$, $\mt_p\simeq 5.97$ and $\mt_{\rm BLP}\simeq 6.49$), $\eta_n$ is much closer to $f$ although $\max_i \mve_{-i}^2 \simeq  0.305$ is much larger than before. We have now $\hISE_{LOO}(\eta_n)\simeq 0.1073$, $\hISE_{BLP}(\eta_n)\simeq 0.0053$ and $\ISE(\eta_n)\simeq 0.0097$; $\hISE_{BLP}(\eta_n)$ thus underestimates $\ISE(\eta_n)$ by a factor of 2, but $\hISE_{LOO}(\eta_n)$ overestimates it by a factor of 10. For the empirical mean $\overline{\eta_n}=\1b_n\TT\yb_n/n$, we obtain $\hISE_{LOO}(\overline{\eta_n})\simeq 0.0758$, suggesting a better prediction of $f$ by $\overline{\eta_n}$ than by $\eta_n$, whereas  $\hISE_{BLP}(\overline{\eta_n})\simeq 0.0610>\hISE_{BLP}(\eta_n)\simeq 0.0053$, indicating that $\eta_n$ is a better predictor than $\overline{\eta_n}$ (and indeed, $\ISE(\overline{\eta_n})\simeq 0.0625 > \ISE(\eta_n)\simeq 0.0097$).

\section{Matlab code for calculating $\hISE_{BLP}(\eta_n)$ and $\hISE_{BLUP}(\eta_n)$}\label{S:Matlab-code}

\begin{verbatim}
function [ ise_BLP,ise_BLP_unbiased,ise_LOOCV,BLUE ] = ...
    ISE_WLOO_BLP( yn,Xn,Rn,Xtest,Mu,WnN,Kn_BLP,knx_BLP,nugget,constant_term )
% function [ ise_BLP,ise_BLP_unbiased,ise_LOOCV,BLUE ] = ...
%    ISE_WLOO_BLP( yn,Xn,Rn,Xtest,Mu,WnN,Kn_BLP,knx_BLP,nugget,constant_term )
% Xn = d*n matrix of design points
% yn = n (column) vector of observations
% Rn = n*n matrix such that LOO errors = Rn'*yn for the predictor evaluated
% Xtest = d*N matrix of test points (the ISE is estimated by the empirical
%   mean on Xtest)
% Mu = 1*N row vector of weights (with sum = 1) defining the measure on Xtest
% WnN = n*N matrix, whose ith column is the vector of weights at the ith
%   test point for the predictor evaluated (predictions over Xtest are
%   given by yn'*WnN)
% Kn_BLP = n*n kernel matrix, with the kernel used for ise_BLP estimation
% knx_BLP = n*N kernel matrix for the n design points and N test points
% nugget = nugget parameter (presence of an additive noise with variance
%   nugget*s2, with s2 the GP variance)
% ise_BLP = ISE estimated by construction of the BLP
% ise_BLP_unbiased = unbiased version of the above
% ise_LOOCV = classical LOOCV estimate (sum of squares of LOO errors)/n
% if constant_term == 1, the construction assumes that there is a constant
%   term in the model and estimates it to correct the ISE estimation
% BLUE = estimator of the constant term in the location model

[~,n]=size(Xn);
BLUE=NaN;
LOO_errors=Rn'*yn;
LOO_errors_squared=LOO_errors.^2;
ise_LOOCV=mean(LOO_errors_squared); % standard LOO ISE estimator
% squared error for the predictor and assumed model (as if no trend)
rhon2_12=1+nugget-2*sum(WnN.*knx_BLP,1)+sum(WnN.*(Kn_BLP*WnN),1);
tn_12=knx_BLP-Kn_BLP*WnN;
un=ones(n,1); Knm1un=Kn_BLP\un;
if constant_term == 1
    % include a constant term in the model for the BLP estimator
    constant=yn'*Knm1un/(un'*Knm1un); % = BLUE of the constant
    BLUE=constant;
    % remove the constant, shift the observations, and proceed as for a
    % model with zero mean, add a suitable constant to the estimated ISE
    yn=yn-constant*un;
    LOO_errors=Rn'*yn;
    LOO_errors_squared=LOO_errors.^2;
    ISE_add_constant=constant^2*mean((un'*WnN-1).^2);
end
Qdum=Rn'*Kn_BLP*Rn;
um=diag(Qdum);
cm=um*rhon2_12+2*(Rn'*tn_12).^2;
Sm=um*um'+2*Qdum.^2;
% weights for the BLP
wLOO=(Sm\cm);
% weights for the BLUP
wLOO_unbiased=Sm\(cm+um*(rhon2_12-um'*wLOO)/(um'*(Sm\um)));
error2_WLOOCV=LOO_errors_squared'*wLOO; error2_WLOOCV=max(error2_WLOOCV,0);
error2_WLOOCV_unbiased=LOO_errors_squared'*wLOO_unbiased;
        error2_WLOOCV_unbiased=max(error2_WLOOCV_unbiased,0);
ise_BLP=sum(Mu.*error2_WLOOCV);
ise_BLP_unbiased=sum(Mu.*error2_WLOOCV_unbiased);
if constant_term==1
    ise_BLP=ise_BLP+ISE_add_constant;
    ise_BLP_unbiased=ise_BLP_unbiased+ISE_add_constant;
end
end
\end{verbatim}

\bibliographystyle{siamplain}


\end{document}